\documentclass[preprint,12pt]{jmlr}

\title[BAI with Unknown Distribution]{Best Arm Identification with Unknown \\
Post-Action Context Distribution}
\usepackage{times}

\makeatletter\@namedef{ver@subcaption.sty}{2099/01/01}\makeatother
\usepackage[utf8]{inputenc} 
\usepackage[T1]{fontenc}    
\usepackage{hyperref}       
\usepackage{url}            
\usepackage{booktabs}       
\usepackage{amsfonts}       
\usepackage{nicefrac}       
\usepackage{microtype}      
\usepackage{xcolor}   

\usepackage{microtype}
\usepackage{graphicx}
\usepackage{booktabs}
\usepackage{enumitem}

\usepackage{mathtools}
\DeclarePairedDelimiter\norm{\lVert}{\rVert}

\usepackage{algorithm}
\usepackage{algpseudocode}

\usepackage{tikz}
\usetikzlibrary{positioning}
\usetikzlibrary{shapes, arrows, arrows.meta, positioning}
\usepackage{caption}
\usepackage{subcaption}

\usepackage{amsmath}
\usepackage{amssymb}
\usepackage{mathtools}
\usepackage{thmtools,thm-restate}
\usepackage{multirow}

\usepackage{wrapfig}

\usepackage{mwe}
\usepackage{tikz} 
\usetikzlibrary{arrows.meta,shapes} 
\usepackage{pgf}
\usepackage{bbm}
\usepackage{extarrows}

\newcommand{\M}[0]{\mathcal{M}}
\newcommand{\A}[0]{\mathcal{A}}
\newcommand{\B}[0]{\mathcal{B}}
\newcommand{\V}[0]{\mathcal{V}}

\newcommand{\ocal}[0]{\mathcal{O}}
\newcommand{\pcal}[0]{\mathcal{P}}
\newcommand{\hcal}[0]{\mathcal{H}}
\newcommand{\lcal}[0]{\mathcal{L}}

\newcommand{\E}[0]{\mathcal{E}}
\newcommand{\I}[0]{\mathcal{I}}

\newcommand{\pr}[0]{\mathbb{P}}

\newcommand{\N}[0]{\mathcal{N}}

\newcommand{\wb}{\mathbf{w}}
\newcommand{\qb}{\mathbf{q}}
\newcommand{\pb}{\mathbf{p}}

\newcommand{\lambb}{\mathbf{\lambda}}

\newcommand{\paran}[1]{\left( #1 \right)}

\newcommand{\bmu}[0]{\boldsymbol{\mu}}
\newcommand{\muh}[0]{\hat{\mu}}
\newcommand{\bmuh}[0]{\hat{\bmu}}
\newcommand{\muht}[0]{\hat{\bmu}(t)}
\newcommand{\lamh}[0]{\hat{\Lambda}}
\newcommand{\lamht}[0]{\hat{\Lambda}(t)}

\newcommand{\delhit}[1]{\hat{\Delta}_{#1}(t)}
\newcommand{\Ah}[0]{\hat{\A}}
\newcommand{\Aht}[0]{\hat{\A}(t)}

\newcommand{\logdel}[0]{\ln(1 / \delta)}

\newcommand{\bettdelmu}[0]{\beta_{\bmu}(t, \delta)}

\newcommand{\bettdel}[0]{\beta(t, \delta)}

\newcommand{\istar}[0]{i^*(\bmu, \A)}
\newcommand{\istarat}[0]{i_{t}^{*}}
\newcommand{\istart}[0]{i_{t}^{*}}
\newcommand{\wstar}[0]{\mathbf{w}^*(\bmu, \A)}
\newcommand{\wstarht}[0]{\mathbf{w}^*(\muht, \Aht)}

\newcommand{\wstart}[0]{\mathbf{w}^*(t)}
\newcommand{\wepsstar}[1]{\mathbf{w}^*_{#1}(\bmu, \A)}
\newcommand{\nbarxt}[0]{\overline{\mathbf{N}}^X(t)}

\newcommand{\nbarzt}[0]{\overline{\mathbf{N}}^Z(t)}
\newcommand{\pbart}[0]{\overline{\mathbf{p}}(t)}

\newcommand{\dls}[2]{dist_{L^2}\left(#1, #2\right)}

\newcommand{\eventmu}{\E_{T}^{\bmu}(\epsilon)}
\newcommand{\eventa}{\E_{T}^{\A}(\epsilon)}
\newcommand{\eventx}{\E_{T}^{X}(\epsilon)}

\newcommand{\expec}[1]{\mathbb{E}[#1]}

\newcommand{\numaction}[0]{N^{X}_{i}(t)}
\newcommand{\numcontext}[0]{N^{Z}_{j}(t)}
\newcommand{\numac}[0]{N_{j, i}(t)}

\newcommand{\argmax}{\operatornamewithlimits{argmax}}

\usepackage{titletoc}

\titlecontents{apdsection}[0em]
  {\addvspace{0.6em}\normalfont}
  {\contentslabel{2.5em}}
  {}
  {\titlerule*[0.6pc]{.}\contentspage}

\usepackage{xcolor}

\author{%
 \Name{Amir Mohammad Abouei \nametag{\thanks{These authors contributed equally to this work.}}} 
 \Email{amir.abouei@epfl.ch} \\
 \addr EPFL
 \AND
 \Name{Mohammad Shahverdikondori \nametag{\footnotemark[1]}} 
 \Email{mohammad.shahverdikondori@epfl.ch} \\
 \addr EPFL
  \AND
   \Name{Emad Emamjomeh}
    \Email{emad.emamjomeh@epfl.ch} \\
  \addr EPFL
  \AND
  \Name{Negar Kiyavash}
  \Email{negar.kiyavash@epfl.ch} \\
  \addr EPFL
}

\begin{document}

\maketitle
\thispagestyle{empty}
\pagestyle{empty}

\begin{abstract}%
    We study best arm identification (BAI) with post-action context in the fixed-confidence setting. After choosing an action, the learner observes a post-action context together with a reward whose distribution depends only on this context. 
    Incorporating such intermediate information can substantially reduce the sample complexity of identifying the best arm. While \citet{shahverdikondori2025optimal} analyzed the case where the distribution of contexts given actions is known, we address the more challenging and practically relevant case where this distribution is unknown.  
    We derive an instance-dependent sample complexity lower bound and  propose a Track-and-Stop algorithm that matches it asymptotically with access to an optimization oracle. Our stopping rule uses a new concentration inequality that jointly controls uncertainty in the context distributions and reward means, removing the factor of two incurred by separate concentration bounds.            
    Implementing the algorithm generally requires solving nonconvex optimization problems. We propose a convex relaxation with instance-dependent approximation guarantees, including asymptotic tightness on hard Gaussian instances.
    Experiments demonstrate that exploiting post-action context reduces the sample complexity relative to baselines that ignore it, even when the context distributions are unknown.
\end{abstract}

\section{Introduction}

Best-arm identification (BAI) in multi-armed bandits (MAB) refers to a class of problems where an agent sequentially interacts with an unknown environment and receives rewards based on its actions. The goal of the agent is to identify the action with the highest mean reward. BAI is typically studied in two settings: (i) the fixed-budget setting, where the time horizon is fixed \citep{budget-confidence-gabillon2012best, SR-audibert2010best, SH-karnin2013almost, lb-budget-carpentier2016tight, budget-new-wang2023best}, and (ii) the fixed-confidence setting, where the objective is to identify the optimal arm with as few interactions as possible while guaranteeing a prescribed level of confidence \citep{even2002pac, lilucb-jamieson2014lil, neurips2022-toptwo}. In this paper, we focus on \emph{best-arm identification in the fixed-confidence setting}.

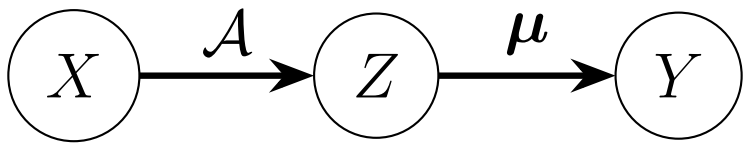
\begin{figure}[t]
        \centering
        \begin{tikzpicture}
            \tikzset{line width=2pt, outer sep=0pt,
            ell/.style={draw,fill=white, inner sep=2pt,
            line width=2pt},
            };
                \node[name=X, shape=circle, draw] at (-3,0){$X$};
                \node[name=Z, shape=circle, draw] at (-1,0){$Z$};
                \node[name=Y, shape=circle, draw] at (1,0){$Y$};
            
            \begin{scope}[>={Stealth[black]},
                          every edge/.style={draw=black,very thick}]
                \path[->] (X) edge node[midway, above] {$\A$} (Z);
                \path[->] (Z) edge node[midway, above] {$\bmu$} (Y);
            \end{scope}
        \end{tikzpicture}
        \caption{Post-action context graph: action $X$, context $Z$, and reward $Y$.}
        \label{fig: causal_graph}
\end{figure}

We investigate a particular form of side information known as \emph{mediator feedback} or \emph{post-action context}~\citep{mann2019learning, metelli2021policy, causal-benign-bilodeau2022adaptively, mediator-cum-eldowa2024information, causal-pareto-liu2024causal}. In this setting, the agent receives a post-action context, $Z$, in addition to the reward, $Y$, after choosing an action, $X$. As illustrated by the causal graph in Figure~\ref{fig: causal_graph},
the reward depends on the chosen action only through the observed
context: $\pr(Y\mid X,Z)=\pr(Y\mid Z)$. For finite action and context spaces, let $\A$ denote the context
probability matrix, with entries
$\A_{j,i}=\pr(Z=j\mid X=i)$, and let $\bmu$ denote the context
reward means, with entries $\mu_j=\expec{Y\mid Z=j}$.
We consider the setting where both $\A$ and $\bmu$ are unknown
to the learner, and also study the special case where $\bmu$ is known.


\begin{table}[t] 
        \centering
        \caption{Different variants of problems with post-action context.}
        \label{table: intro}
        \small
            \begin{tabular} {c | c |c} 
                \toprule
                 Method & Information & Target \\
                \hline
                \citet{mediator-poiani2023pure} & $\A$ & Best Context \\
                \citet{mediator-poiani2023pure, mediator2-reddy2023best} & $\emptyset$ & Best Context \\
                \citet{causal-cucb-lu2020regret, mediator-cum-eldowa2024information} & $\A$ & Regret \\
                \citet{lattimore2016causal}  & $\A$ & Simple Regret \\
                \citet{shahverdikondori2025optimal} & $\A$ & Best Arm \\
                \textbf{Ours} & $\mu$ or $\emptyset$ & Best Arm \\
                \bottomrule
            \end{tabular}
\end{table}

\textbf{Example.} Consider a set of experts, such as doctors, each following an unknown strategy that specifies a distribution over treatment options, such as drug and dosage combinations. Let $X$ denote the selected expert, $Z$ the prescribed treatment, and $Y$ the patient's health outcome. We assume that the expert affects the outcome only through the prescribed treatment. In each round, the learner selects an expert and observes both the treatment and the resulting outcome. The objective is to identify the expert with the highest expected reward. Our framework also covers unobserved confounding between $X$ and $Y$
when actions are interventions on $X$ (see Appendix~\ref{apd: causal interpretation}).


Observing the context can substantially reduce the number of samples
needed to identify the best action. Rewards observed under the same
context $Z=j$ can be pooled across actions to estimate the shared
mean $\mu_j$. \citet{shahverdikondori2025optimal} exploit this structure
when the context probability matrix $\A$ is known.

In the general setting where both $\A$ and $\bmu$ are unknown, the learner must estimate both the context
distributions and the context reward means. Since the expected reward
of action $i$ is $\A_i^\top\bmu$, uncertainty in both quantities
affects the identification of the best action. This introduces two
challenges. Statistically, the stopping rule must jointly control
uncertainty in $\A$ and $\bmu$. Computationally, evaluating the
information gained from a sampling strategy requires optimizing over
alternative context distributions and reward means. These quantities
are coupled through a bilinear constraint, making the optimization
generally nonconvex.

\textbf{Related Work.} In fixed-confidence BAI, an instance-dependent lower bound on
the expected sample complexity is characterized by an optimization over arm sampling proportions \citep{track-stop-garivier2016optimal}. The Track-and-Stop (TaS) algorithm matches this bound as $\delta \to 0$ by solving this problem at the current estimates, sampling the arms so that their empirical proportions follow the solution, and stopping once a generalized likelihood ratio (GLR) statistic exceeds a threshold. It has been extended to structured problems such as linear bandits \citep{linear-lazy-jedra2020optimal}, multiple correct answers \citep{mutiple-correct-answers-degenne2019pure}, and Markov decision processes \citep{rl2-al2021navigating, al2021adaptive}. Our setting needs new versions of the threshold and of the optimization step. The GLR statistic has a categorical part (for $\A$) and an exponential-family part (for $\bmu$), and summing a threshold for each part \citep{rl2-al2021navigating, al2021adaptive} leaves a factor-2 gap to the lower bound, which our joint threshold removes. The optimization is typically handled by Frank-Wolfe \citep{frank-wolf-wang2021fast} or by gamification \citep{linear-degenne2020gamification}, both of which still need to minimize over the alternative instances, a set that is in general non-convex here.

The setting of post-action context, also referred to as mediator feedback, has been studied for regret minimization \citep{mediator-cum-eldowa2024information, nair2021budgeted, causal-pareto-liu2024causal} and for pure exploration. \citet{mediator-poiani2023pure, mediator2-reddy2023best} identify the best \emph{context}, i.e., the largest entry of $\bmu$, with $\A$ known or unknown. There the asymptotic complexity does not depend on whether $\A$ is known, since only $\bmu$ is estimated, unlike in best arm identification, where either $\A$ or $\bmu$ may be the harder parameter to estimate. For best arm identification, \citet{shahverdikondori2025optimal} proposed an asymptotically optimal algorithm that assumes $\A$ is known. We study the general case, in which both $\A$ and $\bmu$ are unknown, which also covers known $\bmu$. Table~\ref{table: intro} summarizes these works. Our setting also falls within causal bandits, with actions interpreted as interventions on $X$; a detailed comparison appears in Appendix~\ref{apd: related work} and \ref{apd: causal interpretation}.

\textbf{Contributions and Organization.} Our contributions are as follows.
\begin{itemize}[leftmargin=*, itemsep=0pt, topsep=0pt]
    \item We derive an instance-dependent sample complexity lower
    bound and propose a Track-and-Stop algorithm that attains it
    asymptotically with access to an optimization oracle. Our results
    cover both unknown $\bmu$ and the special case of known $\bmu$ (Section~\ref{sec: algorithm}).
    \item We introduce a joint stopping threshold that removes the
factor-of-two overhead of separate thresholds and sharpens the
RL sample complexity guarantees of
\citet{rl2-al2021navigating, al2021adaptive}
(Section~\ref{sec: algorithm}).
    \item We study the optimization problems involved in computing the optimal allocation. We show that the cases of known $\bmu$ and $k=2$ contexts reduce to convex programs. For $k\ge3$, we show that natural convexification approaches can fail: the set of alternative instances need not be a finite union of convex sets, and replacing it by its convex hull can make the objective vanish (Section~\ref{sec: allocation}).
    \item We propose a convex relaxation that leads to an efficient algorithm whose asymptotic sample complexity is within an instance-dependent factor of the optimum on every instance. This factor approaches $1$ for a class of hard Gaussian instances (Subsection~\ref{sec: relaxation}).
\end{itemize}

\section{Notation and Problem Setup}\label{sec: preli}
Let $[n] = \{1,2,\ldots,n\}$ and denote by $\Delta^{n-1}$ the $(n-1)$-dimensional standard simplex,
$
\Delta^{n-1} = \big\{w \in \mathbb{R}^n \mid w_i \geq 0 \ \wedge \ \sum_{i=1}^n w_i = 1\big\},
$
with $\Delta^{n-1}_+ \subset \Delta^{n-1}$ denoting the subset of strictly positive vectors.
We consider a stochastic multi-armed bandit environment $\V$ with $n$ arms. At each round $t$, the learner selects an action $X=i_t \in [n]$ and observes an outcome $(z_t,y_t)$, where $z_t$ is the realization of the post-action context $Z$ and $y_t$ is the reward $Y$.

The context variable $Z$ takes values in $[k]$, with probabilities depending on the chosen action. Let $\A = [\A_1 \mid \cdots \mid \A_n] \in \mathbb{R}^{k \times n}$ denote the (unknown) context probability matrix, where $\A_{j,i} = \pr(Z=j \mid X=i)$. Based on the causal graph in Figure \ref{fig: causal_graph}, we assume that the reward distribution depends only on the context, i.e.,
$
\pr(Y \mid X=i, Z=j) = \pr(Y \mid Z=j), \quad \forall i \in [n], \ j \in [k].
$
For any $i \in [n]$, let $P_{\bmu,\A}^i = \pr(Z, Y \mid X=i)$ denote the joint distribution of $(Z,Y)$ conditioned on choosing action $i$. Since the reward is conditionally independent of the action given the context, the expected reward of arm $i$ is
$
\mathbb{E}[Y \mid X=i] = \sum_{j=1}^k \pr(Z=j \mid X=i) \, \mathbb{E}[Y \mid Z=j].
$

Define $\bmu \in \mathbb{R}^k$ with entries $\mu_j \triangleq \mathbb{E}[Y \mid Z=j]$. Then the expected reward of action $i$ can be written as
$
\mathbb{E}[Y \mid X=i] = \sum_{j=1}^k \A_{j,i} \mu_j = \A_i^\top \bmu.
$

Following a common modeling framework in the pure-exploration bandit
literature \citep{kaufmann2016complexity}, we assume that, for each $j \in [k]$, the conditional distribution $\pr(Y \mid Z=j)$ belongs to a one-parameter exponential family of the form
$
    \pcal = \Big\{ (\nu_{\theta})_{\theta \in \Theta} : d\nu_{\theta} /d\xi = \exp\!\big(\theta x - b(\theta)\big) \Big\},
$
where $\Theta \subset \mathbb{R}$, $\xi$ is a reference measure on $\mathbb{R}$ and $b: \Theta \rightarrow \mathbb{R}$ is a strictly convex, twice-differentiable function known as the log-partition function \citep{kaufmann2016complexity}. Each distribution $v_\theta \in \pcal$ is uniquely characterized by its mean $\dot{b}(\theta)$~\citep{exp-family-cappe2013kullback}, where $\dot{b}$ denotes the first derivative of $b$.

Hence, $\pr(Y \mid Z)$ is fully specified by the means vector $\bmu$ defined above. Common examples of distributions in $\pcal$ include Gaussian with known variance and Bernoulli distributions.

For two probability measures $P$ and $Q$, the Kullback–Leibler (KL) divergence is denoted by $\mathrm{KL}(P,Q)$. If $P=\nu_\theta$ and $Q=\nu_{\theta'}$ with means $\mu = \dot{b}(\theta)$ and $\mu' = \dot{b}(\theta')$, we define $d(\mu,\mu')$ as a KL divergence between $\nu_{\theta}$ and $\nu_{\theta'}$. The KL divergence between two Bernoulli distributions with parameters $\mu$ and $\mu'$ defined as $d_B(\mu, \mu') \triangleq \mu \ln(\frac{\mu}{\mu'}) + (1- \mu) \ln(\frac{1- \mu}{1 - \mu'})$. In addition, for any pair of vectors $\textbf{p}, \textbf{q} \in \Delta^{n-1}_+$, we define
$
d(\textbf{p}, \textbf{q} ) \triangleq \sum_{i \in [n]} p_i \ln \!\left(\frac{p_i}{q_i}\right),
$
which corresponds to the KL divergence between the categorical distributions induced by $\textbf{p}$ and $\textbf{q}$.

Based on the above assumptions, any bandit instance with post-action context can be represented by the pair $(\bmu,\A)$. For such an instance, the best arm is defined as
$
\istar \triangleq \argmax_{i \in [n]} \A_i^\top \bmu.
$
We assume that $\istar$ is unique, $\bmu \in (0, 1)^{k}$, and each context has a nonzero probability under each action, i.e., $\min_{i \in [n], j \in [k]} \A_{j,i} > 0$. For simplicity, we occasionally abbreviate $\istar$ as $i^*$. Let $\I$ denote the set of all instances $(\bmu,\A)$ satisfying these assumptions.

We study the problem in the \textit{fixed-confidence} setting: for a prescribed $\delta \in (0,1)$, the learner must identify $\istar$ with probability of error at most $\delta$. A sequential algorithm consists of: (i) a \emph{sampling rule} choosing $i_t$ based on the past history $\mathcal{H}_{t-1} = (i_1,(z_1,y_1),\ldots,i_{t-1},(z_{t-1},y_{t-1}))$, (ii) a \emph{stopping rule} $\tau$ that decides when to terminate, and (iii) a \emph{decision rule} $\hat{i}_\tau$ that recommends an arm at time $\tau$.

An algorithm is \textit{$\delta$-correct} if, for any $(\bmu,\A) \in \I$, it satisfies
\begin{align*}
    \pr_{\bmu,\A} \paran{\tau < \infty, \ \hat{i}_\tau \neq \istar} \leq \delta .
\end{align*}
Our goal is to design $\delta$-correct algorithms with minimum expected stopping time $\mathbb{E}_{\bmu,\A}[\tau]$.

For each arm $i$, the sub-optimality gap is defined as
$ \Delta_i \triangleq \A_{i^*}^\top \bmu - \A_i^\top \bmu.$ Define $\bmuh(t), \Aht, \delhit{i} $ as the empirical estimates of $\bmu, \A, \Delta_i$, respectively, at round $t$, and let $\istart$ denote the corresponding empirical best arm.
Finally, $N^X_i(t)$, $N^Z_j(t)$, and $N_{j,i}(t)$ denote the number of times action $i$, context $j$, and pair $(i,j)$ have been observed up to round $t$, respectively.

\subsection{Lower Bounds}

A lower bound on the expected sample complexity of any $\delta$-correct algorithm was established in \citet{track-stop-garivier2016optimal} for classic bandits and extended to bandits with post-action context in \citet{shahverdikondori2025optimal}. It is expressed through the set of alternative instances
$$
\text{Alt}(\bmu, \A) \triangleq \big\{ (\bmu', \A') \in \I \mid \istar \neq i^*(\bmu', \A') \big\},
$$
which consists of all instances that induce a different best arm. For any instance $(\bmu, \A) \in \I$ and any vectors $\wb \in \mathbb{R}^n$ and $\qb \in \mathbb{R}^k$, we define the \emph{transportation cost}
\begin{align} \label{eq: unknown f formula}
    F(\bmu, \A, \wb, \qb) \triangleq \inf_{(\lambb, \B) \in \text{Alt}(\bmu, \A)} \ \sum_{i \in [n]} w_i d \paran{\A_i, \B_i} + \sum_{j \in [k]} q_j d \paran{\mu_j, \lambda_j},
\end{align}
and its counterpart $F_{\bmu}(\A, \wb)$ for known $\bmu$, in which the alternatives cannot change the rewards, i.e., $\lambb = \bmu$, so that the second sum vanishes:
\begin{align*}
    F_{\bmu}(\A, \wb) \triangleq \inf_{\B : (\bmu, \B) \in \text{Alt}(\bmu, \A)} \sum_{i \in [n]} w_i d \paran{\A_i, \B_i} = \min_{i \neq i^*} \inf_{\B_1, \B_2 : \B_1^{\top} \bmu < \B_2^{\top} \bmu} w_{i^*} d \paran{\A_{i^*}, \B_1} + w_i d \paran{\A_i, \B_2},
\end{align*}
where $\B_1, \B_2 \in \Delta^{k-1}_+$. The second equality holds because only the columns $i$ and $i^*$ need to change. We write $\A\wb$ for the vector of expected context frequencies, $(\A\wb)_j = \sum_{i \in [n]} w_i \A_{j,i}$.

\begin{restatable}{proposition}{lowerBound} \label{pro: lower}
    Let $\delta \in (0,1)$ and $(\bmu, \A) \in \I$. Any $\delta$-correct algorithm with stopping time $\tau_{\delta}$ satisfies $\mathbb{E}_{\bmu, \A}[\tau_{\delta}] \geq T \, d_B(\delta, 1- \delta)$, and hence $\liminf_{\delta \to 0} \mathbb{E}_{\bmu, \A}[\tau_{\delta}] / \logdel \geq T$, where
    \begin{align*}
        T &= T^*(\bmu, \A) \triangleq \Big(\sup_{\wb \in \Delta^{n-1}} F(\bmu, \A, \wb, \A\wb)\Big)^{-1} &&\text{if } \bmu \text{ is unknown,} \\
        T &= T^*_{\bmu}(\A) \triangleq \Big(\sup_{\wb \in \Delta^{n-1}} F_{\bmu}(\A, \wb)\Big)^{-1} &&\text{if } \bmu \text{ is known.}
    \end{align*}
\end{restatable}

When $\bmu$ is known, alternatives are restricted to changes in $\A$, which implies $T^*_{\bmu}(\A) \leq T^*(\bmu,\A)$. Proposition~\ref{pro: lower} follows from the general bound of \citet{shahverdikondori2025optimal} by evaluating the KL divergence between the joint distributions of $(Z, Y)$ (details in Appendix~\ref{apd: lower bound}). The transportation cost has a simple interpretation \citep{kaufmann2016complexity}: if the learner samples arms in proportion to $\wb$, the infimum identifies the closest alternative instance in which some arm $i \neq i^*$ becomes the best arm. For unknown $\bmu$, it has two terms: the first, $\sum_{i} w_i \, d(\A_i, \B_i)$, is the cost of perturbing the context distributions, and the second, $\sum_{j} (\A\wb)_j \, d(\mu_j, \lambda_j)$, is the cost of perturbing the reward means. Maximizing over $\wb$ yields the optimal allocation, which any asymptotically optimal algorithm tracks.


\section{Stopping Rule and Optimal Algorithm}\label{sec: algorithm}
We now present an algorithm that handles both settings (known and unknown $\bmu$). It combines a GLR stopping rule with a new threshold, and the G-tracking sampling rule. We first fix, for each setting, the three quantities that the algorithm uses: the GLR statistic, the stopping threshold, and the target allocation. Writing $\nbarxt = N^X(t)/t$ and $\nbarzt = N^Z(t)/t$ for the empirical frequencies of arms and contexts, these are

\begin{table}
\centering
\small
\caption{Quantities used by the algorithm in each setting.}
\label{table: alg-comp}
\begin{tabular}{l|c|c} 
     & Unknown $\bmu$ & Known $\bmu$ \\ \hline
    GLR statistic & $\lamht = t F\paran{\muht, \Aht, \nbarxt, \nbarzt}$ & $\lamh_{\bmu}(t) = t F_{\bmu}\paran{\Aht, \nbarxt}$ \\
    Threshold & $\bettdel$ in \eqref{eq: new beta} & $\bettdelmu$ in \eqref{eq: known beta} \\
    Allocation $\wstart$ & $ \argmax_{\wb} F\paran{\muht, \Aht, \wb, \Aht\wb}$ & $ \argmax_{\wb} F_{\bmu}\paran{\Aht, \wb}$ 
\end{tabular} 
\end{table}

\textbf{Stopping rule.} The GLR statistic $\lamht$ measures the evidence that the empirical best arm $\istart$ is the best arm (derivation in Appendix~\ref{apd: glr derivation}); the algorithm stops at
\begin{align} \label{eq: stop rule}
    \tau_{\delta} \triangleq \inf \{ t \in \mathbb{N} \mid \lamht > \bettdel \}
\end{align}
(with $\lamh_{\bmu}(t)$ and $\bettdelmu$ in the known-$\bmu$ setting). The threshold is chosen to ensure $\delta$-correctness while keeping the
stopping time short. For known $\bmu$, only the context distributions are estimated, and we use the threshold of \citet{rl2-al2021navigating, jonsson2020planning} for categorical distributions,
\begin{align} \label{eq: known beta}
    \bettdelmu \triangleq \logdel + (k-1) \sum_{i \in [n]} \ln \paran{e \paran{1 + \frac{N^X_i(t)}{k-1}}}.
\end{align}
\textbf{A new threshold for unknown $\bmu$.} When $\bmu$ is unknown, the GLR statistic combines a categorical term
for $\A$ and an exponential-family term for $\bmu$, which the threshold
must control simultaneously. The standard approach adds separate
thresholds at confidence level $\delta/2$ \citep{rl2-al2021navigating, al2021adaptive}, giving a leading term of $2\logdel$ and introducing a factor of $2$ in the sample complexity (Appendix~\ref{apd: sub-optimal threshold}), as in Theorem~6 of \citet{rl2-al2021navigating}. We propose a new threshold that removes this factor by controlling the two terms \emph{jointly}. Combining martingale constructions for categorical and exponential-family variables, our threshold reduces the leading confidence term to $\logdel$
\begin{align} \label{eq: new beta}
    \bettdel \triangleq (k-1) \sum_{i \in [n]} \ln \paran{e \paran{1 + \frac{N^X_i(t)}{k-1}}} + 3 \sum_{j \in [k]} \ln \paran{1 + \ln \paran{N^Z_j(t)}} + k C_{\text{exp}} \paran{\frac{\logdel}{k}},
\end{align}
where $C_{\text{exp}}(x)\approx x+\ln(x)$ is precisely defined in
Appendix~\ref{apd: glr derivation}. Since
$kC_{\text{exp}}(\logdel/k)\sim\logdel$ as $\delta\to0$, the threshold
has a leading confidence term of $\logdel$ rather than $2\logdel$.
This removes the factor of $2$ and, as Theorem~\ref{thm : upper} shows,
allows the resulting sample complexity to match the lower bound
asymptotically. The threshold follows from the deviation inequality
below, which jointly controls categorical and exponential-family
variables uniformly over time. Beyond our setting, this inequality
also sharpens the reinforcement learning guarantees of
\citet{rl2-al2021navigating, al2021adaptive}, reducing their leading
sample complexity bounds by the same factor of $2$.

\begin{restatable}{lemma} {newThreshold} \label{lem: new threshold}
    For any $\delta \in (0,1)$, we have
    \begin{align}
        \pr \Bigg( \exists t \in \mathbb{N} : \sum_{i \in [n]} N_i^X(t) d \paran{\Ah_i(t), \A_i} 
        + \sum_{j \in [k]} N^Z_j(t) d \paran{\muh_j(t), \mu_j} > \bettdel \Bigg) \leq \delta.    \nonumber
    \end{align}
\end{restatable}

The following lemma guarantees that the stopping rule is
$\delta$-correct in both settings, regardless of the sampling rule.

\begin{restatable}{lemma}{stoppingLemma}\label{lem: stopping}
For any $\delta\in(0,1)$, any algorithm using the stopping rule
\eqref{eq: stop rule}, with the statistic and threshold specified
in the corresponding column of Table~\ref{table: alg-comp},
is $\delta$-correct.
\end{restatable}

\textbf{Sampling rule.} The sampling rule tracks the allocation $\wstart$ of Table \ref{table: alg-comp}, which is the optimal allocation of the lower bound evaluated at the current estimates. This maximizer need not be unique, so we use G-tracking \citep{shahverdikondori2025optimal}, which, unlike D-tracking \citep{mutiple-correct-answers-degenne2019pure}, accommodates set-valued optimizers. G-tracking was introduced for settings where actions are chosen probabilistically from a given set of distributions. Since arms can be selected directly in our setting, G-tracking takes the following simple closed form.

Both the empirical frequency $\nbarxt$ and the target allocation $\wstart$ are points of the simplex $\Delta^{n-1} \subset \mathbb{R}^n$. G-tracking draws the straight line that starts at $\nbarxt$ and passes through $\wstart$, and extends it beyond $\wstart$ until it reaches the boundary of the simplex. The intersection point $\pb(t+1) \in \Delta^{n-1}$ is the distribution from which the next arm is sampled, so that the empirical frequency is pushed past $\wstart$ and the average frequency moves toward it. This point can be computed explicitly from the condition that all its coordinates remain nonnegative
\begin{align} \label{eq: p(t+1) in g-tracking}
    \pb(t+1) &= \lambda \wstart + (1-\lambda) \nbarxt, \qquad
    \lambda = \min_{i: \nbarxt_i > \wb^*_i(t)} \frac{\nbarxt_i}{\nbarxt_i - \wb^*_i(t)}.
\end{align}
We augment G-tracking with a forced-exploration step to ensure that
every arm is sampled infinitely often and all estimates converge.
This modification requires a refined analysis
(Appendix~\ref{apd: proof-sample}). At rounds $t+1$ for which
$\sqrt{t+1}$ is an integer, the algorithm samples an arm uniformly
at random. The resulting sampling rule $\pi$ is
\begin{align} \label{eq: g-tracking formula}
    \pi(X_{t+1} = i) =
    \begin{cases}
        \tfrac{1}{n} & \text{if } \sqrt{t+1} \in \mathbb{N}, \\[6pt]
        \pb(t+1)_i & \text{otherwise.}
    \end{cases}
\end{align}
\begin{restatable}{theorem}{upperBound}
\label{thm : upper}
    Consider any instance $(\bmu, \A) \in \I$. Algorithm~\ref{algo: unknown}, given access to an oracle that exactly computes $\wstart$, is $\delta$-correct and achieves
    \begin{equation*}
        \limsup_{\delta \to 0} \frac{\mathbb{E}_{\bmu, \A}[\tau_{\delta}]}{\logdel} \leq T^*(\bmu, \A) \quad \text{(unknown } \bmu \text{)}, \qquad \limsup_{\delta \to 0} \frac{\mathbb{E}_{\bmu, \A}[\tau_{\delta}]}{\logdel} \leq T^*_{\bmu}(\A) \quad \text{(known } \bmu \text{)}.
    \end{equation*}
\end{restatable}

\begin{figure}[t]
    \begin{minipage}[t]{\textwidth}
            \begin{algorithm}[H]
                \caption{G-tracking (known or unknown $\bmu$)}
                \label{algo: unknown}
                \textbf{Input:} \textsc{opt}: oracle that computes 
                $\wstart$; $\bmu$ if it is known.\; \\
                \textbf{Initialization:} Sample until every arm--context pair is observed; set $t$ to the number of pulls and compute the estimates.\; \\
                \While{$\lamht \leq \bettdel$ (or $\lamh_{\bmu}(t) \leq \bettdelmu$ if $\bmu$ is known)}{
                    Compute $\wb^*(t)$ using \textsc{opt}.\; \\
                    Pull arm $X_{t + 1}$ based on G-tracking rule \eqref{eq: g-tracking formula}.\; \\
                    Update $\Aht$, $\bmuh(t)$, and $t \gets t + 1$.\;
                }
                \textbf{Output:} $i^*(\bmuh(t),\Aht)$ (or $i^*(\bmu,\Aht)$ if $\bmu$ is known).\;
            \end{algorithm}
    \end{minipage}\hfill
\end{figure}
\section{Computing the Optimal Allocation}\label{sec: allocation}
Computing the allocation $\wstart$ and the GLR statistic in
Algorithm~\ref{algo: unknown} requires evaluating $F$.
The allocation objective $F(\bmu,\A,\wb,\A\wb)$ is concave in $\wb$,
as it is an infimum of linear functions. The main computational
challenge therefore lies in the minimization over alternative instances.
When $\bmu$ is unknown, the constraint
$\B_i^\top\lambb>\B_{i^*}^\top\lambb$ couples the alternative context
distributions and reward means, making this minimization generally
nonconvex. We first identify two tractable cases in which $F$ can be
computed by solving finitely many convex programs
(Subsection~\ref{sec: easy cases}). We then examine the limitations of
convex decomposition and convex-hull relaxations
(Subsection~\ref{sec: hardness}), before introducing a closed-form convex
relaxation with provable approximation guarantees
(Subsection~\ref{sec: relaxation}).

\subsection{Tractable Cases}\label{sec: easy cases}
If the set of alternative parameters can be written as a finite union of convex sets, evaluating $F$ reduces to solving a convex optimization problem over each set and taking the minimum of their optimal values. Each convex program can be solved efficiently to any desired accuracy.

\textbf{Known $\bmu$.} When $\bmu$ is known, $\lambb=\bmu$ is fixed,
and the constraint $\B_i^\top\bmu>\B_{i^*}^\top\bmu$ is linear in
the alternative context distributions. Thus $\text{Alt}$ is a union
of $n-1$ convex sets, one for each arm $i\ne i^*$. Evaluating
$F_{\bmu}$ therefore reduces to solving $n-1$ convex programs and
taking the minimum of their optimal values. Computing the allocation
$\wstart$ requires maximizing the concave, non-smooth function
$F_{\bmu}(\Aht,\cdot)$ over the simplex. An adaptive variant of the
Frank-Wolfe algorithm \citep{pmlr-v28-jaggi13} can be used for this optimization with a convergence guarantee. \citet{frank-wolf-wang2021fast} study this approach for allocation problems in best-arm identification.

\textbf{Unknown $\bmu$ with two contexts ($k=2$).} With two contexts, the bilinear constraint factorizes and $\text{Alt}$ can be written as a finite union of convex sets. Write $\B_s = (b_s, 1-b_s)^\top$, where $b_s$ is the probability of the first context under arm $s$. Then, for any $(\lambb, \B)$ and any $i \neq i^*$,
\begin{align*}
    (\B_i - \B_{i^*})^\top \lambb = (b_i - b_{i^*})(\lambda_1 - \lambda_2),
\end{align*}
Thus $\B_i^\top \lambb > \B_{i^*}^\top \lambb$ holds if and only if both factors are positive or both are negative. Each case is defined by two linear inequalities, and therefore describes a convex set. Taking the union over all $n-1$ arms $i\ne i^*$ gives
$2(n-1)$ convex sets.

\begin{proposition}\label{prop: k2 decomposition}
Let $k = 2$ and $(\bmu, \A) \in \I$. Then $\text{Alt}(\bmu, \A) = \bigcup_{i \neq i^*} \paran{C_i^{+} \cup C_i^{-}}$, with
\begin{align*}
    C_i^{+} &= \big\{(\lambb, \B) : b_i > b_{i^*}, \ \lambda_1 > \lambda_2 \big\}, &
    C_i^{-} &= \big\{(\lambb, \B) : b_i < b_{i^*}, \ \lambda_1 < \lambda_2 \big\},
\end{align*}
with $\lambb\in(0,1)^2$ and $\B_s=(b_s,1-b_s)^\top\in\Delta_+^1$ for every $s\in[n]$. Each of these $2(n-1)$ sets is convex. Evaluating $F$ therefore reduces to solving $2(n-1)$ convex programs and taking the minimum of their optimal values. Their closures may be used without changing these infima. Algorithm~\ref{algo: unknown} therefore admits an efficient implementation through convex optimization and achieves $\limsup_{\delta \to 0} \mathbb{E}_{\bmu, \A}[\tau_{\delta}]/\logdel \leq T^*(\bmu, \A)$.
\end{proposition}

\subsection{Limitations of Convex Reformulations} \label{sec: hardness}
The two-context case suggests evaluating $F$ by decomposing the
alternative set into finitely many convex sets. For $k\ge3$, however,
such a decomposition need not exist. Another possibility is to
replace the alternative set by its convex hull. The following
proposition shows that this relaxation can reduce the transportation
cost to zero, making it uninformative. Let $F^{\mathrm{conv}}$ denote
$F$ in \eqref{eq: unknown f formula} with $\text{Alt}(\bmu,\A)$
replaced by $\operatorname{conv}(\text{Alt}(\bmu,\A))$.

\begin{restatable}[Limitations of convex reformulations]{proposition}{convexFailure}
\label{prop: convexification failure}
\begin{enumerate}[label=(\roman*), leftmargin=*]
    \item For every $n\ge2$ and $k\ge3$, there exists an instance
    $(\bmu,\A)\in\I$ for which $\text{Alt}(\bmu,\A)$ cannot be
    written as a finite union of convex sets.

    \item For every $n,k\ge2$, there exists a nonempty open set
    $\mathcal U\subset\I$ such that, for every
    $(\bmu,\A)\in\mathcal U$, we have
    $
        (\bmu,\A)\in
        \operatorname{conv}\bigl(\text{Alt}(\bmu,\A)\bigr).
    $
    Consequently, for every $\wb\in\Delta^{n-1}$,
    \[
        F^{\mathrm{conv}}(\bmu,\A,\wb,\A\wb)=0,
    \]
    whereas $F(\bmu,\A,\wb,\A\wb)>0$ whenever
    $\wb\in\Delta_+^{n-1}$.
\end{enumerate}
\end{restatable}

Part (i) shows that the finite convex decomposition used for two
contexts does not extend to all instances with $k\ge3$. Part (ii)
shows why taking the convex hull is also problematic: it can include
the original instance, whose KL cost is zero, making the relaxed
statistic ($F^{\mathrm{conv}}$) uninformative for stopping. This differs from the approach
in Subsection~\ref{sec: easy cases}, where we optimize over each convex
set separately and take the smallest optimal value. Taking the convex
hull adds points between these sets, and those additional points can
destroy the information captured by $F$.


\subsection{Convex Relaxation}\label{sec: relaxation}


When $\bmu$ is unknown and $k>2$, evaluating $F$ generally requires
solving a nonconvex optimization problem. We therefore propose a convex
relaxation based on a closed-form lower bound on the GLR statistic
$\lamht$. As shown below, the resulting relaxed complexity gives an
upper bound on $T^*(\bmu,\A)$. Throughout this subsection, assume
$d(u,v)\ge(u-v)^2/(2\sigma^2)$ for some $\sigma>0$. The inequality
holds with equality for Gaussian rewards of known variance $\sigma^2$
and with $\sigma^2=1/4$ for Bernoulli rewards.

\textbf{Definition of the relaxation.}
Fix $(\bmu,\A)\in\I$. For each arm $i\ne i^*$, let $F_i$ denote
the infimum in \eqref{eq: unknown f formula} restricted to alternatives
in which arm $i$ beats $i^*$. Thus $F=\min_{i\ne i^*}F_i$.
For arm proportions $\wb\in\Delta_+^{n-1}$ and context proportions
$\qb\in\Delta_+^{k-1}$, define
\begin{align*}
    V_l=\sum_j\A_{j,l}(\mu_j-\A_l^\top\bmu)^2,
        \quad l\in\{i,i^*\}, \qquad
    H_i=\frac{V_i}{w_i}+\frac{V_{i^*}}{w_{i^*}}
        +\sigma^2\sum_j
        \frac{(\A_{j,i}-\A_{j,i^*})^2}{q_j}.
\end{align*}
Here $H_i$ is the variance scale of the estimated gap.
Let $\kappa_i$ be the nonnegative coefficient depending on
$\A,\wb,\qb$ defined in Appendix~\ref{apd: convex-relax}. Set
\begin{equation}\label{eq: convex-lb}
    L_i(\wb,\qb)=
    \frac{\Delta_i^2}
    {H_i+\kappa_i\Delta_i+\sqrt{H_i^2+2H_i\kappa_i\Delta_i}},
    \qquad
    L(\bmu,\A,\wb,\qb)=\min_{i\ne i^*}L_i(\wb,\qb).
\end{equation}

Using the empirical frequencies defined in
Section~\ref{sec: algorithm}, the relaxed stopping statistic is
\[
    \lamh^L(t)=tL(\muht,\Aht,\nbarxt,\nbarzt).
\]
We set it to zero if any arm or context is unobserved
or the empirical best arm is tied.

\begin{theorem}\label{thm: simple lower bound}
Let $(\bmu,\A)\in\I$, $\wb\in\Delta_+^{n-1}$ and
$\qb\in\Delta_+^{k-1}$. The bounds in \eqref{eq: convex-lb} satisfy
$L_i\le F_i$ for every $i\ne i^*$ and $L\le F$. Moreover,
$\wb\mapsto L(\bmu,\A,\wb,\A\wb)$ is concave, and
$  \forall t: \lamh^L(t)\le\lamht$.
\end{theorem}

\textbf{Approximation quality.}
We next quantify the gap between $L$ and $F$.
Let $p_i=\min_j\min\{\A_{j,i},\A_{j,i^*}\}$ denote the smallest context
probability among arms $i$ and $i^*$, and define
$
 C_\sigma=2+4\sigma^2+\sqrt{1+4\sigma^2}.
$

\begin{theorem}\label{thm: gap}
Let $(\bmu,\A)\in\I$ and $\wb\in\Delta_+^{n-1}$,
with $\qb=\A\wb$. For every $i\ne i^*$,
\[
    1\le\frac{F_i}{L_i}
    \le C_\sigma\frac{\ln(1/p_i)}{\Delta_i^2}.
\]
If $\Delta_i<\min\{V_i,V_{i^*}\}$, we also have
$$
    \frac{F_i}{L_i}
    \le\frac{C_\sigma}
             {2(\min\{V_i,V_{i^*}\}-\Delta_i)}.
$$
Consequently,
\[
    1\le\frac{F}{L}\le\rho(\bmu,\A)
    :=C_\sigma\max_{i\ne i^*}
        \frac{\ln(1/p_i)}{\Delta_i^2}.
\]
\end{theorem}

This gives an explicit finite approximation factor in terms of the arm
gaps and context probabilities. Although this bound is coarse near a
tie, the next result shows that $L$ approaches $F$ in a small-gap
regime. Define
$
 \Gamma=\max_{i\ne i^*}
     \frac{\Delta_i}{\min\{w_i,w_{i^*}\}H_i},
$
with $H_i$ evaluated at $\qb=\A\wb$.

\begin{theorem}\label{thm: hard relaxation}
Let $(\bmu,\A)\in\I$ have unit-variance Gaussian rewards, and let
$\wb\in\Delta_+^{n-1}$ with $\qb=\A\wb$.
If $\Gamma\le1/20$ and $\bmu\in[1.3\Gamma,1-1.3\Gamma]^k$, then
$
 L\le F\le(1+12\Gamma)L.
$
In particular, $F/L\to1$ as $\Gamma\to0$ under these conditions.
\end{theorem}

The bound becomes nearly exact when the arms have similar expected
rewards, provided that the context variances remain bounded below and
the context reward means stay away from $0$ and $1$. As the arm gaps
shrink, identifying the best arm becomes harder, but $\Gamma$ tends to
zero. The theorem then guarantees that $F/L\to1$: the computable lower
bound captures essentially all of the information in $F$, precisely
in this regime where distinguishing the arms requires more samples.

\textbf{Using the relaxation.}
Define the relaxed complexity by
\begin{equation}\label{eq: relaxation}
 U^*(\bmu,\A)^{-1}
   =\max_{\wb\in\Delta^{n-1}}L(\bmu,\A,\wb,\A\wb),
\end{equation}
where the objective is extended continuously by zero on the simplex
boundary. By Theorem~\ref{thm: simple lower bound}, this is a convex
optimization problem. In Algorithm~\ref{algo: unknown}, we replace the
allocation oracle by an empirical maximizer of this objective and stop
when $\lamh^L(t)>\bettdel$. The G-tracking rule and threshold remain
unchanged.

\begin{theorem}\label{thm : relaxation upper}
Let $(\bmu,\A)\in\I$. Algorithm~\ref{algo: unknown} with the relaxed
stopping statistic and access to an exact optimizer of
\eqref{eq: relaxation} is $\delta$-correct and satisfies
\begin{align*}
 \limsup_{\delta\to0}
 \frac{\mathbb E_{\bmu,\A}[\tau_\delta]}{\logdel}
     &\le U^*(\bmu,\A) \le\rho(\bmu,\A)T^*(\bmu,\A).
\end{align*}
If Theorem~\ref{thm: hard relaxation} applies at a $T^*$-optimal allocation with relative gap $\Gamma^*$, then $U^*\le(1+12\Gamma^*)T^*$.
\end{theorem}


\section{Experiments}\label{sec: exps}

In this section, we present our experimental results. In all experiments, rewards follow Gaussian distributions with unit variance.\footnote{Code: \url{https://github.com/ban-epfl/BAI-Unknown-Post-Action-Context}}

\begin{figure}[t]
    \centering
    \includegraphics[width=0.45\textwidth]{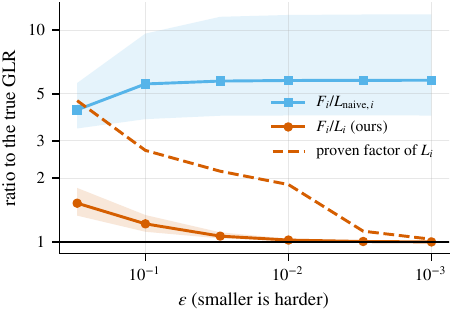}
    \caption{Ratio of $F_i$ to its lower bounds on hard instances (median and
    interquartile range). Dashed: the factor guaranteed by
    Theorems~\ref{thm: simple lower bound} and~\ref{thm: hard relaxation}.}
    \label{fig: tightness}
\end{figure}


\textbf{Tightness of the convex relaxation.} In this experiment, we compare the lower bound $L_i$ in \eqref{eq: convex-lb}
with the numerically computed value of $F_i$
(Appendix~\ref{apd: exp reference}) and the naive lower bound
$L_{\mathrm{naive}, i}$ obtained by adapting the relaxation of
\citet{al2021adaptive} (Lemma~\ref{lem: convexify-rl} in
Appendix~\ref{apd: convex-relax-rl}). We consider a family of hard instances parameterized by $\epsilon$, where smaller values of $\epsilon$ correspond to smaller gaps and
thus harder instances (see Appendix~\ref{apd: exp tightness} for details).

Figure~\ref{fig: tightness} shows the ratios $F_i/L_i$ and
$F_i/L_{\mathrm{naive}, i}$ as functions of $\epsilon$.
As the instances become harder, $F_i/L_i$ approaches $1$, consistent with Theorem~\ref{thm: hard relaxation}, whereas the naive bound remains a constant factor away from $F_i$.
The improvement extends beyond this family: on random instances, $L_i$ is typically within a factor of two of $F_i$ and provides a tighter approximation than $L_{\mathrm{naive}, i}$, with substantially less variation in the approximation ratio (Table~\ref{table: tightness}).

\textbf{Illustration of the gap between the GLR statistic and its relaxations.}
In this experiment, we compare our convex relaxation and the naive relaxation
of the GLR statistic, as well as our proposed thresholds $\bettdel$ against
the naive design $\beta_1(t,\delta/2)+\beta_2(t,\delta/2)$
(Appendix~\ref{apd: sub-optimal threshold}).
Experiments are run on a small instance with Gaussian reward distributions
and two values of $\delta$ (details in Appendix~\ref{apd: experiments}), where
the value of the GLR statistic is computed numerically.

\begin{figure}[t]
    \centering
    \includegraphics[width=0.95\textwidth]{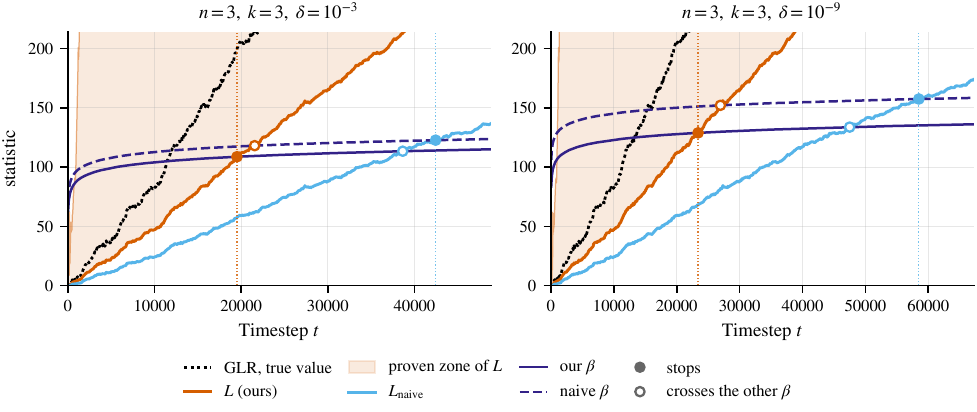}
    \caption{GLR statistic, our relaxation $L$, naive relaxation $L_{\mathrm{naive}}$,
and stopping thresholds for $\delta=10^{-3}$ (left) and $\delta=10^{-9}$
(right). Shading indicates the GLR range guaranteed by
Theorem~\ref{thm: gap}.}
    \label{fig: glr trace}
\end{figure}

Figure~\ref{fig: glr trace} reports how the GLR statistic evolves with time
step $t$ for our convex relaxation, the naive relaxation, and the numerically
computed value, as well as two threshold designs.
The results show that the naive relaxation underestimates GLR values and
delays stopping, while our relaxation stays within a factor of about two of
the GLR statistic and stops after less than half of the rounds needed by the
naive design.
Moreover, our thresholds are tighter than the naive alternatives, with the
gap increasing as $\delta$ decreases (from the left to the right panel),
leading to faster identification.

\textbf{Sample complexity comparison.}
In this experiment, we compare the impact of different levels of prior
knowledge of the environment on sample complexity. We evaluate the
following algorithms.

\textbf{GTS-$\bmu$.} Assumes prior knowledge of the vector $\bmu$ and runs
Algorithm~\ref{algo: unknown} for known $\bmu$, which uses Track-and-Stop
with the G-tracking rule.

\textbf{GTS-$\A$.} Assumes knowledge of the context distribution matrix $\A$
and runs STS algorithm from \citet{shahverdikondori2025optimal} which
performs G-tracking for the given $\A$.

\textbf{GTS-Unknown (ours).} Algorithm~\ref{algo: unknown}, which operates
without any prior knowledge of the environment and uses the convex relaxation
of Subsection~\ref{sec: relaxation} for both the stopping statistic and the
allocation. For $k = 2$, it solves the optimization in
\eqref{eq: unknown f formula} exactly
(Proposition~\ref{prop: k2 decomposition}).

\textbf{GTS-Unknown (Naive).} A baseline that also operates without any prior
knowledge of the environment, but uses the naive relaxation
$L_{\mathrm{naive}}$ and the naive threshold design, both obtained by
following \citet{al2021adaptive}.

\textbf{GTS-SG.} A baseline that ignores post-action context, treating each
arm's reward distribution as sub-Gaussian (valid since means lie in the
bounded interval $(0,1)$), and applies stopping thresholds for sub-Gaussian
random variables~\citep{tuynman2025batch}. Details are provided in
Appendix~\ref{apd: experiments}.

We conducted an experiment to illustrate how prior knowledge of either
$\bmu$ or $\A$ accelerated the learning process. We considered two instances
with $n = 3$, $k = 2$, $\delta = 0.001$, and parameters
\begin{align} \label{eq: instance-hard}
     &\bmu_1 = \begin{bmatrix}
        0.9 \\
        0.1
    \end{bmatrix}, \quad
    \A_1 = \begin{bmatrix}
        0.6 & 0.5 & 0.4 \\
        0.4 & 0.5 & 0.6
    \end{bmatrix}, \quad
    \bmu_2 = \begin{bmatrix}
        0.4 \\
        0.6
    \end{bmatrix}, \quad
    \A_2 = \begin{bmatrix}
        0.9 & 0.5 & 0.1 \\
        0.1 & 0.5 & 0.9
    \end{bmatrix}. 
\end{align} 
In instance $(\bmu_1, \A_1)$, the context distributions of the arms are
close, so identifying the best arm requires more accurate estimates of
$\A_1$, even if $\bmu_1$ is known. Hence, prior knowledge of $\A$ is
particularly beneficial. Conversely, in instance $(\bmu_2, \A_2)$, the
reward means are close, and accurate knowledge of $\bmu_2$ is more important.

\begin{figure}[!t] 
    \centering
    \includegraphics[width=0.85\textwidth]{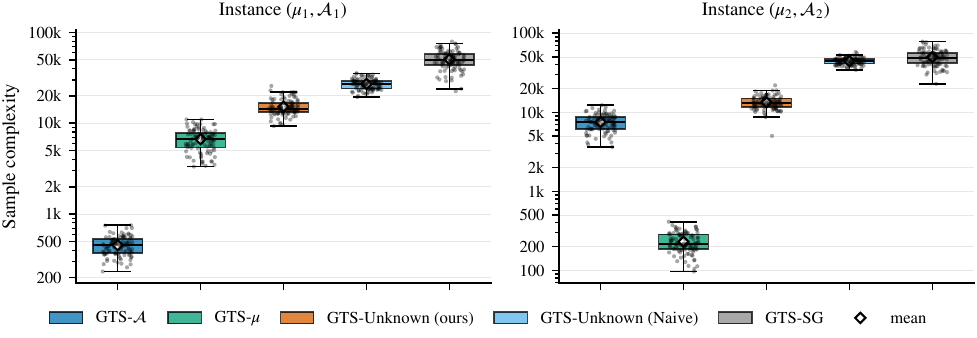}
    \caption{Sample complexity comparison of different algorithms on the
    instances defined in \eqref{eq: instance-hard}.} \label{fig: sample-comp-main}
\end{figure} 

We ran all five algorithms on both instances and report the stopping times in
Figure~\ref{fig: sample-comp-main}. As expected, GTS-$\A$ is more efficient on instance~1 and GTS-$\bmu$ on instance~2. All context-aware algorithms,
including those without prior knowledge, outperform the baseline that ignores
post-action context. Among the algorithms without prior knowledge, GTS-Unknown
(ours) needs roughly two to three times fewer samples than GTS-Unknown (Naive).
Additional experiments in Appendix~\ref{apd: experiments} show similar trends.

\section{Conclusion}
We studied fixed-confidence best-arm identification with unknown
post-action context distributions. We derived an instance-dependent
sample complexity lower bound and proposed a Track-and-Stop algorithm that matches it asymptotically with an optimization oracle. Our joint stopping threshold removes the factor of two incurred by separate concentration bounds. To address the computational challenges, we identified tractable cases and proposed a convex relaxation with instance-dependent approximation guarantees and asymptotic tightness for a class of hard Gaussian instances. Experiments demonstrate that post-action context can improve sample efficiency even when its distribution is unknown.

\bibliography{biblio}

\newpage

\startcontents[appendix]
\thispagestyle{empty}
\begin{center} \label{sec: apd}
    {\Large \textbf{Supplementary Materials}}
\end{center}

\vspace{1em}
\noindent{\large\bfseries Contents}\par
\smallskip\hrule\medskip
\printcontents[appendix]{apd}{1}[2]{}

\begin{table}[H]
    \centering
    \caption{Table of notations.}
    \label{table: notations}
    \small
    \begin{tabular}{c|l} 
        \toprule
        Notation & Description \\
        \hline
        $n$ & Size of action set \\
        $k$ & Size of post-action context \\
        $\Delta^{n-1} (\Delta^{n-1}_+)$ & $(n-1)$-dimensional standard simplex (with strictly positive components)\\
        $d(\mu, \lambda)$ & Kullback–Leibler (KL) divergence between two exponential family with mean $\mu$ and $\lambda$ \\
        $d(\textbf{p}, \textbf{q})$ & The KL divergence between the categorical distributions induced by $\textbf{p}$ and $\textbf{q}$ \\     
        $d_{B}(p, 1-p)$ & KL between two Bernoulli  with parameters $p$ and $1 - p$ \\
        $\A = [\A_1 \vert \A_2 \vert \cdots \vert \A_n]$ & Context probability matrix, where $\A_{j,i} = \pr(Z = j \mid X = i)$ \\
        $\bmu = [\mu_1, \mu_2, \ldots, \mu_k]$ & the means value vector, where $\mu_{i} = \mathbb{E}[Y \mid Z=i]$ \\
        $P^{\A, \bmu}_{i}$ & Joint distribution of $(Z, Y)$ after pulling arm $X = i$ in environment $\A, \bmu$ \\
        $i^*(\bmu, \A), i^*$ & The best arm for an instance parameterized by $\bmu$ and $\A$ \\
        $\Delta_i$  &   Sub-optimality gap of arm $i$ \\
        
        $\muht, \Aht$ &       The empirical estimates of all entries of $\bmu$ and $\A$ up to round $t$ \\
        $i_{t}^*$ & The empirical best arm for an instance parameterized by $\muht$ and $\Aht$ \\
        $\delhit{i}$ &  Estimated sub-optimality gap of arm $i$ and $i_{t}^*$ in around $t$ \\
        Alt$(\bmu, \A)$ & Set of alternative parameter $(\bmu', \A') \in \I$ such that $i^*(\bmu, \A) \neq i^*(\bmu', \A')$ \\
        $\numaction$ &    Total number of times arm $i$ has been played up to round $t$ \\
        $\numcontext$ &    Total number of times context $j$ has been observed up to round $t$ \\
        $\numac$ &   Total number of times arm-context $i, j$ has been observed up to round $t$ \\
        
        $\lamh_{\bmu}(t), \lamht$ &      GLR statistic at around $t$ for the known and unknown case\\
        $\wb^*_{\bmu}(\A) ,\mathbf{w}^*(\bmu, \A)$ &  The optimal weights (proportions) for actions, the known and unknown case \\
        
        \bottomrule
    \end{tabular}
\end{table}

\section{Related Work}\label{apd: related work}

There has been extensive research on pure exploration problems in multi-armed bandits~\citep{bandit-book1-lattimore2020bandit, pure2-thesis-stephens2023pure, shahverdikondori2026active, lb-tsitsiklis-mannor2004sample, shahverdikondori2025best}. The classical best-arm identification (BAI) problem has been primarily studied under two paradigms: the \emph{fixed-confidence} setting, where the learner aims to identify the optimal arm with an error probability of at most $\delta$~\citep{track-stop-garivier2016optimal, SR-audibert2010best, kaufmann2020contributions, confidence-jamieson2014best, lb-tsitsiklis-mannor2004sample}, and the \emph{fixed-budget} setting, where the time horizon is fixed and the goal is to minimize the probability of error~\citep{budget-confidence-gabillon2012best, SH-karnin2013almost, lb-budget-carpentier2016tight, pmlr-v306-balagopalan26a, hwang2026box}.  

The first asymptotically optimal algorithm for BAI in the fixed-confidence regime, \emph{Track-and-Stop}, was introduced by~\citet{track-stop-garivier2016optimal}. This method tracks the optimal sampling proportions of arms and stops once sufficient statistical evidence is accumulated. Since then, the Track-and-Stop framework has served as the foundation for several extensions addressing different problem variants \citep{kaufmann2021mixture, linear-lazy-jedra2020optimal, mutiple-correct-answers-degenne2019pure}, including the one studied in this work.  

Within the fixed-confidence BAI literature, most studies assume that the reward distributions belong to a one-parameter exponential family, most commonly, Gaussian distributions with known variance. \citet{tuynman2025batch} extended this framework to the broader class of sub-Gaussian distributions, showing that, with properly tuned sequential thresholds, similar asymptotic guarantees can be retained. We employ these thresholds in the GTS-SG algorithm used in our experiments; explicit expressions are provided in Appendix~\ref{apd: experiments}.  

Another closely related research direction involves \textit{causal
bandits}, where the learner interacts with a causal model and can
perform interventions on certain variables~\citep{lattimore2016causal,
causal-context-madhavan2024causal, pmlr-v300-shahverdikondori26a, causal-fateme-jamshidi2024confounded,
causal-fairness-huang2022achieving, causal-pure-xiong2022combinatorial}.
In these problems, the feedback includes not only rewards but also
observations of other nodes in the causal graph, distinguishing them
from classical contextual bandits. Our framework can be seen as a
specific instance of causal bandits with three nodes, as depicted
in Figure~\ref{fig: causal_graph}, where actions correspond to
interventions on $X$.

Most work in this area targets cumulative regret
\citep{causal-cucb-lu2020regret} or simple regret
\citep{lattimore2016causal, causal-simple-sawarni2023learning}.
Existing pure-exploration formulations also differ from ours:
some impose assumptions such as binary generalized linear
models~\citep{causal-pure-xiong2022combinatorial}, while others
consider soft interventions and the fixed-budget
setting~\citep{causal-fixed-budget-sen2017identifying}.

Two other works study identification with a context that does not
depend on the action: \citet{context2-kato2021role} consider BAI
with a context observed before the action, and \citet{russac2021b}
consider A/B/n testing against a control arm in the presence of
subpopulations, where the context may also be observed after the action.

\section{Causal Interpretation and Latent Confounding}
\label{apd: causal interpretation}

\begin{figure}[htbp]
 \centering
 \begin{tikzpicture}[
  observed/.style={circle,draw=black,fill=white,line width=0.6pt,
                   minimum size=7mm,inner sep=2pt,outer sep=0pt},
  latent/.style={observed,dash pattern=on 3pt off 2pt},
  causal/.style={->,>={Stealth[black]},draw=black,line width=0.8pt},
  parameter/.style={midway,above=2pt,inner sep=1pt}
 ]
  \node[observed] (X) at (0,0) {$X$};
  \node[observed] (Z) at (1.7,0) {$Z$};
  \node[observed] (Y) at (3.4,0) {$Y$};
  \node[latent] (U) at (1.7,1.4) {$U$};
  \draw[causal] (X) -- node[parameter] {$\A$} (Z);
  \draw[causal] (Z) -- node[parameter] {$\bmu$} (Y);
  \draw[causal] (U) -- (X);
  \draw[causal] (U) -- (Y);
 \end{tikzpicture}
 \caption{Unobserved confounding between $X$ and $Y$. Intervening on
 $X$ removes the edge $U\to X$.}
 \label{fig: causal confounding}
\end{figure}
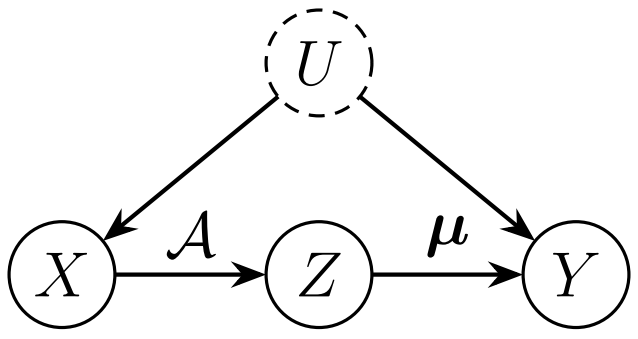

In the causal bandit formulation, each arm $i$ corresponds to an
intervention $\operatorname{do}(X=i)$, after which the learner observes
$(Z,Y)$. Our framework also covers unobserved confounding between
$X$ and $Y$, as illustrated in Figure~\ref{fig: causal confounding}.

A hard intervention removes the edge $U\to X$, so the reward is
conditionally independent of the intervention given $Z$:
\[
 \pr(Y\mid Z=j,\operatorname{do}(X=i))
 =\pr(Y\mid Z=j,\operatorname{do}(X=i')),
 \qquad \forall i,i'.
\]
This is the conditionally benign property of
\citet{causal-benign-bilodeau2022adaptively}. Thus, although the
observational distribution can be confounded, the interventional
feedback satisfies our model. Define
\[
 \A_{j,i}=\pr(Z=j\mid\operatorname{do}(X=i)),
 \qquad
 \mu_j=\mathbb{E}[Y\mid Z=j,\operatorname{do}(X=i)].
\]
Then $\mathbb{E}[Y\mid\operatorname{do}(X=i)]=\A_i^\top\bmu$, so
identifying the best intervention is exactly our best-arm problem.
Under the standing distributional assumptions, our algorithm applies
without modification and retains its guarantees. Neither $\A$ nor
$\bmu$ needs to be known: both can be estimated from intervention
feedback.

\section{Lower Bound Derivation} \label{apd: lower bound}
This section derives the lower bounds stated in Proposition \ref{pro: lower}. First, we review the following proposition from \citet{shahverdikondori2025optimal}.

\begin{proposition}[\citet{shahverdikondori2025optimal}] \label{pro: general lower} 
     For any bandit environment with parameters $(\bmu,\A) \in \I$ and any $\delta$-correct algorithm, we have $\mathbb{E}_{\bmu, \A}[\tau_{\delta}] \geq T^*(\bmu, \A)d_B(\delta, 1- \delta)$, where
        \begin{align} \label{eq: general_lower}
            T^*(\bmu, \A)^{-1} \triangleq \hspace{-0.1cm} \sup_{\mathbf{w} \in \Delta^{n-1}} \inf_{\substack{(\bmu',\A') \\ \in \text{Alt}(\bmu, \A)}} \sum_{i \in [n]} w_i \mathrm{KL} \paran{P_{\bmu, \A}^{i}, P_{\bmu', \A'}^{i}}.
        \end{align}
\end{proposition}

\textbf{Known Reward Means.}  
Fix an instance $(\bmu, \A) \in \I$, where $\bmu$ is known to the learner. From Proposition \ref{pro: general lower}, for any $\delta$-correct algorithm we have
\[
\mathbb{E}_{\bmu, \A}[\tau_\delta] \;\ge\; T^*(\bmu, \A)\, d_B(\delta, 1 - \delta),
\]
where
\[
T^*(\bmu, \A)^{-1} \;=\;
\sup_{\wb \in \Delta^{n-1}}
\ \inf_{(\bmu', \A') \in \text{Alt}(\bmu, \A)}
\ \sum_{i \in [n]} w_i\, \mathrm{KL}\!\left( P_{\bmu, \A}^{i},\, P_{\bmu', \A'}^{i} \right).
\]
Since $\bmu$ is known a priori, the hardest alternative instances can be chosen with $\bmu' = \bmu$. Hence,
\[
T^*_{\bmu}(\A)^{-1} \;=\;
\sup_{\wb \in \Delta^{n-1}}
\ \inf_{\B \in \text{Alt}_{\bmu}(\A)}
\ \sum_{i \in [n]} w_i\, \mathrm{KL}\!\left( P_{\bmu, \A}^{i},\, P_{\bmu, \B}^{i} \right),
\]
where $\text{Alt}_{\bmu}(\A) \triangleq \{\B : i^*(\bmu, \B) \neq i^*(\bmu, \A)\}$.

By the properties of the KL divergence, we have
\begin{align*}
    \mathrm{KL}\!\left( P_{\bmu, \A}^{i},\, P_{\bmu, \B}^{i} \right)
    &= \mathrm{KL}\!\left( \pr_{\bmu, \A}(Y, Z \mid X=i),\,
    \pr_{\bmu, \B}(Y, Z \mid X=i) \right) \\[4pt]
    &= d(\A_i, \B_i) +
    \sum_{j \in [k]} \A_{j,i}\,
    d \big( \pr_{\bmu, \A}(Y \mid Z=j, X=i),\,
    \pr_{\bmu, \B}(Y \mid Z=j, X=i) \big) \\
    &= d(\A_i, \B_i),
\end{align*}
where the last equality follows because $\pr_{\bmu, \A}(Y \mid Z=j, X=i) = \pr_{\bmu, \B}(Y \mid Z=j, X=i)$, as the vector $\bmu$ is shared between the two instances.

Next, decompose $\text{Alt}_{\bmu}(\A)$ as
$\text{Alt}_{\bmu}(\A) = \bigcup_{i \neq \istar} C_i$,  
where $C_i = \{ \B : \B_i^{\top}\bmu > \B_{\istar}^{\top}\bmu \}$. Then,
\[
T^*_{\bmu}(\A)^{-1}
\;=\;
\sup_{\wb \in \Delta^{n-1}}
\min_{i \neq \istar}
\inf_{\B \in C_i}
\sum_{s \in [n]} w_s\, d(\A_s, \B_s).
\]
To minimize the inner infimum, it is sufficient to modify only the two columns of $\A$ with indices $i$ and $\istar$, since changing other columns only increases the value. Thus,
\[
T^*_{\bmu}(\A)^{-1}
\;=\;
\sup_{\wb \in \Delta^{n-1}}
\min_{i \neq \istar}
\inf_{\substack{\B_i,\, \B_{\istar} :\\ \B_i^{\top}\bmu > \B_{\istar}^{\top}\bmu}}
\Big[
    w_{\istar}\, d(\A_{\istar}, \B_{\istar})
    + w_i\, d(\A_i, \B_i)
\Big],
\]
which concludes the derivation for the known-$\bmu$ case.  

\textbf{Unknown Environment.}  
When both $\A$ and $\bmu$ are unknown, the derivation follows directly from the general bound. For any two instances $(\bmu, \A)$ and $(\lambb, \B)$, by properties of the KL divergence we have
\begin{align*}
    \mathrm{KL}\!\left( P_{\bmu, \A}^{i},\, P_{\lambb, \B}^{i} \right)
    &= d(\A_i, \B_i)
    + \sum_{j \in [k]} \A_{j,i}\,
    d \big( \pr_{\bmu, \A}(Y \mid Z=j, X=i),\,
    \pr_{\lambb, \B}(Y \mid Z=j, X=i) \big) \\[4pt]
    &= d(\A_i, \B_i)
    + \sum_{j \in [k]} \A_{j,i}\, d(\mu_j, \lambda_j).
\end{align*}
Hence,
\begin{align*}
    T^*(\bmu, \A)^{-1}
    &= \sup_{\wb \in \Delta^{n-1}}
    \inf_{\substack{(\lambb, \B) \\ \in \text{Alt}(\bmu, \A)}}
    \sum_{i \in [n]} w_i\,
    \mathrm{KL}\!\left( P_{\bmu, \A}^{i},\, P_{\lambb, \B}^{i} \right) \\[3pt]
    &= \sup_{\wb \in \Delta^{n-1}}
    \inf_{\substack{(\lambb, \B) \\ \in \text{Alt}(\bmu, \A)}}
    \left[
        \sum_{i \in [n]} w_i\, d(\A_i, \B_i)
        + \sum_{j \in [k]}
        \Big( \sum_{i \in [n]} w_i\, \A_{j,i} \Big)
        d(\mu_j, \lambda_j)
    \right],
\end{align*}
which completes the derivation for the fully unknown case.

\section{Stopping Rule} \label{apd: glr derivation}
This section contains the proofs for GLR and stopping rules.

\subsection{GLR Derivation} \label{}
The derivation of our GLR is similar to prior work \citep{shahverdikondori2025optimal, kaufmann2021mixture}. We define two hypotheses test between $H_0$ and $H_1$, where 
\begin{align*}
              &\hcal_0: (i^*_t \neq i^*(\bmu, \A)),   \\
              &\hcal_1: (i^*_t = i^*(\bmu, \A)).
\end{align*}
In each round, we run this test, and if we have high confidence to reject $\hcal_0$, then it means that $i^*_t = i^*(\bmu, \A)$. Therefore, we stop and output $i_t^*$. 

Now, we need to find the log-likelihood ratio of the above test. Note that each instance of our problem is parameterized by $(\bmu, \A)$, and $i^*_t \neq i^*(\bmu, \A)$ is equivalent to $(\bmu, \A) \in \text{Alt}(\muht, \Aht)$. Therefore, the log-likelihood ratio is 
$$
    \ln(R_t) = \ln \frac{\sup_{(\lambda, \B) \in I} \lcal((Z_1, Y_1), \dots, (Z_t, Y_t); \lambda, \B)}{\sup_{(\lambda, \B) \in \text{Alt}(\muht, \Aht)} \lcal((Z_1, Y_1), \dots, (Z_t, Y_t); \lambda, \B)}.
$$
The maximizer of the numerator is the maximum likelihood, i.e., $\lambda, \B = \muht, \Aht$. Therefore, 
$$
 \ln(R_t) = \ln \frac{\lcal((Z_1, Y_1), \dots, (Z_t, Y_t); \muht, \Aht)}{\sup_{(\lambda, \B) \in \text{Alt}(\muht, \Aht)} \lcal((Z_1, Y_1), \dots, (Z_t, Y_t); \lambda, \B)}.
$$
And then, we have
\begin{align*}
          \ln(R_t) &= \ln \frac{\prod_{s \in [t]} \pr_{\Aht}(Z = z_s \mid X = i_s) \pr_{\muht} (Y = y_s \mid Z = z_s)}{\sup_{\lambda, \B \in \text{Alt}(\muht, \Aht)} \prod_{s \in [t]} \pr_{\B}(Z = z_s \mid X = i_s) \pr_{\lambda} (Y = y_s \mid Z = z_s)} \\
                &= \ln \inf_{\lambda, \B \in \text{Alt}(\muht, \Aht)} \prod_{s \in [t]} \frac{\pr_{\Aht}(Z = z_s \mid X = i_s)}{\pr_{\B}(Z = z_s \mid X = i_s)}
                \frac{\pr_{\muht} (Y = y_s \mid Z = z_s)}{\pr_{\lambda} (Y = y_s \mid Z = z_s)}     \\  
                &= \inf_{\lambda, \B \in \text{Alt}(\muht, \Aht)} \sum_{s \in [t]} \ln \left ( \frac{\pr_{\Aht}(Z = z_s \mid X = i_s)}{\pr_{\B}(Z = z_s \mid X = i_s)} \right) + \sum_{s \in [t]} \ln \left (  \frac{\pr_{\muht} (Y = y_s \mid Z = z_s)}{\pr_{\lambda} (Y = y_s \mid Z = z_s)} \right ) \\
                &= \inf_{\lambda, \B \in \text{Alt}(\muht, \Aht)} \sum_{i \in [n]} N^{X}_i(t) d(\Ah_i(t), \B_i) + \sum_{s \in [t]} \ln \left (  \frac{\pr_{\muht} (Y = y_s \mid Z = z_s)}{\pr_{\lambda} (Y = y_s \mid Z = z_s)} \right )
\end{align*}
Since the reward distribution belongs to one-parameter exponential family, we simplify the second term, hence,
\begin{equation} \label{eq: log-like}
 \ln(R_t) = \inf_{\lambda, \B \in \text{Alt}(\muht, \Aht)} \sum_{i \in [n]} N_i^X(t) d(\Ah_i(t), \B_i) + \sum_{j \in [k]} N^Z_j(t) d(\muh_j(t), \lambda_j).
\end{equation}

For the unknown case, the value of GLR is exactly equal to equation \eqref{eq: log-like}; therefore, we have
$$
    \lamht =   \ln(R_t)  = \inf_{\lambda, \B \in \text{Alt}(\muht, \Aht)} \sum_{i \in [n]} N_i^X(t) d(\Ah_i(t), \B_i) + \sum_{j \in [k]} N^Z_j(t) d(\muh_j(t), \lambda_j).
$$

For the case of known $\mu$, we do not require involving the reward part in the computation of log-likelihood (because $\bmu$ is not further an unknown parameter in this case). Therefore, the log-likelihood is equal to
$$
      \ln(R_t) = \ln \frac{\sup_{(\bmu, \B) \in I} \lcal((Z_1, Y_1), \dots, (Z_t, Y_t); \bmu, \B)}{\sup_{(\bmu, \B) \in \text{Alt}(\bmu, \Aht)} \lcal((Z_1, Y_1), \dots, (Z_t, Y_t); \bmu, \B)},
$$
Which, after similar computation, will be equal to 
$$
    \ln(R_t) = \inf_{\bmu, \B \in \text{Alt}(\bmu, \Aht)} \sum_{i \in [n]} N_i^X(t) d(\Ah_i(t), \B_i).
$$
Therefore, $\lamh_{\bmu}(t) = \inf_{\bmu, \B \in \text{Alt}(\bmu, \Aht)} \sum_{i \in [n]} N_i^X(t) d(\Ah_i(t), \B_i)$. We can rewrite this optimization problem as 
\begin{align} \label{eq: glr-known-proof}
    \lamh_{\bmu}(t) &= \inf_{\bmu, \B \in \text{Alt}(\bmu, \Aht)} \sum_{i \in [n]} N_i^X(t) d(\Ah_i(t), \B_i) \nonumber \\
                    &= \min_{i \neq i_t^*} \inf_{\B \in \mathcal{C}_i} \sum_{i \in [n]} N_i^X(t) d(\Ah_i(t), \B_i) \nonumber \\
                    &= \min_{i \neq i_t^*} \inf_{\substack{\B_1, \B_2 : \\ \B_1^{\top} \bmu < \B_2^{\top} \bmu}} N^X_{\istarat} (t) d \paran{\Ah_{\istarat}(t), \B_1} + N^X_i(t) d \paran{\Ah_i(t), \B_2},
\end{align}
where $\mathcal{C}_i \triangleq \{\B: \B_i^\top \bmu > \B_{i_t^*}^\top \bmu \}$. This finishes the proof of computation of GLR for known $\bmu$ and unknown case.

\subsection{Proof of $\delta$-correctness (Lemma \ref{lem: stopping})}

We prove the $\delta$-correctness for known and unknown $\bmu$ separately. 

\subsubsection{Proof of $\delta$-correctness of known $\bmu$}

First, consider the following lemma.
\begin{lemma}[\citet{rl2-al2021navigating, jonsson2020planning}] \label{lem: concen-cat}
    Let $\delta \in (0, 1)$ and $\beta_{\bmu}(t, \delta) = \logdel + (k-1) \sum_{i \in [n]} \ln \paran{e \paran{1 + \frac{N^X_i(t)}{k-1}}}$, we have

    $$
        \pr \left [ \exists t \geq 1: \sum_{i \in [n]} N^X_i(t) d(\Ah_i(t), \A_i) \geq \beta_{\bmu}(t, \delta) \right ] \leq \delta
    $$
\end{lemma}
\begin{proof}
    A similar inequality was established in \citet{rl2-al2021navigating}, based on a deviation inequality for KL-divergences.
    Here, we briefly outline the main idea.
    \citet{rl2-al2021navigating} showed that there exists a non-negative martingale $(N_t)_{t \geq 0}$ satisfying:
    \begin{itemize}
        \item $N_0 = 1,$
        \item $\forall t \geq 1: N_t \geq \exp \left ( \sum_{i \in [n]} N^X_i(t) d(\Ah_i(t), \A_i)  - (k-1) \sum_{i \in [n]} \ln \paran{e \paran{1 + \frac{N^X_i(t)}{k-1}}}\right ).$
    \end{itemize}

    Since $N_t$ is a non-negative martingale, by applying Ville's (Doob maximal) inequality, we have 
    $$
        \pr [\exists t \geq 0: N_t \geq \frac{1}{\delta}] \leq \delta \mathbb{E}[N_0] = \delta
    $$
    Using the second property of $N_t$, we obtain
    \begin{align*}
         \pr [\exists t \geq 0 &: N_t \geq \frac{1}{\delta}] = \pr [\exists t \geq 1: N_t \geq \frac{1}{\delta}] = \pr [\exists t \geq 1: \ln(N_t) \geq \logdel] \\
                                               & \geq \pr [\exists t \geq 1: \sum_{i \in [n]} N^X_i(t) d(\Ah_i(t), \A_i)  - (k-1) \sum_{i \in [n]} \ln \paran{e \paran{1 + \frac{N^X_i(t)}{k-1}}} \geq \logdel] \\
                                                & = \pr [\exists t \geq 1: \sum_{i \in [n]} N^X_i(t) d(\Ah_i(t), \A_i)  \geq \beta_{\bmu}(t, \delta)],
    \end{align*}        
    which concludes the proof.
\end{proof}

\begin{restatable}{lemma}{knownStopping} 
\label{lem: known stopping}
    Let $(\bmu, \A) \in \I$ be a bandit instance with post-action context. Any algorithm that has access to $\bmu$ and employs the stopping rule $\beta_{\bmu}(t, \delta)$ is $\delta$-correct, i.e.,
    $
    \pr_{\bmu, \A} \big(\tau_{\delta} < \infty, \ \hat{i}_{\tau} \neq \istar \big) \leq \delta .
    $
\end{restatable}

\begin{proof}
We have
\begin{align*}
     \pr_{\bmu, \A} &\big( \tau_{\delta} < \infty,  \hat{i}_{\tau} \neq \istar \big) = \pr \left [\exists t \geq 1: \lamh_{\bmu}(t) >  \beta_{\bmu}(t, \delta), i^* \neq i_t^* \right] \\
                      &=  \pr \left [\exists t \geq 1: \min_{i \neq i_t^*} \inf_{\substack{\B_1, \B_2 : \\ \B_1^{\top} \bmu < \B_2^{\top} \bmu}} N^X_{\istarat} (t) d \paran{\Ah_{\istarat}(t), \B_1} + N^X_i(t) d \paran{\Ah_i(t), \B_2} >  \beta_{\bmu}(t, \delta), i^* \neq i_t^* \right] \\
                      &= \pr \left [\exists t \geq 1: \inf_{\bmu, \B \in \text{Alt}(\bmu, \Aht)} \sum_{i \in [n]} N_i^X(t) d(\Ah_i(t), \B_i) >  \beta_{\bmu}(t, \delta), i^* \neq i_t^* \right],
\end{align*}
where we use the definition of $\lamh_{\bmu}(t)$ from Equation \eqref{eq: glr-known-proof}. On event $(i^* \neq i_t^*)$, we have $(\bmu, \A) \in  \text{Alt}(\bmu, \Aht)$, and hence,
\begin{align*}
    &\pr \left [\exists t \geq 1: \inf_{\bmu, \B \in \text{Alt}(\bmu, \Aht)} \sum_{i \in [n]} N_i^X(t) d(\Ah_i(t), \B_i) >  \beta_{\bmu}(t, \delta), i^* \neq i_t^* \right] \\
       &\leq \pr \left [\exists t \geq 1: \sum_{i \in [n]} N_i^X(t) d(\Ah_i(t), \A_i) >  \beta_{\bmu}(t, \delta)\right], \\            
\end{align*}
which is less than $\delta$ by Lemma \ref{lem: concen-cat}. Finally, we conclude that
$$
     \pr_{\bmu, \A} \left (\tau_{\delta} < \infty, \hat{i}_{\tau} \neq \istar \right ) \leq \pr \left [\exists t \geq 1: \sum_{i \in [n]} N_i^X(t) d(\Ah_i(t), \A_i) >  \beta_{\bmu}(t, \delta)\right] \leq \delta.
$$

\end{proof}

\subsubsection{Proof of $\delta$-correctness of unknown $\bmu$}
\begin{restatable}{lemma}{unknownStopping}
\label{lem: unknown stopping}
    Let $(\bmu,\A)\in\I$ be a bandit instance with
    post-action context. Any algorithm that employs
    the stopping rule~\eqref{eq: stop rule} with
    threshold $\beta(t,\delta)$ is $\delta$-correct, i.e.,
    $
    \pr_{\bmu,\A}\big(
        \tau_\delta<\infty,\,
        \hat{i}_{\tau}\neq\istar
    \big)\leq\delta.
    $
\end{restatable}
\begin{proof}
Similar to proof of Lemma \ref{lem: known stopping}, we have 
\begin{align*}
  &\pr_{\bmu, \A} \big(\tau_{\delta} < \infty,  \hat{i}_{\tau} \neq \istar \big) = \pr \left [\exists t \geq 1: \lamht >  \beta(t, \delta), i^* \neq i_t^* \right] \\
                    &= \pr \left [\exists t \geq 1:  \inf_{\lambda, \B \in \text{Alt}(\muht, \Aht)} \sum_{i \in [n]} N_i^X(t) d(\Ah_i(t), \B_i) + \sum_{j \in [k]} N^Z_j(t) d(\muh_j(t), \lambda_j) >  \beta(t, \delta), i^* \neq i_t^* \right] \\
                    &\leq \pr \left [\exists t \geq 1:  \sum_{i \in [n]} N_i^X(t) d(\Ah_i(t), \A_i) + \sum_{j \in [k]} N^Z_j(t) d(\muh_j(t), \mu_j) >  \beta(t, \delta)\right] \leq \delta.
\end{align*}
The last inequality is correct by Lemma \ref{lem: new threshold}.

\end{proof}

\subsection{Proof of New Threshold (Lemma \ref{lem: new threshold})}

First, we define the function $C_{exp}$ from \citet{kaufmann2021mixture}.

\begin{definition}[\citet{kaufmann2021mixture}]\label{def: c-g function}
   The function $C_{exp}$ is defined as
    \begin{align*}
            C_{exp}(x) \triangleq \min_{\xi \in (0, \frac12], \lambda \in (0, \frac{1}{1 + \xi})} \frac{g_{\xi}(\lambda) + x}{\lambda},
    \end{align*}
     where for any $\xi \in (0, \frac12]$,  $g_{\xi}: [0, \frac{1}{1 + \xi}) \to \mathbb{R}$ is defined as 
    \begin{equation}\label{eq: g-xi funct}
          g_{\xi} (\lambda) \triangleq \lambda (1 + \xi) \ln (\frac{2\zeta(2)}{(\ln(1 + \xi))^2}) - \ln (1- \lambda(1 + \xi))
    \end{equation}   
    
    and $\zeta$ denotes the Riemann zeta function, defined as $\zeta(s) = \sum_{n=1}^{\infty} n^{-s}$.
\end{definition}

We combined the construction of mixture martingale for categorical and one-parameter exponential family distributions as proposed by \citet{rl2-al2021navigating, kaufmann2021mixture}.

Based on Lemma 11 in \citet{kaufmann2021mixture}, for any fixed $\xi\in(0,1/2]$ and $\lambda\in(0,1/(1+\xi))$, there exists a non-negative martingale $\M_t$ such that

\begin{itemize}
    \item $\M_0 = 1$,
    \item $\forall t \geq 1: \ln(\M_t) \geq \lambda  \left ( \sum_{j \in [k]} N^Z_j(t) d(\muh_j(t), \mu_j) - 3 \sum_{j \in [k]} \ln \paran{1 + \ln \paran{N^Z_j(t)}} \right ) -  k g_{\xi} (\lambda)$,
\end{itemize}
Here $\M_t$ denotes the reward mixture martingale for this fixed pair $(\xi,\lambda)$. Terms corresponding to zero context counts in the divergence sum and its log-log correction are defined to be zero.

Now, recall $\N_t$, the non-negative martingale defined in the proof of Lemma~\ref{lem: concen-cat}. Note that $\N_t$ and $\M_t$ are dependent martingales, and the product of two dependent martingales is not necessarily a martingale. However, in the setting of our problem, fortunately, we can show that $\N_t^\lambda \M_t$ is a non-negative supermartingale. To show this, according to the tower rule of conditional expectation, we have 
\begin{align*}
    \mathbb{E} \left [\M_t  \N_t^\lambda \mid \mathcal{F}_{t-1} \right ] = \mathbb{E} \left [\mathbb{E} [\M_t \N_t^\lambda \mid \mathcal{F}_{t-1}, X_t = i_t, Z_t = z_t ] \mid \mathcal{F}_{t-1} \right ],
\end{align*}
where $\mathcal F_t$ denotes the $\sigma$-algebra generated by the observable variables up to $t$. Note that, conditioned on $\mathcal{F}_{t-1}, X_t = i_t, Z_t = z_t$, the value of martingale $\N_t$ is deterministic (see the construction of $\N_t$ in \citet{rl2-al2021navigating}). Hence,

$$
    \mathbb{E} \left [\mathbb{E} [\M_t \N_t^\lambda \mid \mathcal{F}_{t-1}, X_t = i_t, Z_t = z_t ] \mid \mathcal{F}_{t-1} \right ] = \mathbb{E} \left [\N_t^\lambda \mathbb{E} [\M_t \mid \mathcal{F}_{t-1}, X_t = i_t, Z_t = z_t ] \mid \mathcal{F}_{t-1} \right ]
$$
The reward distribution given $Z_t$ does not depend on $X_t$, and the context-wise reward mixture updates only the factor associated with $Z_t$. Therefore,
$$
    \mathbb{E} [\M_t \mid \mathcal{F}_{t-1}, X_t = i_t, Z_t = z_t ] =  \mathbb{E} [\M_t \mid \mathcal{F}_{t-1}, Z_t = z_t]
$$

The construction of $\M_t$ is based on the mixture of martingales (see section 2.3 of \citet{kaufmann2021mixture}). For each context, the reward likelihood-ratio increment has conditional expectation one, and integration over the fixed mixture preserves this property. Consequently, $\mathbb{E} [\M_t \mid \mathcal{F}_{t-1}, Z_t = z_t] = \M_{t-1}$. Combining the above results, we obtain

\begin{align*}
       \mathbb{E} \left [\M_t \N_t^\lambda \mid \mathcal{F}_{t-1} \right ] &=   \mathbb{E} \left [\N_t^\lambda \mathbb{E} [\M_t \mid \mathcal{F}_{t-1}, X_t = i_t, Z_t = z_t ] \mid \mathcal{F}_{t-1} \right ] \\
                        &= \mathbb{E}  \left [\N_t^\lambda \M_{t-1} \mid \mathcal{F}_{t-1} \right ] \\
                        &= \M_{t-1}\mathbb E[\N_t^\lambda\mid\mathcal F_{t-1}] \\
                        &\leq \M_{t-1}\bigl(\mathbb E[\N_t\mid\mathcal F_{t-1}]\bigr)^\lambda
                         = \M_{t-1}\N_{t-1}^\lambda.
\end{align*}

The inequality is conditional Jensen's inequality, since $0<\lambda<1$ and $x\mapsto x^\lambda$ is concave.

Finally, $\N_t^\lambda \M_t$ is a non-negative supermartingale with $\M_0 \N_0^\lambda = 1$. Applying Ville’s inequality gives
\begin{align*}
    \pr [\exists t \geq 0: \ln(\N_t^\lambda \M_t) \geq \logdel] &= \pr [\exists t \geq 1: \ln(\N_t^\lambda \M_t) \geq \logdel]  \leq \delta \\
\end{align*}

Next, we provide a lower bound for $\ln(\N_t^\lambda \M_t)$. By the definition of the powered product and the two martingale bounds, we have
\begin{align*}
    \ln(\N_t^\lambda \M_t) & = \lambda \ln(\N_t) + \ln(\M_t) \\
                  &\geq \lambda \left( \sum_{i \in [n]} N_i^X(t) d(\Ah_i(t), \A_i) - (k-1) \sum_{i \in [n]} \ln \paran{e \paran{1 + \frac{N^X_i(t)}{k-1}}} + \right. \\ 
                  &\qquad\qquad \left. \sum_{j \in [k]} N^Z_j(t) d(\muh_j(t), \mu_j) - 3\sum_{j \in [k]} \ln \paran{1 + \ln \paran{N^Z_j(t)}} \right) -   k g_{\xi} (\lambda)
\end{align*}

Therefore, 
\begin{align*}
 \delta \geq \pr [\exists t \geq 1: &\ln(\N_t^\lambda \M_t) \geq \logdel] \geq \pr \left [\exists t \geq 1:  \sum_{i \in [n]} N_i^X(t) d(\Ah_i(t), \A_i) + \sum_{j \in [k]} N^Z_j(t) d(\muh_j(t), \mu_j) \geq  \right. \\
 &\left. (k-1) \sum_{i \in [n]} \ln \paran{e \paran{1 + \frac{N^X_i(t)}{k-1}}} + 3\sum_{j \in [k]} \ln \paran{1 + \ln \paran{N^Z_j(t)}} + \frac{\logdel + k g_{\xi} (\lambda)}{\lambda} \right ]
\end{align*}
We are free to optimize over $\xi$ and $\lambda$ in the above inequality. For fixed $\delta$, these parameters are chosen deterministically before observing the data. Finally, based on the definition of $C_{exp}(x)$ (see Definition~\ref{def: c-g function}), we have
\begin{align*}
   &\pr \left [\exists t \geq 1:  \sum_{i \in [n]} N_i^X(t) d(\Ah_i(t), \A_i) + \sum_{j \in [k]} N^Z_j(t) d(\muh_j(t), \mu_j) \geq \right. \\
 &\left. (k-1) \sum_{i \in [n]} \ln \paran{e \paran{1 + \frac{N^X_i(t)}{k-1}}} + 3\sum_{j \in [k]} \ln \paran{1 + \ln \paran{N^Z_j(t)}} + k C_{exp}(\frac{\logdel}{k})\right ] \leq \delta
\end{align*}
which concludes the proof.

\subsection{Naive Thresholds for Unknown Case} \label{apd: sub-optimal threshold}

As discussed earlier, a straightforward approach to building thresholds is to sum the thresholds for the categorical case and for one-parameter exponential families. From \citet{kaufmann2021mixture}, we have the following uniform-time concentration inequality for a one-parameter exponential family:

\begin{equation} \label{eq: concen-exp}
 \pr \left [\exists t \geq 1:  \sum_{j \in [k]} N^Z_j(t) d(\muh_j(t), \mu_j) \geq 3 \sum_{j \in [k]} \ln \paran{1 + \ln \paran{N^Z_j(t)}} + k C_{\text{exp}} \paran{\frac{\logdel}{k}}   \right] \leq \delta
\end{equation}
We define 
\begin{align*}
    \beta_1(t, \delta) &\triangleq \beta_{\mu}(t, \delta), \\
    \beta_2(t, \delta) &\triangleq 3 \sum_{j \in [k]} \ln \paran{1 + \ln \paran{N^Z_j(t)}} + k C_{\text{exp}} \paran{\frac{\logdel}{k}},
\end{align*}
where $\beta_{\mu}(t, \delta)$ is given in \eqref{eq: known beta}. Consider the stopping rule
\begin{equation}\label{eq: naive-stop}
    \tau_{\delta} \triangleq \inf \{ t \in \mathbb{N} \mid \lamh(t) > \beta_1(t, \frac{\delta}{2}) + \beta_2(t, \frac{\delta}{2}) \}.
\end{equation}

This rule is $\delta$-correct (proof given later), but it results in a sample complexity strictly worse than the threshold in \eqref{eq: new beta}. Indeed,
$$
    \beta_1(t, \frac{\delta}{2}) + \beta_2(t, \frac{\delta}{2}) = \beta(t, \delta) + \ln (\frac{2}{\delta}) + k \left ( C_{\text{exp}} \paran{\frac{\ln (\frac{2}{\delta})}{k}} - C_{\text{exp}} \paran{\frac{\ln (\frac{1}{\delta})}{k}} \right).
$$
For small values of $\delta$, this gives
$$
     \beta_1(t, \frac{\delta}{2}) + \beta_2(t, \frac{\delta}{2}) \approx \beta(t, \delta) + \ln (\frac{4}{\delta}).
$$

Thus, the naive threshold incurs an overhead of $\ln(4/\delta)$, leading to suboptimal sample complexity. The sample complexity using this threshold is upper bounded by
$$ 
     \limsup_{\delta \to 0} \frac{\mathbb{E}_{\bmu, \A}[\tau_{\delta}]}{\logdel} \leq 2 T^*(\bmu, \A).
$$ 
However, our threshold avoids this factor-2 overhead and matches the lower bound $T^*(\bmu, \A)$.

\begin{remark}
    The above discussion is also applicable in prior work in RL literature \citep{rl2-al2021navigating, al2021adaptive}. Both \citet{rl2-al2021navigating, al2021adaptive} used a similar idea of naive thresholds, and their sample complexity does not match their lower bound. For example, Theorem 6 of \citet{rl2-al2021navigating} states
            $$
                \limsup_{\delta \to 0} \frac{\mathbb{E}[\tau_{\delta}]}{\logdel} \leq 2 U_0 (\M).
            $$
    We expect our joint-threshold argument to extend to these RL
settings and remove this factor of $2$ from their upper bounds.
\end{remark}

\begin{lemma}
    Let $(\bmu, \A) \in \I$ be a bandit instance with post-action context. Any algorithm that uses the stopping rule in \eqref{eq: naive-stop} is $\delta$-correct.    
\end{lemma}
\begin{proof}
    The proof is straightforward and similar to proof of Lemma \ref{lem: known stopping} and \ref{lem: unknown stopping}.

\begin{align*}
&\pr_{\bmu,\A}\big(
    \tau_\delta<\infty,\,
    \hat{i}_{\tau_\delta}\neq\istar
\big)\\*
&\quad=
\pr\left[
\begin{aligned}
    \exists t\geq1:\quad
    &\lamht>
    \beta_1\left(t,\frac{\delta}{2}\right)
    +\beta_2\left(t,\frac{\delta}{2}\right),\\
    &i_t^*\neq i^*
\end{aligned}
\right]\displaybreak[2]\\
&\quad=
\pr\left[
\begin{aligned}
    \exists t\geq1:\quad
    &\inf_{(\lambb,\B)\in\text{Alt}(\muht,\Aht)}
    \Bigg\{
        \sum_{i\in[n]}N_i^X(t)d(\Ah_i(t),\B_i)\\
    &\qquad+
        \sum_{j\in[k]}N_j^Z(t)d(\muh_j(t),\lambda_j)
    \Bigg\}\\
    &>
    \beta_1\left(t,\frac{\delta}{2}\right)
    +\beta_2\left(t,\frac{\delta}{2}\right),
    \quad i_t^*\neq i^*
\end{aligned}
\right]\displaybreak[2]\\
&\quad\leq
\pr\left[
\begin{aligned}
    \exists t\geq1:\quad
    &\sum_{i\in[n]}N_i^X(t)d(\Ah_i(t),\A_i)\\
    &\quad+
    \sum_{j\in[k]}N_j^Z(t)d(\muh_j(t),\mu_j)\\
    &>
    \beta_1\left(t,\frac{\delta}{2}\right)
    +\beta_2\left(t,\frac{\delta}{2}\right)
\end{aligned}
\right]\displaybreak[2]\\
&\quad\leq
\pr\left[
\begin{aligned}
    \exists t\geq1:\quad
    &\sum_{i\in[n]}N_i^X(t)d(\Ah_i(t),\A_i)
    >\beta_1\left(t,\frac{\delta}{2}\right)\\
    &\text{or}\quad
    \sum_{j\in[k]}N_j^Z(t)d(\muh_j(t),\mu_j)
    >\beta_2\left(t,\frac{\delta}{2}\right)
\end{aligned}
\right]\displaybreak[2]\\
&\quad\leq
\pr\left[
    \exists t\geq1:
    \sum_{i\in[n]}N_i^X(t)d(\Ah_i(t),\A_i)
    >\beta_1\left(t,\frac{\delta}{2}\right)
\right]\\*
&\qquad+
\pr\left[
    \exists t\geq1:
    \sum_{j\in[k]}N_j^Z(t)d(\muh_j(t),\mu_j)
    >\beta_2\left(t,\frac{\delta}{2}\right)
\right]\\*
&\quad\leq
\frac{\delta}{2}+\frac{\delta}{2}=\delta.
\end{align*}
    where we used union bound combining with Lemma \ref{lem: concen-cat} and Equation \ref{eq: concen-exp}.
    
\end{proof}

\section{Proofs of sample complexity upper bound} \label{apd: proof-sample}

This section presents the proofs of the sample complexity upper bounds for our algorithm, Theorem \ref{thm : upper}. We write the general proof for the case of the unknown $\bmu$, and at the end explain the differences for the case where $\bmu$ is known. 
We first present some notation and lemmas, which are used in the proof. 

For a point $p \in \mathbb{R}^k$ and a set $C \in \mathbb{R}^k$, we use $dist_{L^2}(p, C)$ to denote the distance of $p$ to $C$ with respect to $L^2$ norm, more precisely 
\begin{align*}
    dist_{L^2}(p, C) = \inf_{c \in C} \left\|{p - c}\right\|_2.
\end{align*}

\begin{lemma} \label{lem: continuouity}
        For each $(\bmu, \A) \in \mathcal{I}$, the set $\wstar$ defined as
        $$
        \wstar = \argmax_{\wb \in \Delta^{n-1}} F \paran{\bmu, \A, \wb, \qb(\wb)},
        $$
        where $q(\wb)_j = \sum_{i \in [n]} w_i \A_{j,i}$,
        is a convex set and the set-valued mapping $(\bmu, \A) \rightarrow \wstar$ is upper hemicontinuous. 
\end{lemma}
\begin{proof}
        Recall that $\wstar$ is the set of solutions that maximize
        \begin{align} \label{eq: w star set formula}
          F(\bmu, \A, \wb ,\qb(\wb)) = \inf_{(\lambb, \B) \in \text{Alt}(\bmu, \A)} \sum_{i \in [n]} w_i d \paran{\A_i, \B_i} + \sum_{j \in [k]} q_j d \paran{\mu_j, \lambda_j}.
        \end{align}
        Since this function is an infimum over linear functions in $\wb$, thus, $F$ is concave with respect to $\wb$. Consequently, if $\wb_1$ and $\wb_2$ are both maximizers of \eqref{eq: w star set formula}, then any convex combination of them is also a maximizer, implying that $\wstar$ is a convex set. 

        Note that, since the function $F$ satisfies conditions of Theorem 22 in \citet{mutiple-correct-answers-degenne2019pure}, the mapping $(\bmu, \A) \rightarrow \wstarht$ is hemicontinuous.
\end{proof}

Note that the above lemma is also true for known $\mu$ case.

Let $p_{\min} = \min_{j,i} \A_{j,i}$. By assumption, $p_{\min} > 0$. Since our analysis focuses on the asymptotic regime, the initialization phase of the algorithm contributes only a constant term to the total sample complexity. Specifically, for each action $i$ and context $j$, the expected waiting time to observe a sample with $(X=i, Z=j)$ is at most $\frac{1}{p_{\min}}$, as the probability of encountering context $j$ when playing action $i$ is at least $p_{\min}$. Hence, the expected number of initialization steps is bounded by $\frac{kn}{p_{\min}}$, which is constant, and  
$
\limsup_{\delta \to 0} \frac{kn / p_{\min}}{\log(1/\delta)} = 0.
$
For convenience, we set $t = 0$ at the end of the initialization phase.

For any non-negative real number $\alpha$ and any instance $(\bmu, \A)$, we define the $L^2(\alpha)$-approximation set of $\wstar$ as 
\begin{align*}
    \wb_\alpha^*(\bmu, \A) \triangleq \left\{p \in \Delta^{n-1} \mid dist_{L^2}(p, \wstar) \leq \alpha \right \}.
\end{align*}

Based on Lemma~\ref{lem: continuouity}, the set $\wstar$ is convex, which implies that for each $\alpha$, the set $\wb_\alpha^*(\bmu, \A)$ is also convex.  
Moreover, by Lemma~\ref{lem: continuouity} and the fact that the set of instances $(\bmu', \A')$ having the same best arm as $(\bmu, \A)$ is open in $\mathbb{R}^{k + kn}$, for any $\epsilon > 0$ there exists a constant $\xi = \xi(\epsilon, \bmu, \A)$ such that for every instance $(\bmu', \A')$,  
\begin{align*}
    \max \big(\norm{\bmu' - \bmu}_{\infty}, \norm{\A' - \A}_{\infty} \big) \leq \xi 
    \ \Rightarrow \ 
    (\bmu', \A') \in \mathcal{I}, i^*(\bmu', \A') = \istar, 
    \wb^*(\bmu', \A') \subseteq \wb^*_\epsilon(\bmu, \A).
\end{align*}
We then define  
$$
\I_{\epsilon} \triangleq \Big\{ (\bmu', \A') \in \I \ \Big| \ \max \big( \norm{\bmu' - \bmu}_{\infty}, \norm{\A' - \A}_{\infty} \big) \leq \xi \Big\}.
$$
Let $\nbarxt = \frac{N^X(t)}{t}, \pbart = \frac{\sum_{s \in [t]} \pb(s)}{t}$. For each $T$, we define three good events as follows
\begin{align*}
    \eventmu 
    &\triangleq 
    \bigcap_{t=T^{1/4}}^{T} \left( \left\| \muht - \bmu \right\|_{\infty} \leq \xi \right), \\
    \eventa 
    &\triangleq 
    \bigcap_{t=T^{1/4}}^{T} \left( \left\| \hat{\A}(t) - \A \right\|_{\infty} \leq \xi \right), \\
    \eventx 
    &\triangleq 
    \bigcap_{t=T^{1/4}}^{T} \left( \left\| \nbarxt - \pbart \right\|_{2} \leq t^{-\frac14} \right).
\end{align*}
The following lemma shows that these events occur with high probability. 

\begin{lemma} \label{lem: event prob}
    There exist two constants $B$ and $C$ such that
    \begin{align}
        \pr \big( \big(\eventmu \cap \eventa \cap \eventx \big)^c \big) \leq B T \exp(-C T^{1/8}).
    \end{align}
\end{lemma}

\begin{proof}

We bound each event separately and combine them using union bound.

\textbf{Bounding $\eventmu^c $.} Recall the definition of $p_{\min} > 0$ as the minimum entry of matrix $\A$.
Based on the definition and union bound, we have:

$$
    P\left( \mathcal{E}_{T}^\mu(\epsilon)^c \right)
    \leq 
    \sum_{t=T^{1/4}}^T \sum_{j=1}^k P \left( \left\| \hat{\mu}_j(t) - \mu_j \right\| \geq \xi \right).
$$

Then
\begin{align*}
    P \left( \left\| \hat{\mu}_j(t) - \mu_j \right\| \geq \xi \right) 
    &= 
    P \left( \left\| \hat{\mu}_j(t) - \mu_j \right\| \geq \xi \mid N_j^Z(t) <  p_{\min} t^{3/4}\right) P \left( N_j^Z(t) <  p_{\min} t^{3/4} \right) \\ 
    &+
    P \left( \left\| \hat{\mu}_j(t) - \mu_j \right\| \geq \xi \mid N_j^Z(t) \geq  p_{\min} t^{3/4}\right) P \left( N_j^Z(t) \geq  p_{\min} t^{3/4} \right) \\ 
    &\leq 
    P \left( N_j^Z(t) <  p_{\min} t^{3/4} \right) + P \left( \left\| \hat{\mu}_j(t) - \mu_j \right\| \geq \xi \mid N_j^Z(t) \geq  p_{\min} t^{3/4}\right)
\end{align*}

Note that \( N_j^Z(t) \) can be written as
\[
N_j^Z(t) = \sum_{s=1}^t \mathbbm{1}_{(z_s=j)}.
\]

Therefore, if \( W_s \) denotes \( \mathbbm{1}_{(z_s=j)} \), it is a Bernoulli random variable with parameter greater than or equal to \( p_{\min} \). By Hoeffding's inequality, we have
\begin{align*}
P \left( N_j^Z(t) < p_{\min} t^{3/4} \right) &= P \left( \sum_{s=1}^t W_s < p_{\min} t^{3/4} \right) \\
&\leq P \left( \sum_{s=1}^t W_s - \mathbb{E}[N_j^Z(t)] < p_{\min}  t^{3/4} - t \, p_{\min} \right) \\
&\leq \exp \left( \frac{-2p_{\min}^2 (t^{3/4} - t)^2}{t} \right) = \exp \left( -2p_{\min}^2 \, t(t^{-1/4} - 1)^2 \right) \\
&\leq \exp \left( -p_{\min}^2 t \right),
\end{align*}

where the last inequality holds for sufficiently large \( t \).

Then, by Chernoff's inequality, we have

\[
P\left(|\hat{\mu}_j(t) - \mu_j| \geq \xi \mid N_j^Z(t) \geq p_{\min} t^{3/4}\right) \leq 2 \exp \left( - p_{\min} t^{3/4} \min \big(d(\mu_j + \xi, \mu_j), d(\mu_j - \xi, \mu_j) \big) \right).
\]

Therefore,
\[
P\left(|\hat{\mu}_j(t) - \mu_j| \geq \xi\right) \leq \exp \left( -p_{\min}^2 t \right) + 2 \exp \left( - p_{\min} t^{3/4} \min \big(d(\mu_j + \xi, \mu_j), d(\mu_j - \xi, \mu_j) \big) \right),
\]
and then,
\[
P\left(\eventmu^c\right) \leq \sum_{t=T^{1/4}}^T \sum_{j \in [k]} P\left(|\hat{\mu}_j(t) - \mu_j| > \xi\right) \leq B_1 T \exp (-C_1 T^{1/8})
\]
where \( B_1 \) and \( C_1 \) are constants and functions of \( \xi \), \( p_{\min} \) and \( k \):

$$
B_1 = 3k, C_1 = \min \paran{ p_{\min}^2 , \min_{j \in [k]} \big( d(\mu_j + \xi, \mu_j), d(\mu_j - \xi, \mu_j)\big)}
$$

\textbf{Bounding $\eventa^c $.} Our approach is similar to bounding $\mathcal{E}_{T}^\mu(\epsilon)^c$. Based on the definition and union bound, we have

$$
P\left( \eventa)^c \right)
\leq 
\sum_{t=T^{1/4}}^T \sum_{i=1}^n \sum_{j=1}^k P \left( \left| \hat{\A}_{j,i}(t) - \A_{j,i} \right| \geq \xi \right).
$$

Now we have

\begin{align*}
P \left( \left| \hat{\A}_{j,i}(t) - \A_{j,i} \right| \geq \xi \right)
&= 
P \left( \left| \hat{\A}_{j,i}(t) - \A_{j,i} \right| \geq \xi \mid N_i^X(t) <  \frac{\sqrt{t}}{2 n} \right) P \left( N_i^X(t) <  \frac{\sqrt{t}}{2 n} \right) \\ 
&+
P \left( \left| \hat{\A}_{j,i}(t) - \A_{j,i} \right| \geq \xi \mid N_i^X(t) \geq  \frac{\sqrt{t}}{2 n} \right) P \left( N_i^X(t) \geq \frac{\sqrt{t}}{2 n} \right) \\ 
&\leq 
P \left( N_i^X(t) <  \frac{\sqrt{t}}{2 n} \right) + P \left( \left| \hat{\A}_{j,i}(t) - \A_{j,i} \right| \geq \xi \mid N_i^X(t) \geq \frac{\sqrt{t}}{2 n} \right)
\end{align*}

Note that \( N_i^X(t) \) can be written as
\[
N_i^X(t) = \sum_{s=1}^t \mathbbm{1}_{(i_s=i)}.
\]

Considering the sampling rule in \eqref{eq: g-tracking formula}, the algorithm plays uniformly at random on rounds with square indices. Hence, up to round $t$, it performs approximately $\sqrt{t}$ random plays, and during these rounds the probability of selecting arm $i$ is $\frac{1}{n}$.  
Let $W_s = \mathbbm{1}_{(i_{s^2} = i)}$ for $s \le \sqrt{t}$; then $W_s$ is a Bernoulli random variable with parameter $\frac{1}{n}$. By Hoeffding’s inequality, we obtain

\begin{align*}
P \left( N_i^X(t) < \frac{\sqrt{t}}{2 n} \right) &\leq P \left( \sum_{s=1}^{\sqrt{t}} W_s < \frac{\sqrt{t}}{2 n} \right) \\
&\leq P \left( \sum_{s=1}^{\sqrt{t}} W_s - \mathbb{E}[N_i^X(t)] < \frac{\sqrt{t}}{2n} - \frac{\sqrt{t}}{n} \right) \\
&\leq \exp \left({-\frac{\sqrt{t}}{2n^2}} \right).
\end{align*}

Similarly, by applying Hoeffding's inequality for the binary variables $\mathbbm{1}_{(z_s = j)}$, we obtain
\[
 P \left( \left| \hat{\A}_{j,i}(t) - \A_{j,i} \right| \geq \xi \mid N_i^X(t) \geq \frac{\sqrt{t}}{2 n} \right) \leq 2 \exp \left( -2 \xi^2 \frac{\sqrt{t}}{2 n} \right).
\]

Therefore,
\[
P \left( \left| \hat{\A}_{j,i}(t) - \A_{j,i} \right| \geq \xi \right) \leq \exp \left( -\frac{\sqrt{t}}{2n^2} \right) + 2 \exp \left( -2 \xi^2 \frac{\sqrt{t}}{2 n} \right) \leq 3 \exp \left( - {\xi^2}\frac{\sqrt{t}} {2n^2} \right),
\]
which holds for $\xi \leq 1$, and then,
\[
P\left( \eventa^c \right)
\leq 
\sum_{t=T^{1/4}}^T \sum_{i=1}^n \sum_{j=1}^k P \left( \left| \hat{\A}_{j,i}(t) - \A_{j,i} \right| \geq \xi \right) \leq B_2 T \exp (-C_2 T^{1/8}),
\]
where \( B_1 \) and \( C_1 \) are constants and functions of \( \xi \), \( p_{\min} \) and \( k \):

$$
B_2 = 3nk, C_2 = \frac{\xi^2}{2n^2}.
$$

\textbf{Bounding $\eventx^c $.}
This event is exactly the same as the event $\E'_T(\epsilon)$ in \citet{shahverdikondori2025optimal}, where they prove in \text{Lemma F.3} that there are constants $B_3, C_3$ such that
$$
P\left( \eventx^c \right) \leq B_3T \exp (-C_3 T^{1/8}).
$$

\textbf{Overall bound.} Using the individual bounds calculated above, we obtain the following overall bound:

\begin{align*}
    P\left(\left(\eventmu \cap \eventa \cap \eventx \right)^c\right) 
    &\leq 
    P(\eventmu^c) + P(\eventa^c) + P(\eventx^c)  \\ 
    &\leq 
    \sum_{i=1}^3 B_i T \exp (-C_i T^{1/8}) \\
    &\leq
    BT \exp(-C T^{1/8}).
\end{align*}

with $B = 2 \max(B_1, B_2, B_3)$ and $C = \min(C_1, C_2, C_3)$.
\end{proof}

Next, we need to show that when actions are selected according to the modified G-tracking rule and under the defined good events, the empirical frequency vector $\nbarxt$ remains close to the set of optimal frequencies $\wstar$. While this property has been established for the original G-tracking algorithm in~\citet{shahverdikondori2025optimal}, the addition of forced exploration rounds requires a refined analysis. The following lemma provides a new bound on this distance for our modified version.

\begin{lemma} \label{lem: G-tracking}
    For each $\epsilon > 0$, there exist a constant $T_{\epsilon}$ such that for each $T \geq T_{\epsilon}$, on event $\eventmu \cap \eventa \cap \eventx$, we have 
    \begin{align*}
        \forall t \geq \sqrt{T} : dist_{L^2}\paran{\nbarxt, \wstar} \leq 6 \epsilon. 
    \end{align*}
\end{lemma} 

\begin{proof}
    Let  
    \begin{align*}
        Y_t = \dls{\nbarxt}{\wepsstar{2\epsilon}}, 
        \quad 
        X_t = \dls{\pbart}{\wepsstar{2\epsilon}},
    \end{align*}
    where $\wepsstar{2\epsilon}$ denotes the set of points whose distance from the optimal frequency set $\wstar$ is at most $2\epsilon$.  
    For two points $A, B \in \mathbb{R}^k$, let $\overline{AB}$ denote the length of the segment connecting them.  
    
    Lemma~F.4 in~\citet{shahverdikondori2025optimal} establishes that for each $T > \epsilon^{-16}$, under event $\eventmu \cap \eventa \cap \eventx$, the following statements hold for all $t > T^{1/4}$:
    \begin{enumerate}
        \item $\overline{\pbart \nbarxt} < \epsilon$;
        \item $\overline{\pbart, \overline{\pb}(t+1)} \leq \epsilon$, which implies $|X_{t+1} - X_t| \leq \epsilon$;
        \item If $X_t \leq \epsilon$ and $\sqrt{t} \notin \mathbb{N}$, then $X_{t+1} \leq \frac{t}{t+1} X_t$.
    \end{enumerate}
    
    We prove that if, for some $t > T^{1/4}$, $X_t \leq \epsilon$, then  
    \begin{align} \label{eq: wt four eps}
        \forall t' > t: \quad X_{t'} \leq 3\epsilon.
    \end{align}
    
    We prove this later. Having this property, define  
    $$
    \tau = \inf \{ t > T^{1/4} \mid X_t \leq \epsilon \}.
    $$
    
    We show that $\tau < \sqrt{T}$. Suppose, for contradiction, that $\tau > \sqrt{T}$, meaning $X_t > \epsilon$ for all $t \in (T^{1/4}, \sqrt{T}]$.  
    Let $T^{1/4} < m_1 < m_2 < \ldots < m_r \leq \sqrt{T}$ denote all perfect squares in this interval. Forced exploration occurs only at rounds $m_i + 1$; for all other rounds, $X_{t+1} \leq \frac{t}{t+1} X_t$.  
    
    During forced exploration, since  
    $\overline{\pb}(t+1) = \frac{t}{t+1} \pbart + \frac{\mathbf{p}(t+1)}{t+1}$,  
    the convexity of the distance function implies
    $$
    X_{t+1} \leq \frac{t}{t+1} X_t + \frac{\sqrt{2}}{t+1},
    $$
    as the maximum distance between any two points on the simplex $\Delta^{n-1}$ is $\sqrt{2}$.  
    
    Let $\alpha = \left\lceil \frac{\sqrt{2}}{\epsilon} \right\rceil$. For each $i$, we have
    \begin{align*}
        X_{m_i + 1} &\leq X_{m_i}\frac{m_i}{m_i + 1} + \frac{\sqrt{2}}{m_i + 1}, \\
        \Longrightarrow \quad 
        X_{m_i + \alpha} &\leq X_{m_i + 1} \frac{m_i + 1}{m_i + \alpha} \leq X_{m_i} \frac{m_i}{m_i + \alpha} + \frac{\sqrt{2}}{m_i + \alpha}.
    \end{align*}
    After simplification, this gives
    $$
    X_{m_i + \alpha} \leq X_{m_i},
    $$
    whenever $m_{i+1} - m_i > \alpha$, which holds for sufficiently large $T$.  
    Thus, $X_t$ decreases according to the rule $X_{t+1} \leq \frac{t}{t+1} X_t$ between consecutive perfect squares. Consequently,
    \begin{align*}
        X_{m_r} 
        \leq X_{m_{r-1}} \frac{m_{r-1} + \alpha}{m_r}
        \leq X_{m_1} 
        \prod_{i=2}^{r} \frac{m_{i-1} + \alpha}{m_i}.
    \end{align*}
    
    This product equals
    $$
    \frac{m_r - 1}{m_r} \cdot \frac{m_r - 2}{m_r - 1} \cdots \frac{m_1 - 1}{m_1},
    $$
    where $r\alpha$ terms (one block of $\alpha$ terms before each $m_i$) have been removed.  
    Removing these $r\alpha$ terms makes the product smaller compared to the case where the first $ r\alpha$ terms are removed, implying
    $$
    X_{m_r} \leq X_{m_1} \, \frac{m_1 + r\alpha - 1}{m_{r}}.
    $$
    Since $m_{r} = \mathcal{O}(\sqrt{T})$, $m_1 = \mathcal{O}(T^{1/4})$, $r = \mathcal{O}(T^{1/4})$, and $\alpha$ is constant with respect to $T$, the product $X_{m_1} \, \frac{m_1 + r\alpha - 1}{m_{r}}$ becomes smaller than $\epsilon$ for large enough $T$. This contradicts the assumption $\tau > \sqrt{T}$, thus $\tau < \sqrt{T}$.  
    
    Finally, by~\eqref{eq: wt four eps}, for all $t > \sqrt{T}$ we have  
    \begin{align*}
        X_t \leq 3\epsilon 
        \quad \Longrightarrow \quad
        Y_t \leq 4\epsilon 
        \quad \Longrightarrow \quad
        \dls{\nbarxt}{\wstar} \leq 6\epsilon.
    \end{align*}
    We now prove \eqref{eq: wt four eps}.  
    Suppose, for contradiction, that there exist $t > T^{1/4}$ such that $X_t \le \epsilon$, and some $t' > t$ with $X_{t'} > 3\epsilon$.  
    Let $t''$ denote the last round before $t'$ such that $X_{t'' - 1} \le \epsilon$, which implies that for all $s \in [t'', t']$, we have $X_s > \epsilon$.  
    Let $m$ be the smallest perfect square satisfying $m \ge t''$.  
    
    For rounds $s$ between $t''$ and $m$, $X_s$ is monotonically decreasing, so $X_m \le X_{t''} \le 2\epsilon$, where the second inequality follows from $|X_{t''} - X_{t'' - 1}| \le \epsilon$.  
    At the forced-exploration round, we have $X_{m+1} \le 3\epsilon$.  
    For all $s \in [m + 2, (\sqrt{m} + 1)^2]$, $X_s$ again decreases, and as shown earlier, for indices $s > m + 1 + \alpha$, we have $X_s < X_m$.  
    Hence, $X_{(\sqrt{m} + 1)^2} \le X_m$, and for all intermediate $s$, $X_s \le 3\epsilon$.  
    
    Repeating this reasoning recursively for $(\sqrt{m} + 1)^2$, $(\sqrt{m} + 2)^2$, and so on, shows that $X_s$ always remains below $3\epsilon$.  
    Therefore, such a $t'$ cannot exist, completing the proof.

\end{proof}

On the event $\eventmu \cap \eventa \cap \eventx$, for each $\sqrt{T} \leq t \leq T$
    \begin{enumerate}
        \item $\muht$ is close to $\bmu$.
        \item $\Aht$ is close to $\A$.
        \item $i^*(\muht, \Aht) = i^*(\bmu, \A)$, hence Alt$(\muht, \Aht) = $ Alt$(\bmu, \A)$.
        \item $\nbarxt$ is close to $\wb^*$ for a $\wb^* \in \wstar$.
     \end{enumerate}

     Combining these properties allows us to derive a lower bound on $\hat \Lambda (t)$. 
     
     Let $\mathcal{O}_i = \left\{ (\bmu,\A) \in \I \ \mid \A_i^\top \bmu > \A_j^\top \bmu, \ \forall j \neq i \right\}$. We define the function $g: \mathcal{O}_{i^*(\bmu, \A)} \times \Delta ^ {n-1} \rightarrow \mathbb{R}$ as
   
    \begin{align} \label{eq: g-function-proof}
        g(\bmu', \A', \mathbf{w}') = \inf_{(\lambda, \B) \in \text{Alt}(\bmu', \A')} \sum_{i=1}^n w'_i \left(d(\A'_i, \B_i) + \sum_{j=1}^k \A'_{j,i} d(\mu'_j, \lambda_j)\right),
    \end{align}
    
     Note that $\lamht = t g\big(\muht, \Aht, \nbarxt\big)$. We also define the constant
    \begin{align*}
        C^*_\epsilon(\bmu, \A) = \inf_{\substack{(\bmu', \A') \in \I_{\epsilon} \\ \wb': \dls{\mathbf{w}'}{\wstar} \leq 6 \epsilon}} g(\bmu', \A',  \mathbf{w}').
    \end{align*}

    The mapping $(\bmu', \A', \wb') \rightarrow \inf_{(\lambda, \B) \in \text{Alt}(\mu', \A')} \sum_{i=1}^n w'_i \left(d(\A'_i, \B_i) + \sum_{j=1}^k \A'_{ji} d(\mu'_j, \lambda_j)\right)$ is jointly continuous and the constraint set is independent of $(\bmu', \A', \wb')$. Therefore, by Berge’s maximum theorem \citep{berge1963topological}, the function $g$ is continuous and $C^*_\epsilon(\bmu, \A)$ exists.

    Based on the definition of $C^*_\epsilon(\bmu, \A)$, on event $\eventmu \cap \eventa \cap \eventx$ and for $T > T_{\epsilon}$, we have $\forall t > \sqrt{T}$
    $$
        \lamht = t g \paran{\muht, \Aht, \nbarxt} \geq t \, C^*_\epsilon(\bmu, \A).
    $$

    Recall threshold, 
    $$    
         \beta(t, \delta) =  (k-1) \sum_{i \in [n]} \ln \paran{e \paran{1 + \frac{N^X_i(t)}{k-1}}} + 3\sum_{j \in [k]} \ln \paran{1 + \ln \paran{N^Z_j(t)}} + k C_{exp}\paran{\frac{\logdel}{k}}
    $$

    Since $N^X_i(T), N^Z_j(T) \leq T$, we can upper bound $\beta(T, \delta)$ by
    $$
        \beta(T, \delta) \leq n(k-1) \ln \Big(e\Big(1 + \frac{T}{k - 1}\Big)\Big) + 3k \ln ( 1 + \ln(T)) +  k C_{exp}\paran{\frac{\logdel}{k}}.
    $$

    Because all terms inside are sublinear in $T$ (i.e., $\ocal (\ln T)$), there exists a constant $T_2(\epsilon)$ depending on $\epsilon$ such that
    $$  
      \forall T \geq T_2(\epsilon): \beta(T, \delta) \leq (T\epsilon) C^*_\epsilon(\bmu, \A) + k C_{exp}\paran{\frac{\logdel}{k}}.
    $$
    We define another constant $T_3(\epsilon, \delta)$ with the property that
    \begin{align*}
        \forall T \geq T_3(\epsilon, \delta): \quad & T C^*_\epsilon(\bmu, \A) \geq (T\epsilon) C^*_\epsilon(\bmu, \A) + k C_{exp}\paran{\frac{\logdel}{k}} \\
        &  \iff T \geq \frac{k C_{exp}\paran{\frac{\logdel}{k}}}{(1 - \epsilon)C^*_\epsilon(\bmu, \A)}.
    \end{align*}
    
   Hence, we can define
    $$
        T_3(\epsilon, \delta) \triangleq \left \lceil \frac{k C_{exp}\paran{\frac{\logdel}{k}}}{(1 - \epsilon)C^*_\epsilon(\bmu, \A)} \right\rceil
    $$

    On event  $\eventmu \cap \eventa \cap \eventx$, and if $T \geq \max(T_\epsilon, T_2, T_3)$, we have
    \begin{align*}
       (T\epsilon) \, C^*_\epsilon(\bmu, \A) + k C_{exp}\paran{\frac{\logdel}{k}} &\leq  T C^*_\epsilon(\bmu, \A) \\
          \Longrightarrow \beta(T, \delta)  & \leq    T \, C^*_\epsilon(\bmu, \A) \\
          \Longrightarrow \beta(T, \delta)  & \leq   \hat{\Lambda} (T).
    \end{align*}

    Therefore, under this condition, the algorithm must stop before time $T$ because the GLR exceeds the threshold at round $T$. Hence, for all $t \geq \max(T_\epsilon, T_2, T_3)$,
    
         $$\{\tau_\delta \geq t\} \subseteq {\big(\eventmu \cap \eventa \cap \eventx\big)}^c.$$
    
    We can now bound the sample complexity as follows.

    \begin{align*}
        \mathbb{E} [\tau_{\delta}] &= \sum_{t = 1}^{\infty} \pr (\tau_\delta \geq t) \\
        &\leq \max(T_\epsilon, T_2, T_3) + \sum_{t = \max(T_\epsilon, T_2, T_3)}^{\infty} \pr (\tau_\delta \geq t) \\
        &\leq \max(T_\epsilon, T_2, T_3) + \sum_{t = \max(T_\epsilon, T_2, T_3)}^{\infty} \pr \paran{\big(\eventmu \cap \eventa \cap \eventx\big)^c} \\
        &\leq \max(T_\epsilon, T_2, T_3) + \sum_{t = 1}^{\infty} Bt \exp (-Ct^{1/8}) \\
        &\leq T_\epsilon + T_2 + T_3 + \sum_{t = 1}^{\infty} Bt \exp (-Ct^{1/8}).
    \end{align*}    

    Note that $T_\epsilon$, $T_2$, and $\sum_{t = 1}^{\infty} Bt \exp (-Ct^{1/8})$ are constant values with respect to $\delta$. Therefore,
    $$
        \limsup_{\delta \to 0} \frac{\mathbb{E} [\tau_{\delta}]}{\logdel}  \leq \limsup_{\delta \to 0} \frac{T_3}{\logdel}  = \limsup_{\delta \to 0} \frac{k C_{exp}\paran{\frac{\logdel}{k}}}{(1 - \epsilon)C^*_\epsilon(\bmu, \A) \logdel}.
    $$
    Since $C_{exp}(x) \approx x$ as $x \to \infty$, we have
    $$
           \limsup_{\delta \to 0} \frac{\mathbb{E} [\tau_{\delta}]}{\logdel}  \leq \frac{1}{(1- \epsilon)C^*_\epsilon(\bmu, \A)}.
    $$
    Finally, letting $\epsilon \to 0$ yields
    $$
        \limsup_{\delta \to 0} \frac{\mathbb{E} [\tau_{\delta}]}{\logdel} \leq  \lim_{\epsilon \to 0} \frac{1}{(1- \epsilon)C^*_\epsilon(\bmu, \A)} = T^*(\bmu, \A).
    $$

    \textbf{Known $\bmu$}. The extension of the proof to the case of known $\bmu$ is straightforward. Here, we review the important steps. We do not need to define $\eventmu$, because the algorithm is aware of vector $\bmu$. We have similar Lemmas \ref{lem: continuouity}, \ref{lem: G-tracking}, and \ref{lem: event prob}.
    We define new function $g_{\bmu}(\A', \wb')$ as  
    \[
    g_{\bmu}(\A',\wb')
        =
        \inf_{\B:\,(\bmu,\B)\in\text{Alt}(\bmu,\A')}
        \sum_{i=1}^{n} w'_i\,d(\A'_i,\B_i).
    \]

    instead of $g(\bmu', \A', \wb')$. Similarly, we can redefine $T_2(\epsilon)$ and $T_3(\epsilon, \delta)$, and then we reach the desired sample complexity.

    \subsection{Sample complexity of the naive thresholds} \label{apd: upperbound-naive}
        If we use $\beta_1(t, \delta / 2)$ and $\beta_2(t, \delta/2)$ instead of $\beta$, then we need to redefine $T_3(\epsilon, \delta)$ as 
        $$
            T'_3(\epsilon, \delta) \triangleq \frac{\ln(2 /\delta) + k \, C_{exp}(\frac{\ln(\frac{2}{\delta})}{k})}{(1 - \epsilon)C^*_\epsilon(\bmu, \A)}.
        $$
        Therefore, by limiting $\delta \to 0$, we have
        $$
            \limsup_{\delta \to 0} \frac{\mathbb{E} [\tau'_{\delta}]}{\logdel} \leq \lim_{\epsilon \to 0} \limsup_{\delta \to 0} \frac{\ln(2 /\delta) + k \, C_{exp}(\frac{\ln(\frac{2}{\delta})}{k})}{(1- \epsilon)C^*_\epsilon(\bmu, \A) \logdel} = 2T^*(\bmu, \A).
        $$
        As we discussed before, this upper bound is two times worse than the sample complexity upper bound obtained by using the proposed thresholds, according to Theorem \ref{thm : upper}.

\section{Proof of Proposition~\ref{prop: convexification failure}}

\subsection{Proof of Part (i): Alternatives Are Not a Finite Union of Convex Sets}
\label{apd: finite convex}

Here, we prove the existence of a fundamental complexity in the
case of a fully unknown environment by showing that for some
instances $(\bmu,\A)$, the set of alternative instances
$\text{Alt}(\bmu,\A)$ cannot be expressed as a finite union of
convex sets.

For any instance $(\bmu,\A)$ with $n$ actions and $k$ context
values, the pair can be represented as a point in
$\mathbb{R}^{k+kn}$, since $\bmu\in\mathbb{R}^k$ and
$\A\in\mathbb{R}^{k\times n}$. Consequently,
$\text{Alt}(\bmu,\A)$ is a subset of $\mathbb{R}^{k+kn}$.
First consider an instance with $n=2$, $k=3$, and $\istar=1$.
By the definition of valid instances,
$$
\text{Alt}(\bmu,\A)
=
\left\{
(\lambb,\B)\in\mathbb{R}^{9}
\ \Big|\
\B_1^\top\lambb<\B_2^\top\lambb,\ 
\lambb\in(0,1)^3,\ 
\B^\top\mathbbm{1}=\mathbbm{1},\
\B_{j,i}>0\ \text{for all }j,i
\right\},
$$
where $\mathbbm{1}$ is the all-ones vector.

Suppose, for contradiction, that there exist convex sets
$C_1,\ldots,C_M$ such that
$$
\text{Alt}(\bmu,\A)=\bigcup_{m=1}^{M}C_m.
$$
Intersect this set with the affine slice specified by
$\lambda_2=0.5$, $\lambda_3=0.9$,
$\B_{3,1}=0.5$, and $\B_{3,2}=0.6$.
On this slice, the column-sum constraints give
$\B_{2,1}=0.5-\B_{1,1}$ and
$\B_{2,2}=0.4-\B_{1,2}$. The inequality defining the
alternative set becomes
\begin{align*}
\B_1^\top\lambb<\B_2^\top\lambb
&\Longleftrightarrow
\lambda_1(\B_{1,1}-\B_{1,2})
+0.5\bigl(0.1-\B_{1,1}+\B_{1,2}\bigr)-0.09<0 \\
&\Longleftrightarrow
(\lambda_1-0.5)(\B_{1,1}-\B_{1,2})<0.04.
\end{align*}
Define $x=\lambda_1-0.5$ and
$y=\B_{1,1}-\B_{1,2}$. Positivity of the entries of $\B$
implies $\B_{1,1}\in(0,0.5)$ and
$\B_{1,2}\in(0,0.4)$. Hence the affine projection of the
intersection onto $(x,y)$ is exactly
$$
T=
\left\{
(x,y)\ \Big|\
xy<0.04,\quad
x\in(-0.5,0.5),\quad
y\in(-0.4,0.5)
\right\}.
$$
The image under an affine map of the intersection of a convex set
with an affine slice is convex. Thus our supposition would express
$T$ as a union of finitely many convex sets
$D_1,\ldots,D_M$.

Now consider the curve
$$
S=
\left\{
(x,0.04/x)\ \Big|\ x\in(0.1,0.4)
\right\}.
$$
We have $S\subseteq\partial T$. Since the union is finite,
$$
S\subseteq\overline{T}
=\bigcup_{m=1}^{M}\overline{D_m}.
$$
Therefore, two distinct points of $S$ belong to the same
$\overline{D_m}$. This closure is convex, so the segment joining
those points lies in $\overline{D_m}\subseteq\overline{T}$.
However, $x\mapsto0.04/x$ is strictly convex on $(0.1,0.4)$.
Every interior point $(x,y)$ of that segment lies strictly above
the curve and therefore satisfies $xy>0.04$. Such a point is
outside $\overline{T}$, a contradiction.

The same argument extends to every $n\geq2$ and $k\geq3$.
Fix an instance in $\I$ whose unique best arm is arm $1$.
Let $\rho=0$ if $k=3$, and choose $\rho\in(0,1)$ if $k>3$.
Extend each point $(\lambb,\B)$ of the preceding affine slice
to $k$ contexts and $n$ arms as follows. Keep the first three
entries of $\lambb$, set $\lambda_j=0.5$ for $j\geq4$, and,
for $i\in\{1,2\}$, set
$$
\widetilde{\B}_{j,i}
=
\begin{cases}
(1-\rho)\B_{j,i}, & j\in\{1,2,3\},\\
\rho/(k-3), & j\in\{4,\ldots,k\},
\end{cases}
$$
where the second case is used only when $k>3$. For every
additional arm $i\geq3$, set
$$
\widetilde{\B}_i
=
\frac{\widetilde{\B}_1+\widetilde{\B}_2}{2}.
$$
This extension is affine and preserves strict positivity and
the column-sum constraints. Moreover, the difference between
the expected rewards of arms $1$ and $2$ is
$$
\widetilde{\B}_1^\top\widetilde{\lambb}
-
\widetilde{\B}_2^\top\widetilde{\lambb}
=
(1-\rho)(xy-0.04).
$$
Each additional arm has the average of those two expected
rewards. Consequently, an extended instance has a unique best
arm different from arm $1$ precisely when $xy<0.04$.

If the alternative set for this $n$ and $k$ were a finite union
of convex sets, intersecting it with the affine slice defined
by this extension and projecting onto $(x,y)$ would express $T$
as a finite union of convex sets. This contradicts the argument
above and proves part~(i).

\subsection{Proof of Part (ii): Failure of the Convex-Hull Relaxation}

We now prove part~(ii). Let $\varepsilon=0$ if $k=2$, and choose
any $\varepsilon\in(0,1)$ if $k>2$. Construct three instances
indexed by $s\in\{0,+,-\}$ with reward-mean vectors
$$
\bmu^{(0)}=(0.6,0.4,0.5,\ldots,0.5),\quad
\bmu^{(+)}=(0.8,0.2,0.5,\ldots,0.5),\quad
\bmu^{(-)}=(0.4,0.6,0.5,\ldots,0.5).
$$
Set
$$
(p_1^{(0)},p_2^{(0)})=(0.6,0.4),\qquad
(p_1^{(+)},p_2^{(+)})=(0.4,0.6),\qquad
(p_1^{(-)},p_2^{(-)})=(0.8,0.2),
$$
and let $p_i^{(s)}=0.5$ for every $i\geq3$. For each
$s\in\{0,+,-\}$, define
$$
\A^{(s)}_{1,i}=(1-\varepsilon)p_i^{(s)},\qquad
\A^{(s)}_{2,i}=(1-\varepsilon)(1-p_i^{(s)}),
$$
and, when $k>2$, define
$$
\A^{(s)}_{j,i}=\frac{\varepsilon}{k-2},
\qquad j\in\{3,\ldots,k\}.
$$
All entries of these matrices are positive, and their columns
sum to one. Direct calculation gives
$$
(\bmu^{(0)},\A^{(0)})
=
\frac{1}{2}(\bmu^{(+)},\A^{(+)})
+
\frac{1}{2}(\bmu^{(-)},\A^{(-)}).
$$

At $(\bmu^{(0)},\A^{(0)})$, arms $1$ and $2$ have expected
rewards $0.5+0.02(1-\varepsilon)$ and
$0.5-0.02(1-\varepsilon)$, respectively. Every remaining arm
has expected reward $0.5$, so arm $1$ is uniquely optimal.
At both $(\bmu^{(+)},\A^{(+)})$ and
$(\bmu^{(-)},\A^{(-)})$, arms $1$ and $2$ have expected
rewards $0.5-0.06(1-\varepsilon)$ and
$0.5+0.06(1-\varepsilon)$, respectively, while every
remaining arm has expected reward $0.5$. Thus arm $2$ is
uniquely optimal at both instances. They are alternatives to
$(\bmu^{(0)},\A^{(0)})$, whose midpoint therefore belongs to
the convex hull of its alternative set.

This property holds on a nonempty open set. To see this, let
$$
v=
\frac{1}{2}
\bigl((\bmu^{(+)},\A^{(+)})
      -(\bmu^{(-)},\A^{(-)})\bigr).
$$
All parameter constraints and best-arm inequalities in the
construction are strict. By continuity, there is an open
neighborhood $\mathcal U\subset\I$ of
$(\bmu^{(0)},\A^{(0)})$ such that every
$(\bmu,\A)\in\mathcal U$ has arm $1$ as its unique best arm,
whereas $(\bmu,\A)+v$ and $(\bmu,\A)-v$ are valid instances
with arm $2$ as their unique best arm. Since $(\bmu,\A)$ is
their midpoint,
$$
(\bmu,\A)\in
\operatorname{conv}\bigl(\text{Alt}(\bmu,\A)\bigr)
\qquad\text{for every }(\bmu,\A)\in\mathcal U.
$$
The KL objective is nonnegative and vanishes at
$(\lambb,\B)=(\bmu,\A)$, which proves that
$F^{\mathrm{conv}}(\bmu,\A,\wb,\A\wb)=0$ for every
$\wb\in\Delta^{n-1}$.

Finally, let $\wb\in\Delta_+^{n-1}$. If
$F(\bmu,\A,\wb,\A\wb)=0$, there would be a sequence of
alternatives whose KL objective tends to zero. Since every
$w_i>0$ and every $(\A\wb)_j>0$, each categorical and reward
divergence in that objective would tend to zero. Identifiability
of the corresponding distributions would then force the
alternative parameters to converge to $(\bmu,\A)$. This is
impossible: the best arm at $(\bmu,\A)$ is unique and remains
the best arm throughout a neighborhood of that instance.
Hence $F(\bmu,\A,\wb,\A\wb)>0$, completing the proof.

\section{Proof of Convex Relaxation} \label{apd: convex-relax}
We prove the four results of Subsection~\ref{sec: relaxation}. We use the notation and assumptions of Subsection~\ref{sec: relaxation}. Alternative reward means belong to $(0,1)^k$, as required by the definition of $\I$; their limits in $[0,1]^k$ may also be used when taking infima. For the proofs, we extend the formulas for $L_i$ to arbitrary positive arm and context weights. Fix $i\ne i^*$ and write
\[
    D_{ij}=\A_{j,i}-\A_{j,i^*},\qquad
    R_l=1-\min_j\A_{j,l},\qquad
    m_i=\min\{w_i,w_{i^*}\}.
\]
The coefficient $\kappa_i$ in \eqref{eq: convex-lb} is defined by
\begin{align*}
    a_i&=\max\left\{\frac{R_i}{w_i},
                     \frac{R_{i^*}}{w_{i^*}}\right\},\\
    s_i&=\max_j
        \frac{\A_{j,i}/w_i+\A_{j,i^*}/w_{i^*}}{q_j},\\
    \kappa_i&=\frac{a_i+\sqrt{a_i^2+4\sigma^2s_i}}{2}.
\end{align*}
When $\qb=\A\wb$, we have
\begin{equation}\label{eq: relaxation context weights}
    q_j\ge w_i\A_{j,i}+w_{i^*}\A_{j,i^*}
    \qquad\text{for every }j\in[k].
\end{equation}
This inequality is used only for the approximation guarantees.

\subsection{Proof of Theorem~\ref{thm: simple lower bound}}

We first give an inequality for the change in a mean under a change of its context distribution. The following lemma is related to the transportation inequalities
of \citet{talebi2018variance}. Their Corollary~11 bounds the same
mean shift using $d(Q,P)$, whereas we use $d(P,Q)$. Their Lemma~12 uses $d(P,Q)$, but involves two truncated variance terms and the full span $\max_j f_j-\min_j f_j$. Our bound instead uses only $\operatorname{Var}_P(f)$ and the one-sided range $\max_j f_j-P^\top f$.

\begin{lemma}\label{lem: relaxation mean shift}
Let $P,Q\in\Delta^{k-1}$ and $f\in\mathbb R^k$. Define
$V=\operatorname{Var}_P(f)$ and
$R=\max_j f_j-P^\top f$. Then
\[
    (Q-P)^\top f\le\sqrt{2Vd(P,Q)}+R\,d(P,Q).
\]
If $f\in[0,1]^k$, then $R\le1-\min_jP_j$.
The inequality also allows zero entries of $P$.
\end{lemma}

\begin{proof}
Let $x=(Q-P)^\top f$ be the change in the mean. It is sufficient
to consider $x>0$ and $d(P,Q)<\infty$; the other cases are immediate.
In this case $R>0$ and $d(P,Q)>0$.
Center the values by writing $g_j=f_j-P^\top f$. Then
\[
    \sum_jP_jg_j=0,\qquad
    \sum_jP_jg_j^2=V,\qquad
    \sum_jQ_jg_j=x,\qquad g_j\le R.
\]

Introduce a positive scalar $\alpha$ to scale these centered values.
Taking $0<\alpha<1/R$ ensures that $1-\alpha g_j>0$ for every $j$,
so we can apply Jensen's inequality to their logarithms.
We leave $\alpha$ free for now and choose it at the end to minimize
the resulting upper bound on $x$.
Specifically, Jensen's inequality gives
\begin{align*}
    &\sum_{j:P_j>0}P_j
       \ln\left(\frac{Q_j}{P_j}(1-\alpha g_j)\right)\\
    &\qquad\le
       \ln\left(\sum_{j:P_j>0}Q_j(1-\alpha g_j)\right)
       \le\ln(1-\alpha x).
\end{align*}
The last inequality adds the nonnegative terms with $P_j=0$ and
uses $\sum_jQ_jg_j=x$. To identify the KL term, expand the logarithm
on the left using $\ln(ab)=\ln a+\ln b$:
\begin{align*}
    &\sum_{j:P_j>0}P_j
       \ln\left(\frac{Q_j}{P_j}(1-\alpha g_j)\right)\\
    &\qquad=
       \underbrace{\sum_{j:P_j>0}P_j\ln\frac{Q_j}{P_j}}_{-d(P,Q)}
       +\sum_jP_j\ln(1-\alpha g_j).
\end{align*}
Moving the KL term to the right and the last logarithm to the left gives
\[
    d(P,Q)\ge\sum_jP_j\ln(1-\alpha g_j)-\ln(1-\alpha x).
\]

To bound the first term, use
\[
    \ln(1-z)\ge-z-\frac{z^2}{2(1-\alpha R)}
       \qquad(z\le\alpha R<1).
\]
After averaging under $P$, the linear term vanishes and the squared
term gives $V$. Together with $-\ln(1-\alpha x)\ge\alpha x$,
this gives
\[
    d(P,Q)\ge-\frac{\alpha^2V}{2(1-\alpha R)}+\alpha x 
    \Longrightarrow
    x\le\frac{d(P,Q)}{\alpha}
               +\frac{\alpha V}{2(1-\alpha R)}.
\]
For $V>0$, choose
\[
    \alpha=\frac{\sqrt{2d(P,Q)}}
                 {\sqrt V+R\sqrt{2d(P,Q)}}.
\]
This choice lies in $(0,1/R)$, and substitution gives
$x\le\sqrt{2Vd(P,Q)}+R\,d(P,Q)$.
If $V=0$, letting $\alpha\uparrow1/R$ in the same upper bound
gives $x\le R\,d(P,Q)$, as required.

Finally, for $f\in[0,1]^k$, choose $j_0$ attaining its maximum.
The term with $j=j_0$ is zero and every other difference is at most
one. Thus
\[
    R=\sum_jP_j(f_{j_0}-f_j)
      \le1-P_{j_0}\le1-\min_jP_j.
\]
\end{proof}

\begin{proof}[Proof of Theorem~\ref{thm: simple lower bound}]
\textbf{Lower bound.}
Fix $i\ne i^*$ and an alternative $(\lambb,\B)$ satisfying
$\B_i^\top\lambb\ge\B_{i^*}^\top\lambb$.
Write $\eta_j=\lambda_j-\mu_j$, and define
\[
    C_X=w_i d(\A_i,\B_i)+w_{i^*}d(\A_{i^*},\B_{i^*}),
    \qquad
    C_Y=\frac1{2\sigma^2}\sum_jq_j\eta_j^2.
\]
Here $C_X$ is the cost of changing the two context distributions,
and $C_Y$ is a lower bound on the reward cost. Thus the total
cost of the alternative is at least $C=C_X+C_Y$.
In all sums over $l$ below, $l\in\{i,i^*\}$.
Feasibility gives
\begin{equation}\label{eq: relaxation gap decomposition}
    \Delta_i\le
       (\B_i-\A_i)^\top\lambb
       +(\A_{i^*}-\B_{i^*})^\top\lambb
       +\sum_jD_{ij}\eta_j.
\end{equation}

Apply Lemma~\ref{lem: relaxation mean shift} with $f=\lambb$
for arm $i$ and $f=\mathbf1-\lambb$ for arm $i^*$.
Since $\lambb\in[0,1]^k$, the coefficients of the linear KL terms
are bounded by $R_i$ and $R_{i^*}$, respectively.
Moreover, the triangle inequality for centered square-integrable
variables gives
\[
    \sqrt{\operatorname{Var}_{\A_l}(\lambb)}
       \le\sqrt{\operatorname{Var}_{\A_l}(\bmu)}
              +\sqrt{\operatorname{Var}_{\A_l}(\eta)}
       \le\sqrt{V_l}+\sqrt{\sum_j\A_{j,l}\eta_j^2}.
\]
Substituting these bounds into
\eqref{eq: relaxation gap decomposition} yields
\begin{align*}
    \Delta_i\le{}&
       \sum_l\sqrt{2V_l\,d(\A_l,\B_l)}
       +\left|\sum_jD_{ij}\eta_j\right|\\
       &+\sum_l\sqrt{2d(\A_l,\B_l)
                                      \sum_j\A_{j,l}\eta_j^2}
       +\sum_lR_l\,d(\A_l,\B_l).
\end{align*}

We bound the first line using weighted Cauchy--Schwarz:
\begin{align*}
    &\sum_l\sqrt{2V_l\,d(\A_l,\B_l)}
       +\left|\sum_jD_{ij}\eta_j\right|\\
    &\quad\le
       \sqrt{2\left(\frac{V_i}{w_i}
                    +\frac{V_{i^*}}{w_{i^*}}\right)C_X}
       +\sqrt{2\sigma^2\sum_j\frac{D_{ij}^2}{q_j}\,C_Y}\\
    &\quad\le\sqrt{2H_i(C_X+C_Y)}=\sqrt{2H_iC}.
\end{align*}
For the second line, the definitions of $a_i$ and $s_i$ give
\begin{align*}
    \sum_lR_l\,d(\A_l,\B_l)&\le a_iC_X,\\
    \sum_l\sqrt{2d(\A_l,\B_l)\sum_j\A_{j,l}\eta_j^2}
    &\le\sqrt{2C_X\sum_j
           \left(\frac{\A_{j,i}}{w_i}
                  +\frac{\A_{j,i^*}}{w_{i^*}}\right)\eta_j^2}\\
    &\le\sqrt{2s_iC_X\sum_jq_j\eta_j^2}
      =2\sigma\sqrt{s_iC_XC_Y}.
\end{align*}
The correction coefficient satisfies
$\kappa_i(\kappa_i-a_i)=\sigma^2s_i$. Hence
\begin{align*}
    &\kappa_i(C_X+C_Y)-a_iC_X-2\sigma\sqrt{s_iC_XC_Y}\\
    &\qquad=
       \left(\sqrt{(\kappa_i-a_i)C_X}
                       -\sqrt{\kappa_iC_Y}\right)^2\ge0.
\end{align*}
Combining the bounds proves
\begin{equation}\label{eq: relaxation scalar bound}
    \Delta_i\le\sqrt{2H_iC}+\kappa_iC.
\end{equation}
The right-hand side is increasing in $C$. Its equality solution is
$L_i(\wb,\qb)$ in \eqref{eq: convex-lb}, so every feasible
alternative has cost at least $C\ge L_i(\wb,\qb)$.
Taking infima gives $F_i\ge L_i$, and taking the minimum over
$i\ne i^*$ gives $F\ge L$.

\textbf{Concavity.}
We show that $L_i$ is jointly concave and nondecreasing in
$(w_i,w_{i^*},\qb)$. The reciprocal $1/H_i$ is a concave,
nondecreasing weighted harmonic mean. Also, the two quantities
\[
    u=\frac1{a_i}
      =\min\left\{\frac{w_i}{R_i},\frac{w_{i^*}}{R_{i^*}}\right\},
    \qquad
    v=\frac1{\sigma\sqrt{s_i}}
      =\frac1\sigma\min_j
          \sqrt{\frac{q_j}
                        {\A_{j,i}/w_i+\A_{j,i^*}/w_{i^*}}}
\]
are concave and nondecreasing. For $u$, this follows by taking the
minimum of two linear functions. For $v$, each term is the geometric
mean of $q_j$ and a concave weighted harmonic mean; the geometric
mean is itself concave and nondecreasing in both arguments.

To obtain the same properties for $1/\kappa_i$, observe that
\[
    \frac1{\kappa_i}=v h(u/v),\qquad
    h(z)=\frac{2z}{1+\sqrt{1+4z^2}}.
\]
Direct differentiation shows that $h$ is increasing and concave
on $[0,\infty)$, with $h(0)=0$. For such a function, its perspective
$(u,v)\mapsto v h(u/v)$ is concave and nondecreasing in both
arguments. Composing it with the concave functions $u$ and $v$
therefore proves that $1/\kappa_i$ is concave and nondecreasing.

Finally, let $g$ be the inverse of the increasing convex function
\[
    z\longmapsto\frac{2z}{(\Delta_i-z)^2},
       \qquad 0\le z<\Delta_i.
\]
Then $g$ is increasing and concave, with $g(0)=0$.
At equality in \eqref{eq: relaxation scalar bound}, rearrangement
shows that
\[
    L_i=\frac1{\kappa_i}\,
                 g\left(\frac{\kappa_i}{H_i}\right).
\]
This is the perspective of $g$ evaluated at
$(1/H_i,1/\kappa_i)$. Since both arguments are concave and
nondecreasing in the weights, the same composition rule proves
joint concavity and monotonicity of $L_i$.
The linear substitution $\qb=\A\wb$ and the minimum over
$i\ne i^*$ then prove concavity of
$\wb\mapsto L(\bmu,\A,\wb,\A\wb)$.

Scaling all weights by $c>0$ divides $H_i$ and $\kappa_i$ by $c$.
Thus $L_i(c\wb,c\qb)=cL_i(\wb,\qb)$.
We also have
\begin{equation}\label{eq: relaxation boundary bound}
    L_i\le\frac{\Delta_i}{\kappa_i}
          \le2\Delta_i\min\{w_i,w_{i^*}\},
\end{equation}
using $\kappa_i\ge a_i$ and $R_l\ge1/2$ for $k\ge2$.
Taking the minimum over arms shows that $L$ tends to zero when
any arm weight tends to zero, justifying its continuous zero
extension on the simplex boundary.

\textbf{Empirical statistic.}
The lower-bound argument allows zero empirical context probabilities:
Lemma~\ref{lem: relaxation mean shift} applies to them as well.
It also allows real Gaussian empirical means, since boundedness
was needed only for the alternative means. The reward-KL inequality
is applied on the corresponding empirical-to-alternative domain.
Homogeneity therefore gives
\[
    \lamh^L(t)
      =tL(\muht,\Aht,\nbarxt,\nbarzt)
      =L(\muht,\Aht,N^X(t),N^Z(t))\le\lamht.
\]
The zero convention for missing counts or a tied empirical best arm
also preserves this inequality.
\end{proof}

\subsection{Proof of Theorem~\ref{thm: gap}}

\begin{proof}
Fix $i\ne i^*$ and keep the reward means unchanged,
$\lambb=\bmu$. If $w_i\le w_{i^*}$, replacing $\A_i$ by
$\A_{i^*}$ makes the two arms tie at cost
$w_i d(\A_i,\A_{i^*})$. If $w_{i^*}<w_i$, replacing
$\A_{i^*}$ by $\A_i$ gives a tie at cost
$w_{i^*}d(\A_{i^*},\A_i)$.
Strict alternatives can be obtained by arbitrarily small
perturbations: increase arm $i$'s mean in the first case,
or decrease the best arm's mean in the second case, breaking
any remaining best-arm ties if necessary. These perturbations
are possible because all context probabilities are positive
and the context reward means are not all equal.
Since every entry of the two original columns is at least $p_i$,
\[
    d(\A_i,\A_{i^*}),\ d(\A_{i^*},\A_i)\le\ln(1/p_i).
\]
Thus, in either case,
\begin{equation}\label{eq: relaxation one column cost}
    F_i\le m_i\ln(1/p_i).
\end{equation}

We next lower bound $L_i$. Equation~\eqref{eq: relaxation context weights}
and $w_i,w_{i^*}\ge m_i$ imply
\[
    (w_i,w_{i^*},\qb)\ge m_i(1,1,\A_i+\A_{i^*})
\]
coordinatewise. Monotonicity and homogeneity from the proof of
Theorem~\ref{thm: simple lower bound} therefore give
\begin{equation}\label{eq: relaxation reference weights}
    L_i(\wb,\qb)\ge m_iL_i(1,1,\A_i+\A_{i^*}).
\end{equation}
Here $L_i(1,1,\A_i+\A_{i^*})$ means that the two relevant arm
weights are set to one and the context weights to
$\A_i+\A_{i^*}$; the other arm weights do not enter $L_i$.
At these reference weights,
\begin{align*}
    H_i&=V_i+V_{i^*}
        +\sigma^2\sum_j\frac{D_{ij}^2}{\A_{j,i}+\A_{j,i^*}}
        \le\frac12+2\sigma^2,\\
    a_i&=1-p_i\le1,\qquad s_i=1,\qquad
    \kappa_i\le\frac{1+\sqrt{1+4\sigma^2}}2.
\end{align*}
The bound on $H_i$ uses $V_l\le1/4$ and
$D_{ij}^2\le(\A_{j,i}+\A_{j,i^*})^2$, together with
$\sum_j(\A_{j,i}+\A_{j,i^*})=2$.
Since $\Delta_i\le1$ and
$\sqrt{H_i^2+2H_i\kappa_i\Delta_i}\le H_i+\kappa_i\Delta_i$,
the denominator defining $L_i$ at these reference weights is at most
\[
    2H_i+2\kappa_i\Delta_i
    \le2+4\sigma^2+\sqrt{1+4\sigma^2}=C_\sigma.
\]
Consequently,
\[
    L_i(\wb,\qb)\ge
    m_iL_i(1,1,\A_i+\A_{i^*})
    \ge\frac{m_i\Delta_i^2}{C_\sigma}.
\]
Combining this with \eqref{eq: relaxation one column cost} and
$L_i\le F_i$ gives
\[
    1\le\frac{F_i}{L_i}
    \le C_\sigma\frac{\ln(1/p_i)}{\Delta_i^2}.
\]
Each upper factor is at most $\rho(\bmu,\A)$, so
$F_i\le\rho(\bmu,\A)L_i$ for every $i\ne i^*$.
Taking minima over arms gives $F\le\rho(\bmu,\A)L$.
Together with $L\le F$, this proves the theorem.
\end{proof}

\textbf{A sharper approximation factor.}
Instead of replacing one context column by the other, we can choose
the least costly distribution that closes the gap. Define
\[
    \mathcal K_i=\max\left\{
       \inf_{\substack{P\in\Delta_+^{k-1}:\\
                P^\top\bmu\ge\A_{i^*}^\top\bmu}}d(\A_i,P),
       \quad
       \inf_{\substack{P\in\Delta_+^{k-1}:\\
                P^\top\bmu\le\A_i^\top\bmu}}d(\A_{i^*},P)
       \right\}.
\]
The first infimum is the cost of raising arm $i$ to the best arm's
mean; the second is the cost of lowering the best arm to arm $i$'s
mean. Taking the maximum covers either choice of the less-weighted
arm. The first step of the proof then gives $F_i\le m_i\mathcal K_i$.
Combining this with \eqref{eq: relaxation reference weights}, without
upper bounding the reference denominator by $C_\sigma$, gives
\begin{equation}\label{eq: relaxation sharper factor}
    1\le\frac{F_i}{L_i}
    \le\frac{\mathcal K_i}{L_i(1,1,\A_i+\A_{i^*})}.
\end{equation}
This is the sharper comparison underlying the theorem:
$\mathcal K_i\le\ln(1/p_i)$, and the reference value of $L_i$
is kept exactly. The following lemma further bounds $\mathcal K_i$
when the gap is small relative to the context variances.

\begin{lemma}\label{lem: relaxation linear tilt}
Let $P\in\Delta_+^{k-1}$, $f\in[0,1]^k$, and
$V=\operatorname{Var}_P(f)>0$. Put $r=P^\top f-\min_jf_j$.
For $0\le x<V/r$, define
\[
    Q_j=P_j\left[1+\frac{x}{V}(f_j-P^\top f)\right].
\]
Then $Q\in\Delta_+^{k-1}$, $(Q-P)^\top f=x$, and
\[
    d(P,Q)\le\frac{x^2}{2(V-rx)}.
\]
\end{lemma}

\begin{proof}
The centered values $f_j-P^\top f$ have mean zero under $P$,
so $\sum_jQ_j=1$. Moreover,
\[
    \frac{Q_j}{P_j}\ge1-\frac{rx}{V}>0,
\]
which proves positivity. The change in the mean is
\[
    (Q-P)^\top f
    =\frac{x}{V}\left(\sum_jP_jf_j^2-(P^\top f)^2\right)=x.
\]
To bound the KL divergence, write
$u_j=(x/V)(f_j-P^\top f)$, so $Q_j/P_j=1+u_j$ and
$u_j\ge-rx/V$. The inequality
\[
    -\ln(1+u_j)\le-u_j+\frac{u_j^2}{2(1-rx/V)}
\]
therefore gives
\[
    d(P,Q)=-\sum_jP_j\ln(1+u_j)
    \le\frac{\sum_jP_ju_j^2}{2(1-rx/V)}
    =\frac{x^2}{2(V-rx)}.
\]
Here the linear terms vanish because $\sum_jP_ju_j=0$, and
$\sum_jP_ju_j^2=x^2/V$.
\end{proof}

If $\Delta_i<\min\{V_i,V_{i^*}\}$, apply
Lemma~\ref{lem: relaxation linear tilt} with $x=\Delta_i$,
using $f=\bmu$ for arm $i$ and $f=\mathbf1-\bmu$ for arm $i^*$.
The first choice raises arm $i$'s mean by $\Delta_i$, while the
second lowers the best arm's mean by $\Delta_i$.
In both cases $r\le1$, so $x<V/r$ and
\[
    \mathcal K_i\le
    \frac{\Delta_i^2}{2(\min\{V_i,V_{i^*}\}-\Delta_i)}.
\]
The reference-weight calculation above also gives
\[
    L_i(1,1,\A_i+\A_{i^*})\ge\frac{\Delta_i^2}{C_\sigma}.
\]
Combining these inequalities with
\eqref{eq: relaxation sharper factor}, we obtain
\begin{align*}
    \frac{F_i}{L_i}
    &\le\frac{\mathcal K_i}{L_i(1,1,\A_i+\A_{i^*})}\\
    &\le
    \frac{\Delta_i^2}
         {2(\min\{V_i,V_{i^*}\}-\Delta_i)}
    \frac{C_\sigma}{\Delta_i^2}\\
    &=\frac{C_\sigma}
            {2(\min\{V_i,V_{i^*}\}-\Delta_i)},
\end{align*}
which proves the additional bound in Theorem~\ref{thm: gap}.

\subsection{Proof of Theorem~\ref{thm: hard relaxation}}

\begin{proof}
Fix $i\ne i^*$. Since $\sigma=1$,
\[
    H_i=\frac{V_i}{w_i}+\frac{V_{i^*}}{w_{i^*}}
                       +\sum_j\frac{D_{ij}^2}{q_j},
    \qquad
    \frac{\Delta_i}{m_iH_i}\le\Gamma\le\frac1{20}.
\]
Put $h=1+3\Gamma\le1.15$. We construct
\begin{align*}
    B_{j,i}&=\A_{j,i}\left[
       1+\frac{h\Delta_i}{w_iH_i}(\mu_j-\A_i^\top\bmu)\right],\\
    B_{j,i^*}&=\A_{j,i^*}\left[
       1-\frac{h\Delta_i}{w_{i^*}H_i}
                              (\mu_j-\A_{i^*}^\top\bmu)\right],\\
    \lambda_j&=\mu_j+\frac{h\Delta_i}{H_i}\frac{D_{ij}}{q_j},
\end{align*}
leaving the other context columns unchanged. Write
$\eta=\lambb-\bmu$ for the change in the reward means.
In all sums over $l$ below, $l\in\{i,i^*\}$.

\textbf{Feasibility.}
The two modified columns sum to one because
$\sum_j\A_{j,l}(\mu_j-\A_l^\top\bmu)=0$.
Their entries remain positive: the centered means have absolute
value at most one, and
\[
    \frac{h\Delta_i}{w_lH_i}
       \le\frac{h\Delta_i}{m_iH_i}\le h\Gamma<1.
\]
By \eqref{eq: relaxation context weights},
\[
    q_j\ge m_i(\A_{j,i}+\A_{j,i^*})\ge m_i|D_{ij}|.
\]
Consequently,
\[
    |\eta_j|=\frac{h\Delta_i}{H_i}\frac{|D_{ij}|}{q_j}
       \le\frac{h\Delta_i}{m_iH_i}
       \le1.15\Gamma<1.3\Gamma.
\]
Together with $\bmu\in[1.3\Gamma,1-1.3\Gamma]^k$, this proves
$\lambb\in(0,1)^k$.

We next show that arm $i$ beats $i^*$. Expanding its advantage gives
\begin{align*}
    (\B_i-\B_{i^*})^\top\lambb
    ={}&-\Delta_i+
       \frac{h\Delta_i}{H_i}
       \left(\frac{V_i}{w_i}+\frac{V_{i^*}}{w_{i^*}}
                              +\sum_j\frac{D_{ij}^2}{q_j}\right)\\
       &+\frac{h\Delta_i}{H_i}
          \sum_l\frac{\operatorname{Cov}_{\A_l}(\bmu,\eta)}{w_l}\\
    ={}&(h-1)\Delta_i+
       \frac{h\Delta_i}{H_i}
          \sum_l\frac{\operatorname{Cov}_{\A_l}(\bmu,\eta)}{w_l}.
\end{align*}
The first term is positive. To control the covariance terms,
use $w_l\A_{j,l}\le q_j$ to obtain
\[
    w_l\sum_j\A_{j,l}\eta_j^2
       \le\sum_jq_j\eta_j^2
       =\frac{h^2\Delta_i^2}{H_i^2}\sum_j\frac{D_{ij}^2}{q_j}.
\]
Cauchy--Schwarz gives
\[
    |\operatorname{Cov}_{\A_l}(\bmu,\eta)|
       \le\sqrt{V_l\sum_j\A_{j,l}\eta_j^2}.
\]
Using these inequalities and $w_l\ge m_i$,
\begin{align*}
    \frac{h\Delta_i}{H_i}
          \sum_l\frac{|\operatorname{Cov}_{\A_l}(\bmu,\eta)|}{w_l}
    &\le\frac{h^2\Delta_i^2}{m_iH_i^2}
           \sqrt{\sum_j\frac{D_{ij}^2}{q_j}}
           \sum_l\sqrt{\frac{V_l}{w_l}}\\
    &\le\frac{\sqrt2h^2\Delta_i^2}{m_iH_i}\\
    &\le\sqrt2h^2\Gamma\Delta_i\\
    &<3\Gamma\Delta_i=(h-1)\Delta_i.
\end{align*}
For the second inequality, we used
$\sum_jD_{ij}^2/q_j\le H_i$ and
$\sum_l\sqrt{V_l/w_l}\le\sqrt{2H_i}$.
The strict inequality follows from $\sqrt2(1.15)^2<3$.
Thus $(\B_i-\B_{i^*})^\top\lambb>0$.
If other arms tie for the best mean, an arbitrarily small
perturbation breaks those ties while preserving this strict
inequality and changing the cost by an arbitrarily small amount.

\textbf{Cost of the alternative.}
Apply Lemma~\ref{lem: relaxation linear tilt} to the modified
columns. For arm $i$, use $f=\bmu$ and
$x=h\Delta_iV_i/(w_iH_i)$; for arm $i^*$, use
$f=\mathbf1-\bmu$ and $x=h\Delta_iV_{i^*}/(w_{i^*}H_i)$.
In each application, $x/V_l\le h\Gamma<1$ and $r\le1$,
so the lemma gives
\[
    d(\A_l,\B_l)
       \le\frac{h^2\Delta_i^2V_l}
                   {2w_l^2H_i^2(1-h\Gamma)}.
\]
For unit-variance Gaussian rewards, the reward cost is exactly
$\frac12\sum_jq_j\eta_j^2$. Adding the context and reward costs,
\begin{align*}
    F_i
    &\le\sum_lw_ld(\A_l,\B_l)+\frac12\sum_jq_j\eta_j^2\\
    &\le\frac{h^2\Delta_i^2}{2H_i^2}
       \left[\frac{V_i/w_i+V_{i^*}/w_{i^*}}{1-h\Gamma}
                         +\sum_j\frac{D_{ij}^2}{q_j}\right]\\
    &\le\frac{h^2\Delta_i^2}{2H_i(1-h\Gamma)}\\
    &\le\frac{\Delta_i^2}{2H_i}
                     \frac{(1+3\Gamma)^2}{1-2\Gamma}.
\end{align*}
The last inequality uses $h\le1.15<2$.

\textbf{Comparison with $L_i$.}
The context-weight inequality also gives
\[
    m_ia_i\le1,\qquad
    m_i^2s_i
       \le\max_j\frac{m_i(\A_{j,i}+\A_{j,i^*})}{q_j}\le1.
\]
Hence, by the definition of $\kappa_i$,
\[
    m_i\kappa_i\le\frac{1+\sqrt5}{2}\le\frac53.
\]
Using
$\sqrt{H_i^2+2H_i\kappa_i\Delta_i}\le H_i+\kappa_i\Delta_i$
in the denominator of $L_i$,
\[
    L_i\ge\frac{\Delta_i^2}{2H_i+2\kappa_i\Delta_i}
       \ge\frac{\Delta_i^2}{2H_i(1+5\Gamma/3)}.
\]
Combining the upper bound on $F_i$ with this lower bound gives
\begin{equation}\label{eq: relaxation hard factor}
    \frac{F_i}{L_i}
       \le\frac{(1+5\Gamma/3)(1+3\Gamma)^2}{1-2\Gamma}.
\end{equation}
Finally, since $0<\Gamma\le1/20$,
\begin{align*}
    &(1+12\Gamma)(1-2\Gamma)
                   -(1+5\Gamma/3)(1+3\Gamma)^2\\
    &\qquad=\Gamma\left(\frac73-43\Gamma-15\Gamma^2\right)
       \ge\frac7{48}\Gamma>0.
\end{align*}
Since $1-2\Gamma>0$, \eqref{eq: relaxation hard factor} implies
$F_i\le(1+12\Gamma)L_i$ for every $i\ne i^*$.
Taking minima over arms and using
Theorem~\ref{thm: simple lower bound},
\[
    L\le F\le(1+12\Gamma)L.
\]
The ratio therefore tends to one as $\Gamma\to0$ under the stated
conditions.
\end{proof}

\subsection{Proof of Theorem~\ref{thm : relaxation upper}}

We first verify the properties needed to apply the sampling analysis
of Appendix~\ref{apd: proof-sample} to the relaxed objective.

\begin{lemma}\label{lem: relaxation optimizer}
The objective $L(\bmu,\A,\wb,\A\wb)$, with its zero boundary
extension, is jointly continuous locally around every true instance.
Its optimizer set
\[
    W_L(\bmu,\A)=\argmax_{\wb\in\Delta^{n-1}}
                         L(\bmu,\A,\wb,\A\wb)
\]
is nonempty, compact, convex, and contained in $\Delta_+^{n-1}$.
The mapping $(\bmu,\A)\mapsto W_L(\bmu,\A)$ is upper
hemicontinuous.
\end{lemma}

\begin{proof}
In a sufficiently small neighborhood of a true instance, the best
arm remains unchanged and the context probabilities stay bounded
away from zero. For positive allocations, joint continuity follows
from the formulas defining $L_i$.

To verify continuity at the allocation boundary, recall
\eqref{eq: relaxation boundary bound}:
\[
    L_i\le2\Delta_i\min\{w_i,w_{i^*}\}.
\]
If any arm weight tends to zero, at least one term in the minimum
defining $L$ therefore tends to zero. The gaps remain bounded over
nearby instances, so this convergence is uniform in those instances.
This proves joint continuity of the zero boundary extension.

Continuity on the compact simplex implies that a maximizer exists
and that the optimizer set is compact. Since the objective is concave,
its optimizer set is convex. Moreover, the objective is positive
at every interior allocation and zero on the boundary. Thus every
maximizer belongs to $\Delta_+^{n-1}$.

Since the feasible simplex is compact and independent of the instance,
Berge's maximum theorem \citep{berge1963topological} implies that
$W_L(\bmu,\A)$ is upper hemicontinuous. A similar application to
optimal bandit allocations appears in \citet[Lemma~28]{mutiple-correct-answers-degenne2019pure}.
\end{proof}

\begin{proof}[Proof of Theorem~\ref{thm : relaxation upper}]
\textbf{$\delta$-correctness.}
If the empirical best arm differs from the true best arm, the true
instance is an alternative to the empirical instance. Therefore,
Theorem~\ref{thm: simple lower bound} and the definition of the
GLR statistic give
\[
    \lamh^L(t)\le\lamht
    \le\sum_iN_i^X(t)d(\Ah_i(t),\A_i)
       +\sum_jN_j^Z(t)d(\muh_j(t),\mu_j).
\]
An erroneous recommendation can therefore occur only if the sum
on the right exceeds $\beta(t,\delta)$ at some round.
Lemma~\ref{lem: new threshold} bounds the probability of this event
by $\delta$, proving $\delta$-correctness.

\textbf{Expected stopping time.}
We apply the proof of Theorem~\ref{thm : upper} in
Appendix~\ref{apd: proof-sample}, replacing the objective $g$ in
\eqref{eq: g-function-proof} by
$L(\bmu,\A,\wb,\A\wb)$ and the optimal allocation set by
$W_L(\bmu,\A)$. Lemma~\ref{lem: relaxation optimizer} verifies
the required continuity, convexity, and upper hemicontinuity.
The exploration and concentration analysis in
Lemma~\ref{lem: event prob}, and the geometric tracking argument
in Lemma~\ref{lem: G-tracking}, apply with these replacements.

The identity $\nbarzt=\Aht\nbarxt$ gives
\[
    \lamh^L(t)
    =tL(\muht,\Aht,\nbarxt,\Aht\nbarxt).
\]
On the good events used in Appendix~\ref{apd: proof-sample},
the empirical instance is close to the true instance and the
empirical allocation is close to $W_L(\bmu,\A)$.
Every allocation in this set has objective value $1/U^*(\bmu,\A)$.
Joint continuity therefore implies that, for any
$0<\epsilon<\min\{1,1/U^*(\bmu,\A)\}$,
\[
    \lamh^L(T)\ge
    T\left(\frac1{U^*(\bmu,\A)}-\epsilon\right)
\]
for sufficiently large $T$, with the estimation and tracking
tolerances chosen sufficiently small.
By Lemma~\ref{lem: event prob}, the complementary event has
probability at most $BT\exp(-CT^{1/8})$.

The threshold and initialization are unchanged. Hence the threshold
comparison and summation of stopping-time tails in
Appendix~\ref{apd: proof-sample} give
\[
    \limsup_{\delta\to0}
    \frac{\mathbb E_{\bmu,\A}[\tau_\delta]}{\logdel}
    \le\frac1{(1-\epsilon)(1/U^*(\bmu,\A)-\epsilon)}.
\]
Letting $\epsilon\to0$ proves
\[
    \limsup_{\delta\to0}
    \frac{\mathbb E_{\bmu,\A}[\tau_\delta]}{\logdel}
    \le U^*(\bmu,\A).
\]

\textbf{Comparison with the optimal complexity.}
For every positive allocation, Theorems~\ref{thm: simple lower bound}
and~\ref{thm: gap} give
\[
    \frac{F(\bmu,\A,\wb,\A\wb)}{\rho(\bmu,\A)}
    \le L(\bmu,\A,\wb,\A\wb)
    \le F(\bmu,\A,\wb,\A\wb).
\]
These inequalities also hold on the simplex boundary, where
$L=F=0$. Indeed, a zero-weight arm's context distribution can be
changed without cost. If this arm is not the best arm, its mean
can be raised above the current best mean; if it is the best arm,
its mean can be lowered below another arm's mean. Such changes
are possible because all true arm means lie strictly between
the smallest and largest context reward means. Any remaining ties
can be broken at arbitrarily small cost.

Taking suprema over allocations gives
\[
    \frac1{\rho(\bmu,\A)T^*(\bmu,\A)}
    \le\frac1{U^*(\bmu,\A)}
    \le\frac1{T^*(\bmu,\A)}.
\]
Taking reciprocals proves
\[
    T^*(\bmu,\A)\le U^*(\bmu,\A)
    \le\rho(\bmu,\A)T^*(\bmu,\A).
\]

Finally, let $\wb^*$ be an optimal allocation for $T^*$ satisfying
the conditions of Theorem~\ref{thm: hard relaxation}, with relative
gap $\Gamma^*$. Evaluating the relaxed objective at this allocation,
\[
    \frac1{U^*}
    \ge L(\bmu,\A,\wb^*,\A\wb^*)
    \ge\frac{F(\bmu,\A,\wb^*,\A\wb^*)}{1+12\Gamma^*}
    =\frac1{(1+12\Gamma^*)T^*}.
\]
Taking reciprocals gives
$U^*\le(1+12\Gamma^*)T^*$, completing the proof.
\end{proof}

\section{Extension of Convex Relaxation \citet{al2021adaptive}} \label{apd: convex-relax-rl}

This section presents a convex relaxation inspired by the lower-bound derivation in \citet{al2021adaptive}.

\begin{lemma} \label{lem: convexify-rl}
For any instance $(\bmu, \A) \in \I$, under the assumptions of Subsection~\ref{sec: relaxation},
$\lamht$ is lower bounded by
    \begin{align*}
        \min_{i \neq i_t^*}
        \delhit{i}^2
        \left(
            \frac{1}{2N_i^X(t)}
            +\frac{1}{2N_{i_t^*}^X(t)}
            +\sum_{j=1}^k
            \frac{\big(\Ah_{j,i}(t)-\Ah_{j,i_t^*}(t)\big)^2}
                 {CN_j^Z(t)}
        \right)^{-1},
    \end{align*}
where $C=1/(2\sigma^2)$, and $T^*(\bmu,\A)\le U_1(\bmu,\A)$, with $U_1$ defined by the refined expression at the end of the proof.
\end{lemma}

\begin{proof}
    First, note that GLR statistic $\lamht$ can be written as
    \begin{align*}
        \lamht
        &= \inf_{(\lambb,\B)\in\text{Alt}(\muht,\Aht)}
        \sum_{s\in[n]}N_s^X(t)d\paran{\Ah_s(t),\B_s}
        +\sum_{j\in[k]}N_j^Z(t)d\paran{\muh_j(t),\lambda_j} \\
        &= \min_{i\neq\istart}
        \inf_{(\lambb,\B)\in C_i(\muht,\Aht)}
        \left[
        \sum_{s\in[n]}N_s^X(t)d\paran{\Ah_s(t),\B_s}
        +\sum_{j\in[k]}N_j^Z(t)d\paran{\muh_j(t),\lambda_j}
        \right],
    \end{align*}
    where $C_i\triangleq\{(\lambb,\B)\mid
    \B_i^\top\lambb>\B_{\istart}^\top\lambb\}$.
    This is because $\text{Alt}(\muht,\Aht)=\bigcup_{i\neq\istart}C_i(\muht,\Aht)$.
    Note that for each $i\neq\istart$, the other columns of $\B$ can be kept equal to those of $\Aht$ when taking the infimum.
    Then, for each such $i$, we can write the optimization problem over $C_i$ as
    \begin{align*}
        \inf_{(\lambb,\B)\in\I}\quad&
        N_i^X(t)d(\Ah_i(t),\B_i)
        +N_{\istart}^X(t)d(\Ah_{\istart}(t),\B_{\istart}) \\
        &\qquad+\sum_{j\in[k]}N_j^Z(t)d\paran{\muh_j(t),\lambda_j},\\
        \text{s.t.}\quad&
        \B_i^\top\lambb>\B_{\istart}^\top\lambb.
    \end{align*}
    The constraint then can be written equivalently as
    \begin{align*}
        \B_i^\top\lambb
        &>\B_{\istart}^\top\lambb+\delhit{i}
        +\Ah_i(t)^\top\muht-\Ah_{\istart}(t)^\top\muht\\
        \Longleftrightarrow\quad\delhit{i}
        &<(\B_i-\B_{\istart})^\top\lambb
        -(\Ah_i(t)-\Ah_{\istart}(t))^\top\muht,
    \end{align*}
    which is equivalent to
    \begin{align*}
        \delhit{i}
        &<(\B_i-\Ah_i(t))^\top \lambb
        +(\Ah_{\istart}(t)-\B_{\istart})^\top\lambb\\
        &\qquad+\sum_{j=1}^k
        \big(\Ah_{j,i}(t)-\Ah_{j,\istart}(t)\big)
        (\lambda_j-\muh_j(t)).
    \end{align*}
    Intuitively, this formulation of the constraint divides the \emph{contribution} of the change of $\Ah_i(t),\Ah_{\istart}(t)$, and $\muht$ for compensating the $\delhit{i}$ gap between arms $i$ and $\istart$.
    Then we can write these contributions as a fraction of $\delhit{i}$ as
    \begin{align*}
        \alpha_i\delhit{i}
        &\triangleq(\B_i-\Ah_i(t))^\top \lambb,\\
        \alpha_{\istart}\delhit{i}
        &\triangleq(\Ah_{\istart}(t)-\B_{\istart})^\top\lambb,\\
        \forall j\in[k]:\quad \gamma_j\delhit{i}
        &\triangleq \big(\Ah_{j,i}(t)-\Ah_{j,\istart}(t)\big)
        (\lambda_j-\muh_j(t)).
    \end{align*}
    Then the constraint $\B_i^\top\lambb>\B_{\istart}^\top\lambb$ is equivalent to
    \[
        \alpha\triangleq\alpha_i+\alpha_{\istart}+\sum_{j=1}^k\gamma_j>1.
    \]

    We first consider positive counts and a unique empirical best arm; otherwise, we use the lower bound zero. The coefficients $\alpha_i,\alpha_{\istart},\gamma_j$ are real-valued and need not be nonnegative.
    We now derive a lower bound on GLR by bounding the terms in the inner infimum.

    \textbf{Bounding $N_i^X(t)\,d(\Ah_i(t),\B_i)$.}
    For this term, we have
    \begin{align*}
        |\alpha_i| \delhit{i}
        &=\left|(\B_i-\Ah_i(t))^\top\lambb\right|\\
        &\le\tfrac12\|\B_i-\Ah_i(t)\|_1
        \le\tfrac12\sqrt{2d(\Ah_i(t),\B_i)},
    \end{align*}
    which implies
    \[
        2\alpha_i^2\delhit{i}^2\le d(\Ah_i(t),\B_i).
    \]
    The first inequality follows from the fact that the entries of $(\B_i-\Ah_i(t))$ sum to zero and that
    $\lambb\in(0,1)^k$, since the alternative instance belongs to $\I$.
    The second inequality is a direct consequence of Pinsker's inequality.
    Multiplying both sides by $N_i^X(t)$ yields
    \begin{equation}\label{eq: convex first bound}
        2N_i^X(t)\alpha_i^2\delhit{i}^2
        \le N_i^X(t)d(\Ah_i(t),\B_i).
    \end{equation}

    \textbf{Bounding $N_{\istart}^X(t)\,d(\Ah_{\istart}(t),\B_{\istart})$.}
    Similar to the previous bound, we have
    \begin{align*}
       |\alpha_{\istart}|\delhit{i}
        &=\left|(\Ah_{\istart}(t)-\B_{\istart})^\top\lambb\right|\\
        &\le\tfrac12\|\Ah_{\istart}(t)-\B_{\istart}\|_1
        \le\tfrac12\sqrt{2d(\Ah_{\istart}(t),\B_{\istart})},\\
        \Rightarrow\quad
        2\alpha_{\istart}^2\delhit{i}^2
        &\le d(\Ah_{\istart}(t),\B_{\istart}).
    \end{align*}
    And after multiplying both sides by $N_{\istart}^X(t)$ we get:
    \begin{equation}\label{eq: convex second bound}
        2N_{\istart}^X(t)\alpha_{\istart}^2\delhit{i}^2
        \le N_{\istart}^X(t)d(\Ah_{\istart}(t),\B_{\istart}).
    \end{equation}

    \textbf{Bounding $N_j^Z(t)\,d(\muh_j(t),\lambda_j)$.}
    By the reward-divergence assumption already imposed in
    Subsection~\ref{sec: relaxation},
    \[
        d(\muh_j(t),\lambda_j)
        \ge\frac{(\muh_j(t)-\lambda_j)^2}{2\sigma^2}
        =C(\muh_j(t)-\lambda_j)^2,
    \]
    where $C=1/(2\sigma^2)$. Hence,
    \[
        |\muh_j(t)-\lambda_j|
        \le\sqrt{\frac1C d(\muh_j(t),\lambda_j)}.
    \]

    Now, we can bound $\gamma_j\delhit{i}$ as
    \begin{align*}
        |\gamma_j|\delhit{i}
        &=|\Ah_{j,i}(t)-\Ah_{j,\istart}(t)|\,|\lambda_j-\muh_j(t)|\\
        &\le |\Ah_{j,i}(t)-\Ah_{j,\istart}(t)|
        \sqrt{\frac1C d(\muh_j(t),\lambda_j)}.
    \end{align*}
    Write $D_{ij}(t)=\Ah_{j,i}(t)-\Ah_{j,\istart}(t)$ and
    $J_i(t)=\{j\in[k]:D_{ij}(t)\ne0\}$.
    If $D_{ij}(t)=0$, then $\gamma_j=0$, so that coordinate contributes nothing.
    After multiplying both sides of the squared inequality by $N_j^Z(t)$, we obtain
    \begin{equation}\label{eq: convex third bound}
        \frac{CN_j^Z(t)}{D_{ij}(t)^2}
        \gamma_j^2\delhit{i}^2
        \le N_j^Z(t)d(\muh_j(t),\lambda_j),
        \qquad j\in J_i(t).
    \end{equation}

    \textbf{Combining the Bounds.}
    Inserting the lower bounds shown in \eqref{eq: convex first bound}, \eqref{eq: convex second bound}, and \eqref{eq: convex third bound} into our optimization problem, we obtain the following lower bound:
    \begin{align*}
        &\inf_{(\lambb,\B)\in C_i(\muht,\Aht)}
        \left[
        \sum_{s\in[n]}N_s^X(t)d(\Ah_s(t),\B_s)
        +\sum_{j\in[k]}N_j^Z(t)d(\muh_j(t),\lambda_j)
        \right]\\
        &\ge\inf_{\alpha>1}
        \left[
        2N_i^X(t)\alpha_i^2\delhit{i}^2
        +2N_{\istart}^X(t)\alpha_{\istart}^2\delhit{i}^2
        +\sum_{j\in J_i(t)}
       \frac{CN_j^Z(t)}{D_{ij}(t)^2}
        \gamma_j^2\delhit{i}^2
        \right].
    \end{align*}
 Here the infimum on the right is over real contribution coefficients with $\alpha_i+\alpha_{\istart}+\sum_{j\in J_i(t)}\gamma_j>1$.

    We now show that, for any strictly positive vector
    $\mathbf b\in (0,\infty)^m$, the optimal value of the problem
    \[
        \inf_{a\in \mathbb R^m:\,
        \sum_{r=1}^m a_r > 1}
        \sum_{r\in[m]}a_r^2b_r
    \]
    is equal to $\left(\sum_{r\in[m]}1/b_r\right)^{-1}$. By scaling and continuity, its infimum equals the minimum on the boundary $\sum_r a_r=1$.
    This follows from the method of Lagrange multipliers: the first-order condition of the Lagrangian
    \[
        \mathcal L(a,\lambda)
        =\sum_{r=1}^m b_ra_r^2+\lambda\left(1-\sum_{r=1}^m a_r\right)
    \]
    yields $a_r=\lambda/(2b_r)$ for all $r$.
    Enforcing the constraint $\sum_{r=1}^m a_r=1$ gives
    $\lambda=2\left(\sum_{r=1}^m1/b_r\right)^{-1}$,
    and substituting back gives the minimum value
    $\left(\sum_{r=1}^m1/b_r\right)^{-1}$.

    Using this result, we obtain
    \begin{align*}
        &\inf_{\alpha>1}
        \left[
        2N_i^X(t)\alpha_i^2\delhit{i}^2
        +2N_{\istart}^X(t)\alpha_{\istart}^2\delhit{i}^2
        +\sum_{j\in J_i(t)}
            \frac{CN_j^Z(t)}{D_{ij}(t)^2}
        \gamma_j^2\delhit{i}^2
        \right]\\
        &=\delhit{i}^2\left(
        \frac1{2N_i^X(t)}+\frac1{2N_{\istart}^X(t)}
        +\sum_{j=1}^k\frac{D_{ij}(t)^2}{CN_j^Z(t)}
        \right)^{-1}.
    \end{align*}
    Substituting this inequality into the GLR formulation yields
    \begin{align*}
        \lamht
        &=\min_{i\neq\istart}
        \inf_{(\lambb,\B)\in C_i(\muht,\Aht)}
        \left[
        \sum_{s\in[n]}N_s^X(t)d(\Ah_s(t),\B_s)
        +\sum_{j\in[k]}N_j^Z(t)d(\muh_j(t),\lambda_j)
        \right]\\
        &\ge\min_{i\neq\istart}\delhit{i}^2
        \left(
        \frac1{2N_i^X(t)}+\frac1{2N_{\istart}^X(t)}
        +\sum_{j=1}^k
        \frac{\big(\Ah_{j,i}(t)-\Ah_{j,\istart}(t)\big)^2}
        {CN_j^Z(t)}
        \right)^{-1}.
    \end{align*}

    Finally, replacing $N_s^X(t)$ by $w_s$ and $N_j^Z(t)$ by
    $\sum_{s\in[n]}w_s\A_{j,s}$, and applying the same argument to the population instance $(\bmu,\A)$,
    we directly obtain
    \begin{align*}
        T^*(\bmu,\A)^{-1}
        &=\sup_{\wb\in\Delta^{n-1}}
        \min_{i\neq i^*}
        \inf_{(\lambb,\B)\in C_i(\bmu,\A)}
        \left[
        \sum_{s\in[n]}w_sd(\A_s,\B_s)\right.\\
        &\hspace{7em}\left.
        +\sum_{j\in[k]}\left(\sum_{s\in[n]}w_s\A_{j,s}\right)
        d(\mu_j,\lambda_j)
        \right]\\
        &\ge\sup_{\wb\in\Delta^{n-1}}
        \min_{i\neq i^*}
        \Delta_i^2
        \left(
        \frac1{2w_i}+\frac1{2w_{i^*}}
        +\sum_{j=1}^k
        \frac{(\A_{j,i}-\A_{j,i^*})^2}
        {C\sum_{s\in[n]}w_s\A_{j,s}}
        \right)^{-1}.
    \end{align*}
    Defining $U_1(\bmu,\A)^{-1}$ by the right-hand side,
    therefore, $T^*(\bmu,\A)\le U_1(\bmu,\A)$, which completes the proof.
\end{proof}

\section{Additional Experiments} \label{apd: experiments}

We first present further details on the experimental setups in the main text and then present the results of more experiments. 

\subsection{Details on GTS-SG Baseline.} We describe the baseline \textbf{GTS-SG} algorithm used in our experiments. In all settings, the reward distribution ($Y \vert Z$) is a Gaussian distribution with unit variance, and the reward means lie in $(0,1)$. The distribution $\pr(Y \mid X = i)$ is a mixture of $k$ Gaussian distributions. In the following lemma, we further show that $Y \mid X = i$ is a sub-Gaussian with parameter $\frac{5}{4}$.

\begin{lemma}\label{lem: sub-gaussian}
    Assume that $X_1, X_2, \ldots, X_k$ are Gaussian random variables with parameter $\sigma = 1$ and means $\mathbb{E}[X_i] = \mu_i \in [a, b]$. Let $X$ denote a mixture of these variables, where $X = X_i$ with probability $p_i$. Then, $X$ is a sub-Gaussian with parameter $\sigma^2 = 1 + \frac{(b - a)^2}{4}$.
\end{lemma}

\begin{proof}
    The result follows by writing $X = Y + \epsilon$, where $Y$ is a discrete random variable satisfying $\pr(Y = \mu_i) = p_i$, and $\epsilon$ is a standard normal variable. Since $Y$ is bounded random variable in [a, b], it is a $\tfrac{(b - a)}{2}$-sub-Gaussian and $\epsilon$ is $1$-sub-Gaussian. Therefore, since $Y$ and $\epsilon$ are independent variable, their sum is sub-Gaussian with parameter $\sigma^2 = 1 + \tfrac{(b - a)^2}{4}$.
\end{proof}

Using this result, the GTS-SG algorithm ignores the post-action context and assumes that each arm’s reward follows an independent sub-Gaussian distribution with parameter $\sigma^2 = \tfrac{5}{4}$. This simplification prevents optimality, as it neglects shared reward parameters across arms.

We implement GTS-SG as a standard track and stop algorithm \citep{track-stop-garivier2016optimal} performing on an instance with Gaussian reward distributions with $\sigma ^2 = \frac{5}{4}$. For the stopping thresholds, we used the results from~\citet{tuynman2025batch}, which develops stopping rules for BAI with sub-Gaussian reward distributions. The stopping threshold presented in that work and used in our experiments is given by  
\begin{align}\label{eq: beta sub-gaussian}
    \beta_{\text{SG}}(t,\delta)
    = \frac{n}{2}\,\overline{W}\!\left(
        2\ln\!\left(\frac{e\pi^2}{6}\right)
        + \frac{2}{n}\ln\!\left(
            \prod_{i=1}^{n}\bigl(1+\ln N_i^X(t)\bigr)^2
        \right)
        + \frac{2}{n}\ln\frac{1}{\delta}
    \right).
\end{align}
where $\overline{W}(x) \approx x + \ln(x)$; the exact definition can be found in~\citet{tuynman2025batch}.

\subsection{Numerical Evaluation of $F$.} \label{apd: exp reference}
We compute reference values of the transportation cost $F$ and its pairwise
components $F_i$ for Figures~\ref{fig: tightness}, \ref{fig: glr trace},
and~\ref{fig: heatmap} and Table~\ref{table: tightness}.
For $k \geq 3$, these values are used only to evaluate the bounds; the
numerical solver described below is not used by any algorithm.

For $k = 2$, we compute $F_i$ exactly using
Proposition~\ref{prop: k2 decomposition}.
An alternative can make arm $i$ better than $i^*$ by reversing either the
order of the two reward means or the order of the first-context probabilities
of arms $i$ and $i^*$.
For Gaussian rewards, the minimum cost of each type of reversal has a closed
form, and $F_i$ is the smaller of the two costs.

For $k \geq 3$, we compute a numerical reference for each $F_i$ by minimizing
over the alternative context distributions $(\B_i, \B_{i^*})$.
For each candidate pair, we compute the optimal alternative reward means
$\lambb$ exactly, leaving an objective in the context distributions alone.
We minimize this objective using BFGS \citep{fletcher1987practical} followed
by Nelder--Mead \citep{nelder1965simplex}.
We initialize the search at $(\A_i, \A_{i^*})$, at the optimal context
distributions obtained with the reward means fixed at $\lambb = \bmu$,
and at $20$ random pairs of context distributions.
We retain the smallest cost across these runs and take the minimum over
$i \neq i^*$ to obtain the reference for $F$.
The local search does not certify a global minimum, but each retained cost
corresponds to a feasible alternative and therefore upper-bounds $F_i$.
Consequently, the reported ratios of the reference values to the lower bounds
can overestimate, but cannot underestimate, the true approximation ratios.

\subsection{Tightness of the Relaxation.} \label{apd: exp tightness}
We compare the bounds $L_i$ and $L_{\mathrm{naive}, i}$ with the reference value
of $F_i$ for a fixed allocation $\wb$ and $\qb = \A \wb$, on two families of
instances.

\textbf{Random instances.} We draw $600$ instances with $n$ and $k$ uniform in
$\{2, \ldots, 5\}$, $\bmu$ uniform in $[0,1]^k$, and each column of $\A$ from a
Dirichlet distribution with all parameters equal to $\alpha$, where $\alpha$ is
uniform in $\{0.3, 1, 3\}$.
A small $\alpha$ produces context probabilities close to zero.
The allocation $\wb$ is drawn uniformly from the simplex.

\textbf{Hard instances.} For each $k \in \{3, 4, 5\}$, we draw four instances
with $n = 3$, $\bmu$ uniform in $[0,1]^k$, and columns $\A^0_i$ uniform on the
simplex.
For each value of $\epsilon$ in Figure~\ref{fig: tightness}, the best arm keeps
its context distribution, $\A_{i^*} = \A^0_{i^*}$, and that of each suboptimal
arm is replaced by $\A_i = (1 - \epsilon) \A_{i^*} + \epsilon \A^0_i$.
The allocation $\wb$ is uniform.
The parameter $\epsilon \in (0, 1]$ controls the hardness of the instance.
It is the fraction of the original difference between arm $i$ and the best arm
that remains, since $\A_i - \A_{i^*} = \epsilon (\A^0_i - \A_{i^*})$.
Every gap is therefore multiplied by $\epsilon$, while the reward means and the
variances $V_i$ keep the same order.
A smaller $\epsilon$ thus gives a harder instance, and a smaller $\Gamma$ in
Theorem~\ref{thm: hard relaxation}.
Figure~\ref{fig: tightness} shows the median and the interquartile range over
the resulting pairs $(i, i^*)$.

Table~\ref{table: tightness} reports the median and the $90\%$ quantile of the
two ratios over the pairs $(i, i^*)$ of each family.
On all pairs, we observed $L_i \leq F_i \leq \rho_i L_i$, where $\rho_i$ is the
factor guaranteed by Theorem \ref{thm: gap}.
On the random instances, $L_i$ is within a factor of two of $F_i$ for a majority
of the pairs and within a factor of five for almost all of them.
The naive bound is looser on most pairs and much more variable: its upper
quantiles are several times larger, for every value of $\alpha$.
The two bounds do not dominate each other, however, and $L_{\mathrm{naive}, i}$
is the larger one on a minority of the pairs.
On the hard instances, both the median and the upper quantile of $F_i / L_i$
decrease to one with $\epsilon$, while $F_i / L_{\mathrm{naive}, i}$ does not
improve (Figure~\ref{fig: tightness}).

\begin{table}[ht]
    \centering
    \caption{Tightness of the relaxations: median of the ratio of $F_i$ to each
    lower bound over the pairs $(i, i^*)$, with the $90\%$ quantile in
    parentheses.}
    \label{table: tightness}
    \small
    \begin{tabular}{l|cc}
        \toprule
        Instances & $F_i / L_{\mathrm{naive}, i}$ & $F_i / L_i$ \\
        \hline
        Random, $\alpha = 0.3$ & $2.03$ $(11.5)$ & $1.74$ $(3.83)$ \\
        Random, $\alpha = 1$ & $3.02$ $(9.5)$ & $1.92$ $(3.32)$ \\
        Random, $\alpha = 3$ & $4.01$ $(18.1)$ & $1.86$ $(3.29)$ \\
        Random, all & $2.93$ $(12.5)$ & $1.84$ $(3.46)$ \\
        \hline
        Hard, $\epsilon = 0.3$ & $4.21$ $(10.5)$ & $1.53$ $(2.37)$ \\
        Hard, $\epsilon = 0.1$ & $5.57$ $(15.5)$ & $1.22$ $(1.62)$ \\
        Hard, $\epsilon = 0.03$ & $5.75$ $(19.7)$ & $1.07$ $(1.19)$ \\
        Hard, $\epsilon = 0.01$ & $5.79$ $(20.6)$ & $1.02$ $(1.06)$ \\
        Hard, $\epsilon = 0.003$ & $5.80$ $(20.9)$ & $1.01$ $(1.02)$ \\
        Hard, $\epsilon = 0.001$ & $5.80$ $(21.0)$ & $1.00$ $(1.01)$ \\
        \bottomrule
    \end{tabular}
\end{table}

\subsection{Instance details for Figure~\ref{fig: glr trace}.} \label{apd: exp trace}
Figure~\ref{fig: glr trace} uses the following instance with $n = 3$ arms and
$k = 3$ contexts.
It is run with $\delta = 10^{-3}$ and $\delta = 10^{-9}$.
\begin{align*}
    \bmu = \begin{bmatrix} 0.58 \\ 0.87 \\ 0.88 \end{bmatrix}, \quad
    \A = \begin{bmatrix}
        0.52 & 0.04 & 0.36 \\
        0.11 & 0.31 & 0.23 \\
        0.37 & 0.65 & 0.41
    \end{bmatrix}.
\end{align*}

\subsection{Comparison optimal frequency with different prior knowledge.} 
We conduct an experiment to demonstrate that different levels of prior knowledge about the environment, knowing $\A$, knowing $\bmu$, knowing neither, or ignoring the post-action context, affect not only the sample complexity bounds but also the optimal sampling frequencies. This indicates that the optimal policy can differ substantially across these settings.

\begin{figure} [t]
    \centering
    \includegraphics[width=\linewidth]{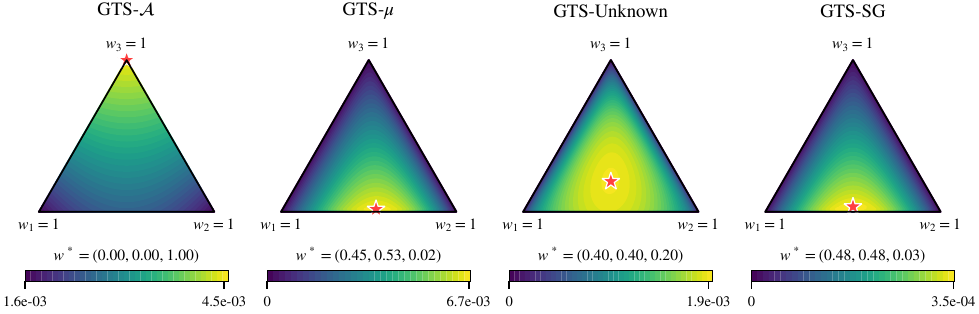}
    \caption{Heatmaps of $F(\wb)$ across the simplex for different prior
    knowledge levels.}
    \label{fig: heatmap}
\end{figure}

Consider the following instance:  
\begin{align}
    &\bmu = \begin{bmatrix}
        0.9 \\
        0.5 \\
        0.3
    \end{bmatrix}, \quad
    \A = \begin{bmatrix}
        0.50 & 0.50 & 0.10\\
        0.40 & 0.10 & 0.45\\
        0.10 & 0.40 & 0.45
    \end{bmatrix}.
\end{align}
For all cases with different prior knowledge, the sample complexity lower bound
takes the form $\mathbb{E}_{\bmu, \A}[\tau_{\delta}] \geq T^*(\bmu,
\A)d_B(\delta, 1- \delta)$, where $(T^*(\bmu, \A))^{-1} = \sup_{\wb \in
\Delta^{n-1}} F(\wb)$, with $F$ defined differently in each setting.
Since $n = 3$ in this example, we can visualize the function $F(\wb)$ over the
simplex $\Delta^2$.

Figure~\ref{fig: heatmap} illustrates the value of $F(\wb)$ for the four
algorithms introduced in the main text, each corresponding to a different
setting of prior knowledge.
The figure clearly shows that the optimal sampling policy can vary significantly
depending on the available information.
Therefore, the algorithms achieving optimal sample complexity may exhibit
entirely different behaviors.

\subsection{Sample Complexity Comparison.} \label{apd: exp large}
We further compared the sample complexity of our algorithm on larger random
instances, i.e., larger values $n$ and $k$.
Note that the sample complexity of algorithms for the problem, heavily depends
on the gaps between the arms, thus bigger instances do not necessarily have
higher sample complexity.

\begin{figure}[p]
    \centering
    \includegraphics[width=0.94\textwidth]{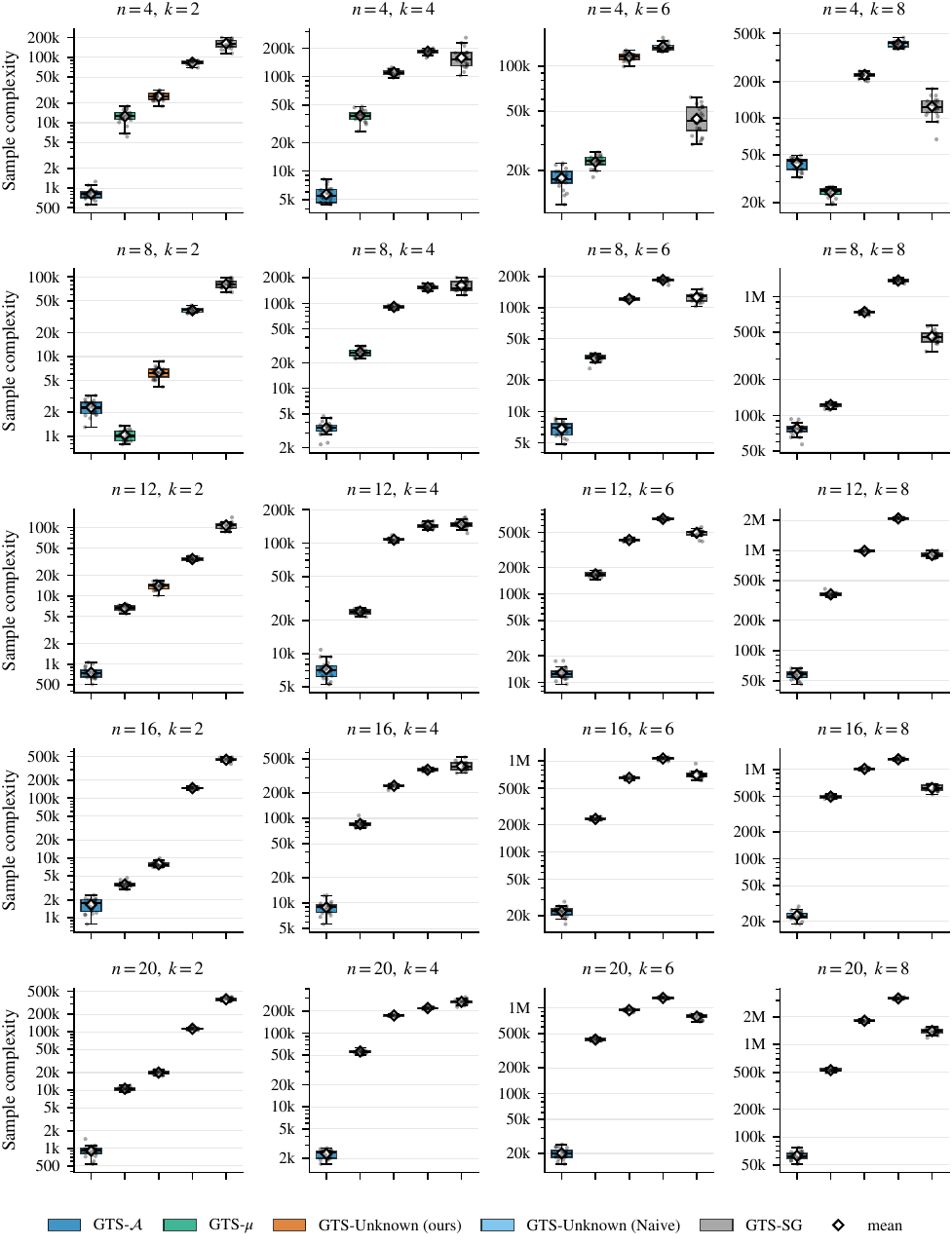}
    \caption{Sample complexity on random instances with $n$ arms and $k$
    contexts, $\delta = 10^{-3}$, $20$ runs per algorithm.
    Boxes show the quartiles, dots the individual runs, and diamonds the means.}
    \label{fig: sample-comp-appendix}
\end{figure}

We run the five algorithms of Figure~\ref{fig: sample-comp-main} on $20$ random
instances, one for each $n \in \{4, 8, 12, 16, 20\}$ and $k \in \{2, 4, 6, 8\}$,
with $\delta = 10^{-3}$ and $20$ runs per algorithm.
The entries of $\bmu$ are uniform in $[0,1]$ and each column of $\A$ is a vector
of $k$ independent uniform variables, normalized to sum to one. To keep the running times moderate, we only keep instances in which the gap between the two best arms is at least $0.05$ and all entries of $\A$ are at least $0.005$.
Figure~\ref{fig: sample-comp-appendix} shows the results. 

\textbf{Improvement over GTS-Unknown (Naive).} GTS-Unknown (ours) achieves a
lower mean stopping time than the naive variant on every tested instance,
typically using only one half to two thirds as many samples.
This consistent improvement demonstrates the practical benefit of our
relaxation and stopping threshold when both $\bmu$ and $\A$ must be learned.

\textbf{Effect of knowing $\bmu$ or $\A$.} GTS-$\bmu$ and GTS-$\A$ show how
prior knowledge can further reduce the learning task.
When $\bmu$ is known, only the context distributions must be estimated;
when $\A$ is known, only the shared reward means must be estimated.
The value of each type of prior knowledge depends on the instance:
closely spaced context distributions make knowledge of $\A$ particularly
useful, whereas closely spaced reward means make knowledge of $\bmu$ more
useful, as illustrated in Figure~\ref{fig: sample-comp-main}.
GTS-Unknown (ours) addresses the general setting in which neither component
is available and still improves consistently over the naive variant.

\textbf{Benefit of exploiting post-action context.} Our approach also yields
substantial gains over GTS-SG without requiring prior knowledge of either
$\bmu$ or $\A$.
With two contexts, it often uses more than an order of magnitude fewer samples,
and with four contexts it improves on GTS-SG on every tested instance.
In the oracle setting, our approach attains the optimal asymptotic complexity
$T^*(\bmu,\A)$, which is no larger than the complexity achievable by GTS-SG,
since an algorithm using post-action context can also choose to ignore it.
At the tested confidence level $\delta = 10^{-3}$, however, estimating the
$n(k-1)$ free parameters of $\A$ introduces an overhead that can outweigh
the information gains for larger context spaces, allowing GTS-SG to stop
earlier.
For fixed $n$ and $k$, the corresponding terms in $\bettdel$ grow only
logarithmically with the stopping time and are lower-order relative to
$\logdel$ as $\delta \to 0$, so this threshold overhead does not contribute
to the leading asymptotic complexity.
For the implementation using our convex relaxation, the asymptotic complexity
is controlled by the approximation guarantee in
Theorem~\ref{thm : relaxation upper}.
\end{document}